\documentclass[sigconf]{aamas} 

\usepackage{balance} 
\usepackage{subcaption}
\usepackage[most]{tcolorbox}
\usepackage{tikz}
\usepackage{multirow}

\setcopyright{ifaamas}
\acmConference[AAMAS '24]{Proc.\@ of the 23rd International Conference
on Autonomous Agents and Multiagent Systems (AAMAS 2024)}{May 6 -- 10, 2024}
{Auckland, New Zealand}{N.~Alechina, V.~Dignum, M.~Dastani, J.S.~Sichman (eds.)}
\copyrightyear{2024}
\acmYear{2024}
\acmDOI{}
\acmPrice{}
\acmISBN{}

\acmSubmissionID{214}

\title[AAMAS-2024 Formatting Instructions]{LLM-Powered Hierarchical Language Agent for Real-time Human-AI Coordination}

\author{
Jijia Liu$^{1*}$, 
Chao Yu$^{1*}$, 
Jiaxuan Gao$^{1*}$, 
Yuqing Xie$^{1}$, 
Qingmin Liao$^{1}$, 
Yi Wu$^{1}$, 
Yu Wang$^{1}$
}
\affiliation{
\institution{$^1$ Tsinghua University, China}
\country{liujj23@mails.tsinghua.edu.cn, \hspace{0.1mm} \{zoeyuchao, samjia2000\}@gmail.com}
}
\email{}

\newcommand{\slowmind}{{Slow Mind}}
\newcommand{\intstage}{{Intention Reasoning Stage}}
\newcommand{\assestage}{{Chat \& Assessment Stage}}
\newcommand{\fastmind}{{Fast Mind}}
\newcommand{\execmind}{{Executor}}
\newcommand{\method}{{Hierarchical Language Agent}}
\newcommand{\mapring}{\emph{Ring}}
\newcommand{\mappart}{\emph{Partition}}
\newcommand{\mapbot}{\emph{Bottleneck}}
\newcommand{\maphard}{\emph{Quick}}
\newcommand{\purelarge}{{Slow-Mind-Only Agent}}
\newcommand{\puresmall}{{Fast-Mind-Only Agent}}
\newcommand{\purellm}{{No-Executor Agent}}

\makeatletter
\newcommand{\DrawLine}{%
  \begin{tikzpicture}
  \path[use as bounding box] (0,0) -- (\linewidth,0);
  \draw[color=gray!150!black,dashed,dash phase=2pt]
        (0-\kvtcb@leftlower-\kvtcb@boxsep,0)--
        (\linewidth+\kvtcb@rightlower+\kvtcb@boxsep,0);
  \end{tikzpicture}%
  }
\makeatother

\newcommand{\BibTeX}{\rm B\kern-.05em{\sc i\kern-.025em b}\kern-.08em\TeX}

\begin{abstract}
AI agents powered by Large Language Models (LLMs) have made significant advances, enabling them to assist humans in diverse complex tasks and leading to a revolution in human-AI coordination. LLM-powered agents typically require invoking LLM APIs and employing artificially designed complex prompts, which results in high inference latency. While this paradigm works well in scenarios with minimal interactive demands, such as code generation, it is unsuitable for highly interactive and real-time applications, such as gaming. Traditional gaming AI often employs small models or reactive policies, enabling fast inference but offering limited task completion and interaction abilities. In this work, we consider Overcooked as our testbed where players could communicate with natural language and cooperate to serve orders. We propose a Hierarchical Language Agent (HLA) for human-AI coordination that provides both strong reasoning abilities while keeping real-time execution. In particular, HLA adopts a hierarchical framework and comprises three modules: a proficient LLM, referred to as Slow Mind, for intention reasoning and language interaction, a lightweight LLM, referred to as Fast Mind, for generating macro actions, and a reactive policy, referred to as Executor, for transforming macro actions into atomic actions. Human studies show that HLA outperforms other baseline agents, including slow-mind-only agents and fast-mind-only agents, with stronger cooperation abilities, faster responses, and more consistent language communications.
\end{abstract}

\keywords{Large Language Models; Language Agents; Real-time Human-AI Coordination; Hierarchical Reasoning and Planning}

\begin{document}


\pagestyle{fancy}
\fancyhead{}


\maketitle
\def\thefootnote{*}\footnotetext{These authors contributed equally to this work}

\section{Introduction}

Developing Artificial Intelligence (AI) agents that can attain human-level performance has been a long-standing goal for AI research~\cite{wooldridge1995intelligent,goodwin1995formalizing}. Large Language Models (LLMs)~\cite{openai2023gpt} have emerged as promising tools in this endeavor, owing to their strong reasoning and generalization abilities. LLM-powered AI agents have exhibited significant potential across diverse domains, including code generation~\cite{li2023starcoder,rozière2023code,shen2023pangucoder2,luo2023wizardcoder}, content creation~\cite{ramesh2021zero,ramesh2022hierarchical,park2023generative}, tool utilization~\cite{nakano2022webgpt,schick2023toolformer,zhou2023agents,qin2023toolllm}, and robotics~\cite{ichter2022do,10160591,wang2023voyager,driess2023palme}. 
Meanwhile, LLM-powered agents have exhibited the ability to mimic human-like behaviors when interacting with other players~\citep{park2023generative,gao2023s}, leading to a revolution in human-AI coordination.

AI agents powered by LLMs commonly rely on LLM APIs and hand-crafted complex prompts. A notable challenge within this paradigm is the high latency associated with the inference process due to API calls, ranging from seconds to minutes \cite{bommasani2023holistic}. The inference time may not be regarded as bottleneck in scenarios with low-frequency interactions, such as code generation. However, the limitation of high inference latency becomes apparent in human-AI coordination applications, which requires real-time responses and high-frequency interactions, such as video games~\cite{wang2023describe,wang2023voyager, park2023generative}. 

\begin{figure*}
    \vspace{-2mm}
    \centering
    \includegraphics[width=0.9\textwidth]{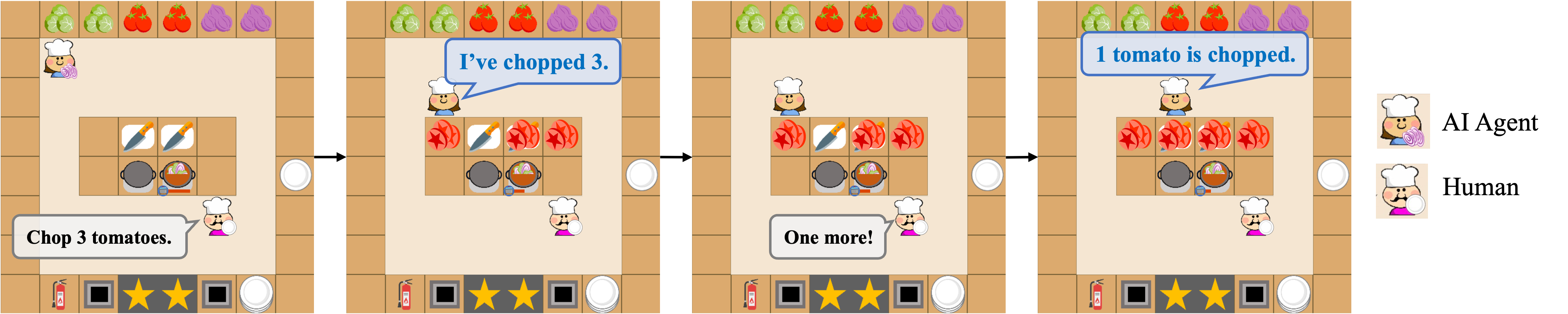}
    \vspace{-3.5mm}
    \caption{A concrete example of cooperation and communication between a human player and an AI player in Overcooked.}
    \label{fig:env_case}
\end{figure*}

In this work, we consider Overcooked as our real-time human-AI coordination testbed. Overcooked is a cooperative cooking game where players collaborate at a rate of approximately 3Hz to serve orders within a time budget. We further develop language-based communication in Overcooked to allow human-like cooperation. An ideal language agent is expected to exhibit real-time responsiveness, strong reasoning capabilities, and effective language-based communication with human players for the best performance in such a fast-paced game. Fig.~\ref{fig:env_case} shows a concrete example. When paired with an AI player, a human player might instruct the AI player by saying ``Chop 3 tomatoes.'' The AI player needs to accurately interpret and follow the command by swiftly picking up and chopping the specified number of tomatoes. Upon completion, the AI player needs to inform the human player of the ongoing progress, 
e.g., by saying ``I've chopped 3'', for future cooperation.
Afterward, the human player might directly say ``one more'', which itself looks semantically ambiguous. The AI player must correctly infer the true command from the history commands and promptly chop one more tomato. 

Traditional gaming AI usually employs smaller models or script policies, emphasizing fast inference for real-time responses to the game dynamics. Yet, this efficiency comes at the cost of limiting task completion and intra-player interaction abilities \cite{bajcsy2017learning,fisac2018probabilistically, yu2023learning,zhao2023maximum,strouse2021collaborating}.

We propose a \emph{\method} \emph{(HLA)} for real-time human-AI coordination in Overcooked. Inspired by System 1 and System 2 thinking~\cite{kahneman2011thinking}, HLA combines both robust reasoning and interaction capabilities from a large model and real-time inference from a smaller model and a reactive policy. 
In particular, HLA employs a hierarchical framework that consists of three modules: a proficient LLM for intention reasoning and language-based communication, a lightweight LLM for interpreting commands and high-level planning, and a script policy for executing low-level actions swiftly. We denote the proficient LLM as \emph{\slowmind}. The lightweight LLM, referred to as \emph{\fastmind}, generates macro actions, and the reactive script policy is referred to as \emph{\execmind}, which transforms macro actions into atomic actions. Human commands are processed simultaneously through both the {\slowmind} and {\fastmind} to enhance real-time performance. The reactive policy further ensures the feasibility of actions and high-frequency interactions. 

We consider three baseline agents, each lacking a specific HLA component. Our evaluation involves experiments on action latency, reasoning with simple and complex commands, and human studies. HLA demonstrates a remarkable advantage, being \textbf{an order of magnitude faster} than the best competitor in real-time action responsiveness. Furthermore, HLA outperforms the baseline agents significantly in command reasoning ability. Human studies confirm these findings, showing that HLA achieves approximately \textbf{50\% higher} game scores and receives the highest human preference.
More demonstrations can be seen on our website \url{https://sites.google.com/view/overcooked-hla/}.

\section{Related Works} 

\textbf{Language Agents.} Prior works train instruction-following agents with paired datasets of text and trajectories in games~\cite{meta2022human,xu2022grounded}, visual navigation~\cite{hermann2017grounded,wu2018building} and robotics~\cite{jiang2023vima,huang2023voxposer}. However, these works are limited to simple domains. Recently, with the advances of Large Language Models, a class of works start to utilize the strong reasoning power of LLMs to interact with complex domains including web scenarios~\cite{nakano2021webgpt, yao2022webshop, deng2023mind2web}, simulated environments~\cite{huang2022language,zhu2023ghost,wang2023describe,park2023generative}, real-world  environments~\cite{huang2022inner,vemprala2023chatgpt,zitkovich2023rt,ichter2022do}. These recent efforts use prompt engineering to elicit the power of LLMs~\cite{wei2022chain,yao2023tree}. Some recent works try to combine LLMs that play different functionalities in a cooperative manner~\cite{li2023camel,chen2023autoagents,chen2023agentverse,wang2023voyager}. In our work, we study the availability of language agents in response speed sensitive environments. 

\textbf{Human-AI Cooperation.} Building AI agents that can cooperate with humans is a longstanding challenge. Prior works study human-AI cooperation without communication in games such as Hanabi~\cite{hu2020other,cui2021k} and Overcooked~\cite{yu2023learning,zhao2023maximum,strouse2021collaborating}, and in robotics ~\cite{bajcsy2017learning,fisac2018probabilistically}. Language commands from humans are used to guide intelligent agents in visual navigation~\cite{hermann2017grounded,wu2018building} and robotics~\cite{jiang2023vima,huang2023voxposer}. Cicero~\cite{meta2022human} trains a language agent that can speak and make decisions like a human in the game of Diplomacy from a large volume of human play data. However, Diplomacy is a turn-based game and therefore has low real-time requirements. Recently, there are attempts in using LLM for human-AI interaction in domains including web scenarios~\cite{gur2023real,deng2023mind2web}, health~\cite{yang2023zhongjing,ali2020virtual}, games~\cite{wang2023describe,wang2023voyager} and other applications~\cite{gao2023assistgpt,schick2022peer,lu2023dialgen}. Similar to us, there are some concurrent works that use LLM for decision-making in the game of Overcooked~\cite{yan2023ask,zhang2023proagent,agashe2023evaluating}. We focus on improving real-time user experience in settings that demand rapid responses during human-AI interaction.

\textbf{LLM Inference Speedup.} The long inference latency of existing LLM-based agents is not desirable in domains that demand a fast response speed. ``Skeleton-Of-Thought''~\cite{ning2023skeleton} generates responses with a skeleton structure and uses parallel inference to reduce latency. Model compression methods achieve faster inference by distilling the knowledge of LLMs into smaller language models~\cite{dong2022survey,wei2022chain,wang2023large,wang2022self,brooks2023instructpix2pix} and performing quantization~\cite{nagel2020up,liu2021post,fang2020post}. In our work, we adopt a hierarchical design that leverages LLMs for reasoning and language interaction and small models for fast reaction in real-time gaming.
\section{Testbed: The Overcooked Game}
\label{sec:testbed}

\begin{figure*}[ht!]
\vspace{-2.5mm}
\centering
    \begin{subfigure}[t]{0.20\textwidth}
        \includegraphics[width=0.9\textwidth]{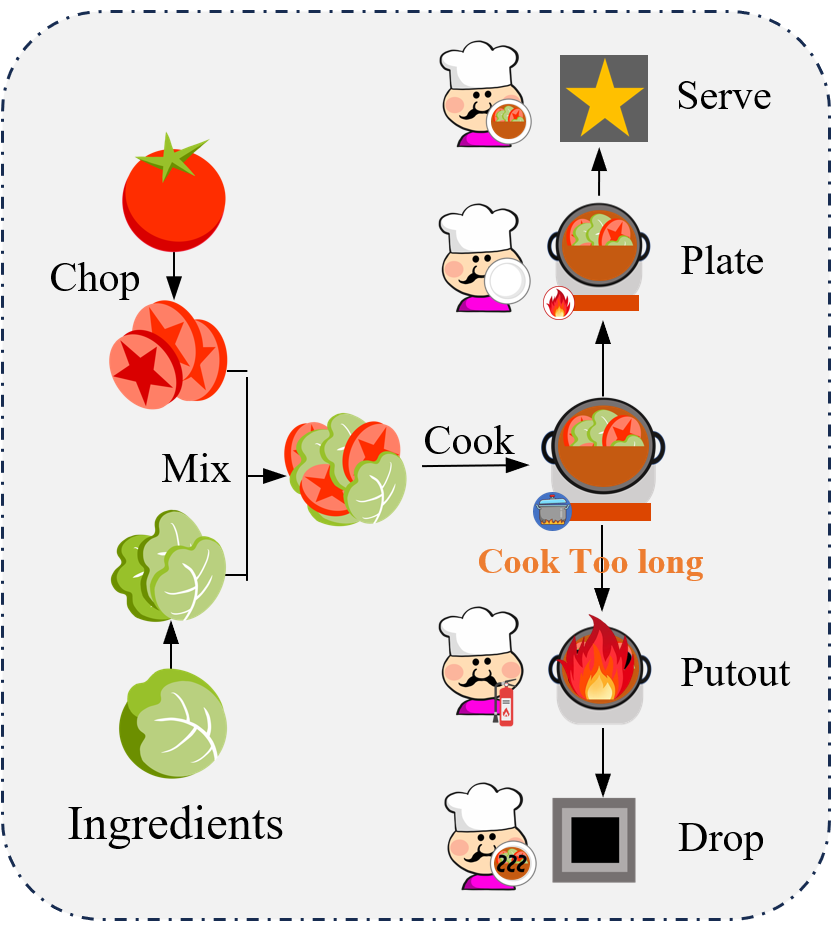}
        \vspace{-2mm}
        \caption{Cooking process.}
        \label{fig:env_process}
    \end{subfigure}
    \begin{subfigure}[t]{0.70\textwidth}
        \includegraphics[width=0.95\textwidth]{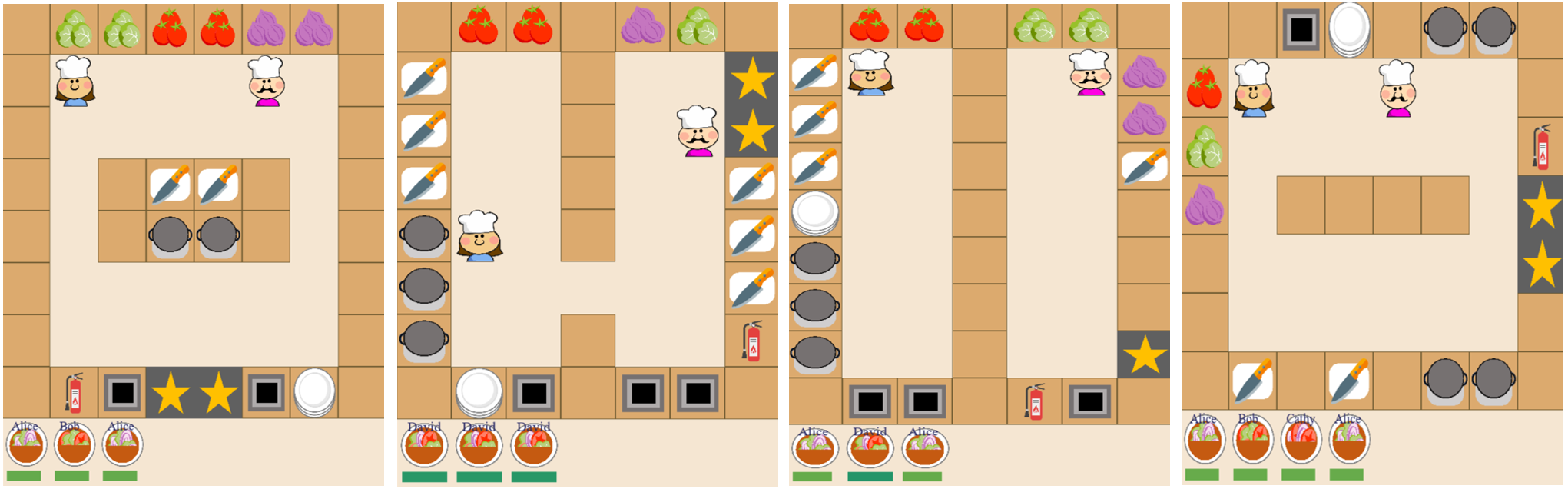}
        \vspace{-2mm}
        \caption{Designed maps. From left to right are {\mapring}, {\mapbot},{\mappart} and {\maphard}}
        \label{fig:exp_map}
    \end{subfigure}
\vspace{-3mm}
\caption{The cooking process and the maps in the Overcooked testbed.}
\vspace{-2mm}
\end{figure*}

\subsection{Environment Details}
Overcooked is a cooperative cooking game where participants must work together to prepare, cook, and promptly serve a variety of dishes. 
The Overcooked environment \cite{toomanycooks, NEURIPS2022_4f2accaf} simulates and simplifies the original Overcooked game and provides an RL training interface as a common testbed for real-time human-AI coordination.  
Throughout the game, orders continuously appear, each accompanied by a strict deadline. Players must prepare dishes in accordance with these orders and ensure they are delivered on time for rewards. Failed deliveries incur penalties. 
To finish an order, players must follow a precise sequence of steps: retrieving the necessary ingredients, chopping them, mixing them as per the recipe, cooking them to make a dish, plating the dish before it overcooks, and finally, serving it at the counter. Fig.~\ref{fig:env_process} depicts the cooking process of the game. Each player can perform one of the 4 actions at each time step, i.e., \emph{up, down, left, and right}, to control the character's position and to interact with other objects.

We implemented a collection of enhancements to the environment with full details described in Appendix~\ref{app:env}.
In particular, we further extend the original environment by developing a chat interface to allow natural language communication between human and AI players.
Human players can either pause the game and send text messages to the AI through an additional chat window or directly speak to the AI player during gameplay. Similarly, the AI player can respond to human players through either text or speech.

We also design 4 distinct maps as shown in Fig. \ref{fig:exp_map}.
In each game, a human player (the pink beard character) collaborates with an AI player (the blue character) to complete the orders. The first two maps, {\mapring} and {\mapbot}, assess general human-AI cooperation capabilities. The third map, {\mappart}, completely separates the two players, therefore they must cooperate to finish any order.
In the fourth map, {\maphard}, we raise the number of concurrent orders and accelerate the action frequency to intensify the gameplay.

\subsection{Challenges of Overcooked}
\label{sec: challenges}

\textbf{Real-time Cooperation. }
The Ovecooked game is particularly time-sensitive. It requires all players to adjust their game strategy in time and take swift actions, so that the orders do not timeout, and the dishes are not overcooked. 
The real-time requirement of the game indicates that, if an AI agent wants to leverage LLM's strong capabilities, it cannot adopt the conventional approach of directly prompting the LLM to generate its next action. This approach leads to substantial latency, making it impractical for this game.

\textbf{Command Reasoning.}
When playing the original Overcooked game, human players often give commands that are semantically ambiguous or overly complex. We collect some common human commands and present the typical scenarios as follows. 

\vspace{-1mm}
\begin{itemize}
    \item Quantity specification. The player asks others to ``chop 3 tomatoes.''
    \item Semantic analysis. Instead of giving direct commands such as ``cook the soup,'' the player gives hints, ``the soup order is about to timeout.''
    \item Ambiguous reference. The player first asks others to ``chop some tomatoes,'' then says ``one more.'' Here the ``one'' implicitly refers to a chopped tomato. 
\end{itemize}
\vspace{-1mm}

In such scenarios, the AI agent must consider various factors, such as human commands, environment composition, and historical actions, to determine the true intentions of the human player and respond accordingly. A basic reactive agent lacking reasoning capabilities will struggle to comprehend human commands, let alone collaborate effectively with humans to complete the game.

\section{Method}
\label{method}

\subsection{Overview}

\begin{figure*}
    \centering
    \includegraphics[width=0.65\textwidth]{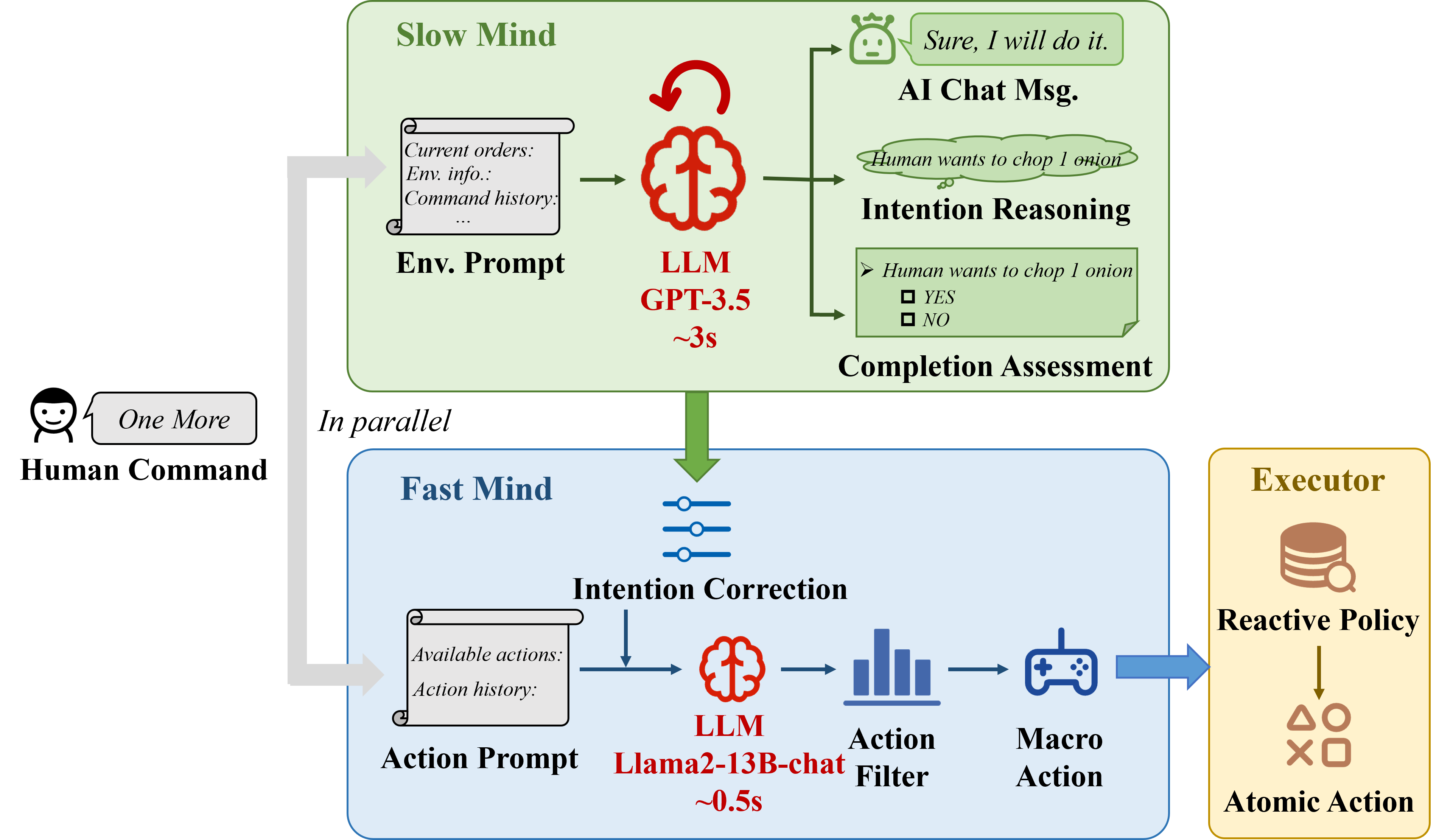}
    \vspace{-3mm}
    \caption{Framework of {\method}, including a {\slowmind} for intention reasoning and language interaction, a {\fastmind} for macro actions generation, and an {\execmind} to execute atomic actions.}
    \label{fig:framework}
    \vspace{-2.5mm}
\end{figure*}

As described in Sec.~\ref{sec:testbed}, the AI agent needs real-time responsiveness, command reasoning ability, and bilateral communication capability. A key observation in our testbed is that the AI agent can perform reasoning about human commands and communicate with humans at a regular frequency while atomic actions should be generated at a high frequency. Meanwhile, LLM-based agents are better at generating high-level moves than producing atomic actions. Therefore we propose \emph{\method} \emph{(HLA)} as depicted in Fig.~\ref{fig:framework}. HLA consists of three components: a proficient LLM, i.e., \emph{\slowmind}, that interprets human commands from the full game history and generates chat feedback; a lightweight LLM, i.e., {\fastmind}, that delivers high-level moves, which we call ``macro actions'', at a medium frequency; and a reactive policy, i.e., \emph{\execmind}, that is implemented as pre-defined scripts and transforms macro actions into atomic actions to interact with the environment at a high frequency. 
We will describe {\slowmind}, {\fastmind}, and {\execmind} in the following sections respectively.

\subsection{{\slowmind}}
\label{slowmind}

\begin{figure*}
    \centering
    \includegraphics[width=0.87\textwidth]{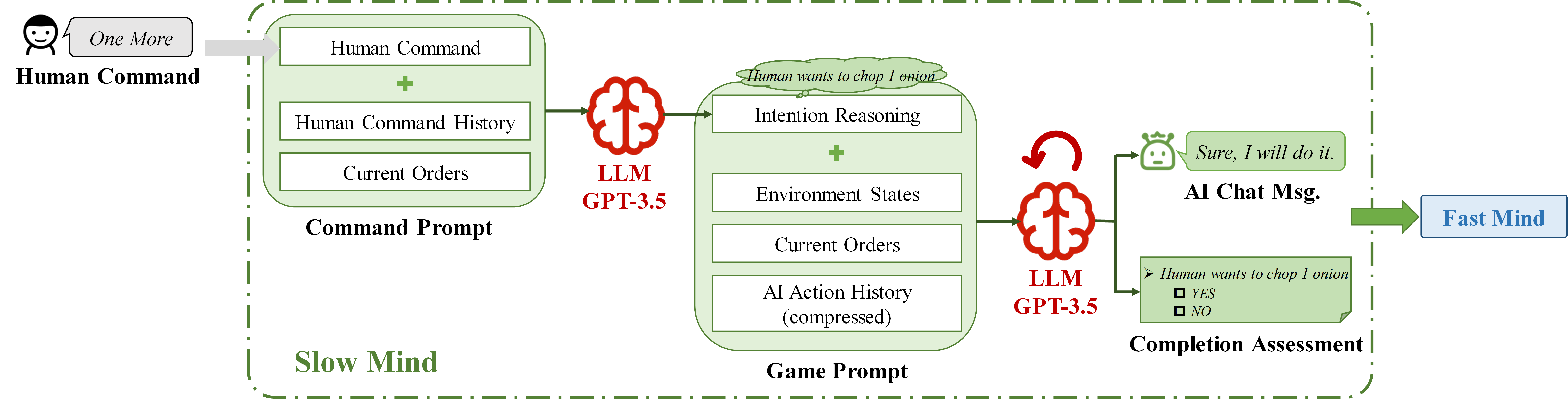}
    \vspace{-3mm}
    \caption{Workflow of {\slowmind}. {\slowmind} employs a two-stage design. It reasons human intention according to human commands in the first stage, then generates chat message and performs completion assessment periodically in the second stage.}
    \label{fig:slowmind}
    \vspace{-2.5mm}
\end{figure*}

{\slowmind} is empowered by a proficient LLM. The main functionality of {\slowmind} is to interpret vague human commands into concrete intentions, keep track of command completion progress, and provide responses containing key information and useful suggestions to the human partner. Also, {\slowmind} sends the inferred human intention to {\fastmind} and indicates to {\fastmind} whether the command is successfully completed.

To this end, we devise {\slowmind} to have two stages. The first stage, {\intstage}, infers human intention given the command and command history when the human issues a new command. The second stage, {\assestage}, periodically checks command completion and generates reply messages to the human partner based on the inferred intention. We use GPT-3.5~\cite{openai2022chatgpt35} here as GPT-3.5 features strong reasoning and communication power.

\subsubsection{\emph{{\intstage}}}

To interpret human commands, the LLM is provided with the command history and environmental information. The LLM interprets the vague command into concrete intentions that best reflect needs of the human partner. For instance, when the current soup orders are ``Alice Soup'' and ``David Soup'', the command ``Cook the first soup on the orders'' should be interpreted as ``Cook Alice Soup''. We note that this intention reasoning stage is carried out only once when a command is issued. The inferred intention is then stored by {\slowmind} and sent to {\fastmind}. The prompt used in {\intstage} is shown in Fig.~\ref{fig:fastmind-intstage}.

\begin{center}
\begin{tcolorbox}[width=80mm, colback=gray!5!white, colframe=gray!150!black, left=1mm, top=0.5mm, right=1mm, bottom=0.5mm]
    \textbf{Input:} \\
    You are in a virtual environment where ... \\
    Current orders: Alice Soup with plenty of time, ... \\
    Human's message: ``Cook the first soup'' \\
    Human's last intention: ...
    \vspace{-0.2em}
    \tcblower
    \vspace{-0.2em}
    \textbf{Output:} \\
    Human's intention now: \underline{Cook Alice Soup}
\end{tcolorbox}
\begin{minipage}{0.5\textwidth}
    \vspace{-2.5em}
    \captionof{figure}{{\slowmind} prompt in {\intstage}.}
    \label{fig:fastmind-intstage}
\end{minipage}
\end{center}

\subsubsection{\emph{{\assestage}}}

In {\assestage}, the LLM generates chat messages to the human partner and performs completion assessment. The chat messages cover different aspects of information including consent to the commands, planning of the AI agent, and information about the environment. The completion assessment is used as a tool to keep track of command completion progress. Once the human command is judged as completed, the inferred intention will be cleared and {\fastmind} will be reminded of completion of the command.

The LLM takes as input the inferred intention, current orders, environment state, and action history. We note that we use a compact representation for the action history by expressing consecutive repeated actions in the form of \emph{``action $\times$ repetition times''} to reduce the length of the action history.

Different from {\intstage} that is only executed once when a new command arrives, {\assestage} runs periodically to actively monitor the command completion progress and provide positive messages to the human partner. Notably, {\assestage} runs in different modes according to whether there exists an unaccomplished human command. When a human command is not completed yet, {\assestage} runs at full functionality. When there are no ongoing human commands, {\slowmind} switches to a more casual mode that only generates chat messages and does not perform completion assessment. 

Outline of the prompt used in {\assestage} is shown in Fig.~\ref{fig:fastmind-assestage}. We note that, when there are no ongoing human commands, completion assessment is disabled and the human's intention is ommited from the input prompt. 

\begin{center}
\begin{tcolorbox}[width=80mm, colback=gray!5!white, colframe=gray!150!black, left=1mm, top=0.5mm, right=1mm, bottom=0.5mm]
    \textbf{Input:} \\
    You are in a virtual environment where ... \\
    Current orders: Alice Soup (plenty of time), ... \\
    Environment state: ... \\
    Human's intention is: Cook Alice Soup once \\
    Action you've done: Chop Lettuce twice, ...
    \vspace{-0.2em}
    \tcblower
    \vspace{-0.2em}
    \textbf{Output:} \\
    Reasoning: I haven't cook Alice Soup yet. \\
    My chat message is: \underline{Sure, I will cook it for you.} \\
    Intention satisfied? \underline{No} 
\end{tcolorbox}
\begin{minipage}{0.5\textwidth}
    \vspace{-2.5em}
    \captionof{figure}{{\slowmind} prompt in {\assestage}.}
    \label{fig:fastmind-assestage}
\end{minipage}
\end{center}


\subsection{{\fastmind}}
\label{fastmind}

\begin{figure*}
    \centering
    \includegraphics[width=0.95\textwidth]{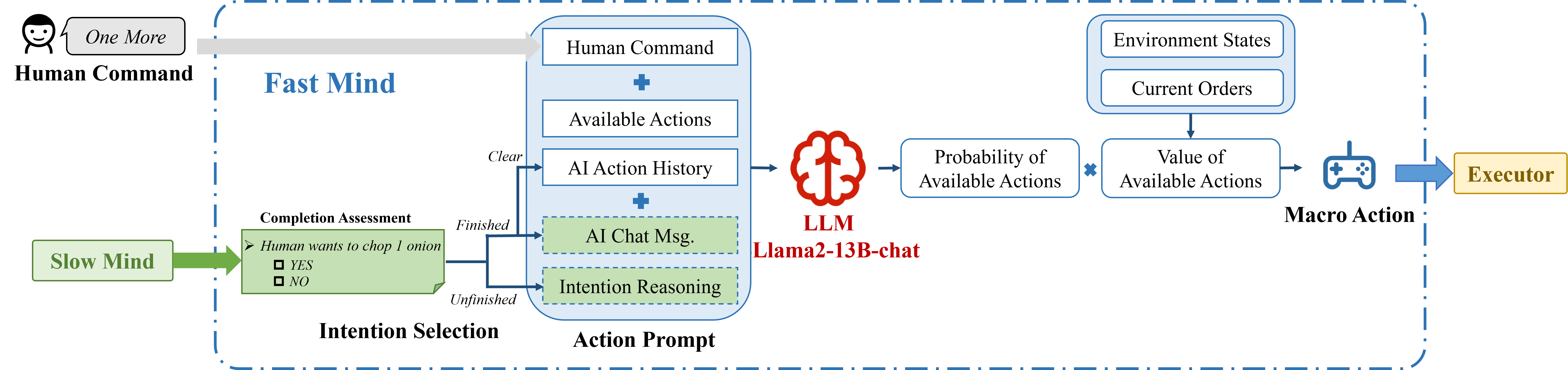}
    \vspace{-3mm}
    \caption{Workflow of {\fastmind}. {\fastmind} is empowered by a lightweight LLM. It works with {\slowmind} cooperatively with a conditional prompt mechanism and avoids sub-optimal moves with an action-filtering mechanism.}
    \label{fig:fastmind}
    \vspace{-3mm}
\end{figure*}

As argued in Sec.~\ref{method}, {\fastmind} produces high-level moves at a medium frequency while following human commands. We call these high-level moves macro actions. The current macro action set in our testbed includes chopping vegetables, serving soup, putting out the fire, and so on. A detailed description of macro actions can be found in Sec.~\ref{execmind}.

Despite having superior decision-making capabilities, a powerful LLM as used in {\slowmind} is not applicable in {\fastmind} due to huge inference latency. Therefore a lightweight LLM is more suitable. On the other hand, while a lightweight LLM such as Llama2-13B-chat can successfully follow a human command when the command is concise and clear, it often generates sub-optimal moves that lead to a lower score when there are no ongoing commands and violates the command when the human partner gives a vague command. 

Taking these factors into account, we design {\fastmind} that generates macro actions to interact with the environment, as depicted in Fig.~\ref{fig:fastmind}. {\fastmind} is empowered by a lightweight LLM (quantized version \cite{thebloke2023llama2gptq} of Llama2-13B-chat \cite{touvron2023LLaMA}). To better ground human commands into moves, {\fastmind} works with {\slowmind} cooperatively with a conditional prompt mechanism. Lastly, {\fastmind} avoids sub-optimal moves with an action-filtering mechanism.

\begin{center}
\begin{tcolorbox}[width=80mm, colback=gray!5!white, colframe=gray!150!black, left=1mm, top=0.5mm, right=1mm, bottom=0.5mm]
    \textbf{Input:} \\
    You are in a virtual environment where ... \\
    \textcolor{orange}{[If intention is unclear]} Human sends a message: ... \\
    \textcolor{orange}{[If intention is reasoned]} Human's intention is: ... \\
    \textcolor{orange}{[If intention is satisfied]} Your chat message is: ...
    \vspace{-0.2em}
    \tcblower
    \vspace{-0.2em}
    \textbf{Output:} \\
    My actions are: Chop Lettuce, \underline{<next action>}
\end{tcolorbox}
\begin{minipage}{0.5\textwidth}
    \vspace{-2.5em}
    \captionof{figure}{{\fastmind} prompt.}
    \label{fig:fastmind-prompt}
\end{minipage}
\end{center}

\subsubsection{Intention Selection}

To better align macro action generation with vague human commands, {\fastmind} also uses the inferred intention from {\slowmind} as input. To prevent macro action generation from being blocked by the intention reasoning stage of {\slowmind}, {\fastmind} uses the raw human command as input before {\slowmind} successfully infers the intention. This asynchronous execution nature results in the conditional prompt mechanism used by {\fastmind} as depicted in Fig.~\ref{fig:fastmind-prompt}. 
When {\slowmind} is analyzing the command, {\fastmind} uses as input the raw command, action history, and availability of actions. As soon as {\slowmind} completes {\intstage}, the inferred intention is sent to {\fastmind}, and {\fastmind} switches to use the inferred intention instead of the raw command to generate macro actions. Later when the command is evaluated as completed by {\slowmind}, the conditional input for {\fastmind} is switched to the chat messages produced by {\slowmind}, which helps the agent to better align its action with conversational response.

\subsubsection{Macro Action Generation} 

We obtain action probability by the output probability of each macro action conditioned on the prompt prefix. Fig.~\ref{fig:fastmind-prompt} illustrates the structure of the employed prompt. Each potential macro action candidate fills in the \underline{$<$next action$>$} variable to evaluate its output probability. The action history contained in the output prompt is to remind the agent of its past actions.

Lightweight LLM used in {\fastmind} may result in sub-optimal actions when managing complicated tasks. We therefore employ an action filter to filter out sub-optimal macro actions. The selection probabilities are calculated based on the probabilities output by the language model and the task-relevant value of each available macro action according to Eq.~(\ref{eq:fastmind}).

\vspace{-3mm}
\begin{equation}
    \log U(a|s) \propto \log P_{LLM}(a|s) + \alpha V(a| s)
    \label{eq:fastmind}
\end{equation}
\vspace{-4mm}

Here, $U(a|s)$ represents the selection probability of macro action $a$ under current state $s$, $P_{LLM}(a|s)$ represents the probability of the language model outputting macro action $a$, $V(a| s)$ represents the value of macro action $a$ contributing to the task reward. The values for $V(a| s)$ are hardcoded, and their detailed information can be found in the Appendix \ref{script}. $\alpha$ is a dynamic adjustment term and is smaller when the human command is not assessed to be met and is larger when it is. Finally, we adopt a greedy selection scheme, i.e. macro action with the greatest $U(a|s)$ is selected.

\subsection{{\execmind}}
\label{execmind}

At the lowest level of HLA, {\execmind} employs a script policy to convert macro actions generated by {\fastmind} into atomic actions to interact with the environment. 
The established macro action set includes the following types.

\begin{itemize} 
    \item \emph{Chop:} Transform an ingredient into its chopped form. 
    \item \emph{Mix:} Combine chopped ingredients for cooking. 
    \item \emph{Cook:} Utilize mixed ingredients to cook a soup. 
    \item \emph{Plate:} Transfer a ready soup to a plate. 
    \item \emph{Serve:} Deliver a plated soup to the delivery point. 
    \item \emph{Putout:} Extinguish a fire on a pot using an extinguisher. 
    \item \emph{Drop:} Plate charred soup and discard it into a bin. 
\end{itemize}
\vspace{-1mm}
Most macro actions are equipped with a specified target, such as \emph{``Chop Lettuce''} and \emph{``Cook Bob Soup''}, consequently resulting in a total set of $21$ macro actions. Implementation details of the reactive policy can be found in Appendix~\ref{script}. Given a macro action as a high-level goal, {\execmind} chooses the most appropriate move and performs path planning to the target. We empirically found this to be particularly beneficial as shown in Sec.~\ref{sec:exp-abl-simple}.

Note that the reactive policy of {\execmind} shares similarities with goal-conditioned reinforcement learning \cite{eysenbach2022contrastive, ma2022offline}, which can, in turn, be trained by typical reinforcement learning algorithms \cite{schulman2017proximal}. We remark this as a topic for future studies.


\section{Experiment}

In this section, we consider three baseline agents to verify the hierarchical structure of HLA. We first test the real-time responsiveness of HLA and baselines by measuring action response latency. Moreover, we use a simple command set to evaluate the cooperative ability of each agent and a more complex command set to test command reasoning ability. Finally, we conduct human studies to collect the game scores and the human preference of all agents. 

\subsection{Baseline}
\label{sec:exp-baseline}

To validate the effectiveness of the proposed hierarchical framework, we introduce three baseline agents, each lacking a certain component of the original HLA. 
\vspace{-1mm}
\begin{itemize}
    \item \emph{{\purelarge} (SMOA)}. We remove the {\fastmind} and let the {\slowmind} produce macro actions. Current available macro actions are added to the input of the {\slowmind}. The action filter is disregarded as GPT-3.5 is incapable of evaluating the probability for each macro action.
    
    \item \emph{{\puresmall} (FMOA)}. We remove the {\slowmind}, including both {\intstage} and {\assestage}. Due to the absence of {\assestage}, the dynamic adjustment term $\alpha$ in held static in Eq.~(\ref{eq:fastmind}). We set a maximum length for action history, beyond which the human intention is assumed as fulfilled. History of human commands, current orders, and environment states are incorporated additionally into the input of {\fastmind}. 

    \item \emph{{\purellm} (NEA)}. We remove the {\execmind} and let the {\fastmind} choose atomic actions to control the agent directly. In addition to the original input, the {\fastmind} of the NEA incorporates environmental state information, such as the positions of all items on the map. 
\end{itemize} 
\vspace{-1mm}
Details of the baseline agents can be found in Appendix~\ref{app:baseline}.

\subsection{Latency}
\label{sec:exp-abl-latency}

We use \emph{macro action latency} and \emph{atomic action latency} to measure the real-time responsiveness of HLA and baseline agents. The macro action latency is defined as the time interval between receiving a human command and subsequently generating a macro action. And the atomic action latency is, in turn, the latency of an atomic action. In order to simulate the natural human-AI coordination, we build a command set and issue each command to the AI agent to evaluate its response latency. Details about latency measurement method and the command set can be found in Appendix \ref{app:abl-latency}.

The results of macro action latency and atomic action latency are depicted in Tab. \ref{tab:exp-abl3_latency}, where lower number implies a faster action response time. The macro action latency of NEA is marked as ``/'' since it generates atomic actions directly. In HLA, when a human command is received, it is concurrently dispatched to both the {\slowmind} and the {\fastmind} therefore we report the macro action generation time of the {\fastmind}. 

Regarding the macro action latency, HLA produces $74.3\%$ lower latency than SMOA and $53.5\%$ lower latency than FMOA, suggesting that the hierarchical design significantly contributes to reducing response latency. FMOA exhibits a marginally shorter response time than SMOA, which can be attributed to the shorter inference time of the lightweight LLM, as well as the elimination of the intention completion assessment module in {\slowmind}. Regarding the atomic action latency, HLA exhibits an order of magnitude advantage over the best competitor (0.28 vs. 0.08), demonstrating the real-time responsiveness of HLA. Among all agents, NEA performs the highest atomic action latency, indicating the importance of {\execmind} that interacts with the environment with high frequency.

\begin{table}[ht]
\vspace{-1mm}
\centering
\begin{tabular}{ccccc}
    \toprule
    & NEA & SMOA & FMOA  & HLA  \\
    \midrule
    Mac. Act. Latency(s) & / & 4.16 \scriptsize{(1.01)} & 2.30 \scriptsize{(1.81)}  & \textbf{1.07 \scriptsize{(0.22)}} \\
    Ato. Act. Latency(s) & 0.71 \scriptsize{(0.08)} & 0.61 \scriptsize{(0.65)} & 0.28 \scriptsize{(0.31)}  & \textbf{0.08 \scriptsize{(0.06)}}\\
    \bottomrule
\end{tabular}
\centering
\caption{Macro action latency and atomic action latency of different agents. The format is ``mean (standard deviation)''.}
\label{tab:exp-abl3_latency}
\vspace{-5mm}
\end{table}

\subsection{Performances with Simple Commands}
\label{sec:exp-abl-simple}

An ideal AI agent should cooperate seamlessly with human players to achieve high game score, either in the absence of explicit human commands or after human commands is fulfilled.
Thus, we design two test cases where the human player either remains silent or speaks very little. 
In both cases, the human player only chops ingredients and does not perform other tasks. 

\begin{figure}[ht]
    \centering
    \includegraphics[width=0.75\linewidth]{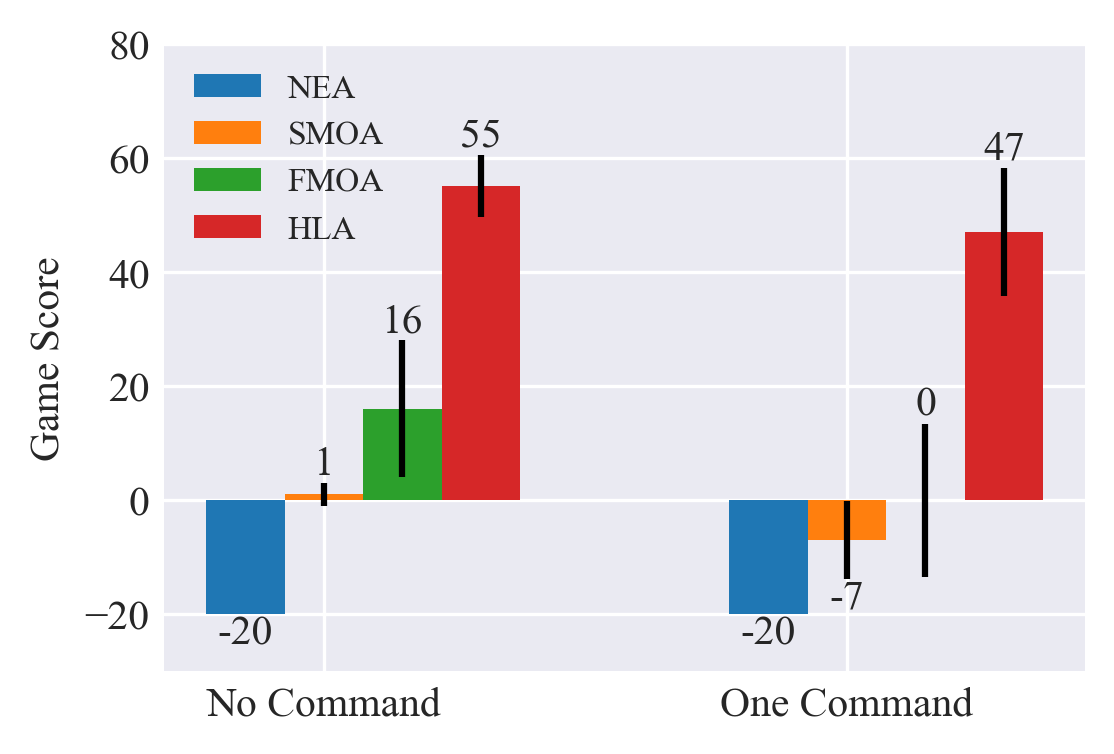}
    \vspace{-4.5mm}
    \caption{Average game scores of HLA and baseline agents. Black line denotes standard deviation.}
    \label{fig:exp-abl1_reward}
    \vspace{-4mm}
\end{figure}

\vspace{-1mm}
\begin{itemize}
\item \emph{No Command}: The human player does not issue any command throughout the game. 

\item \emph{One Command}: The human player asks the AI agent to prepare a specific soup at the start of the game, and the order for this soup only appears once at the start of the game. 
\end{itemize}
\vspace{-1mm}
We employ game score as a metric to assess the cooperative ability of HLA and baseline agents. The game score ascends every time an order is finished in time and descends whenever an order expires. We test both \emph{No Command} and \emph{One Command} on Map {\maphard} and replicate the experiment five times with the same order sequence and keep the same experiment conditions. The average game scores under different test cases are depicted in Fig.~\ref{fig:exp-abl1_reward}, where the black line denotes standard deviation. Visualization of the gaming process can be found in Appendix~\ref{app:abl-score}. 

NEA fails to complete basic tasks, such as picking up things or approaching the target, and frequently gets stuck, therefore misses all soup orders and achieves minimum score of $-20$. This underscores the vital role played by the {\execmind}, which translates macro actions into atomic actions. 
SMOA adequately follows the human command in the \emph{One Command} case, but often overcooks dishes due to its high response latency, thus also performs poorly. 
In the \emph{No Command} case, FMOA serves the desired soups successfully but underperforms compared to HLA, primarily due to a high latency. However, in the \emph{One Command} case, FMOA keeps making the requested soup even if it does not appear on the order list. In other words, FMOA fails to confirm the completion of a human command leads to suboptimal actions that do not fulfill the orders.

\subsection{Interpreting Complex Commands}
\label{sec:exp-abl-complex}

To see how well the agents comprehend and respond to commands of varying complexity, we design a complex command set comprising the following three challenges outlined in Sec. \ref{sec: challenges}. 
\vspace{-1mm}
\begin{itemize}
    \item \emph{Quantity specification (Quantity).} We consider commands with specific numbers, e.g. ``Chop 3 Lettuce.'' or ``Cook Bob Soup once.''
    \item \emph{Semantic analysis (Semantics).} For example, the human gives a hint, ``Alice Soup is about to timeout!'' instead of a direct command like ``Cook Alice Soup.''
    \item \emph{Ambiguous reference (Ambiguity).} For example, after asking the agent to chop two onion, the human player asks, ``Chop 1 more.'' The AI agents need to infer the true intention based on environmental context and history commands.
\end{itemize}
\vspace{-1mm}
We generate 10 different commands for each challenge. The details of the complex command set can be found in Appendix~\ref{app:abl-complex}. 
Each individual command is tested for 5 times. A command is considered successful if it succeeds at least 3 out of 5 attempts within 60s. In such cases, the command is labeled as passed, and its completion time is calculated as the average time taken for successful attempts. Otherwise, it is marked as failed and its completion time is marked as the maximum time of 60s. We report the average success rate and average completion time of commands in each challenge subset.

Tab.~\ref{tab:exp-abl4_instruct1} shows the results of HLA and baseline agents. NEA fails to execute any complex command and thus produces the worst performance.
Except NEA, FMOA exhibits the lowest success rate and the highest completion time when handling ambiguous commands, highlighting the significance of intention reasoning through the {\slowmind}. SMOA performs poorly on quantity and semantics commands, displaying the longest completion time, indicating that although the {\slowmind} can interpret commands, it suffers from prolonged latency. HLA surpasses baseline agents with a notable advantage in both success rate and completion time, which demonstrates the effectiveness of the hierarchical design.

\begin{table}[ht]
\vspace{-2mm}
\centering
\begin{tabular}{c|cc|cc|cc}
    \toprule
    \multirow{2}{*}{AI Agents} & \multicolumn{2}{c|}{Quantity} & \multicolumn{2}{c|}{Semantics} & \multicolumn{2}{c}{Ambiguity} \\ 
    \cmidrule{2-7}
    & Suc.$\uparrow$ & Time$\downarrow$ & Suc.$\uparrow$ & Time$\downarrow$ & Suc.$\uparrow$ & Time$\downarrow$ \\
    \midrule
    NEA & 0.00 & 60.00 & 0.00 & 60.00 & 0.00 & 60.00 \\
    SMOA & 0.40 & 47.14 & 0.60 & 40.20 & \textbf{0.70} & 39.57 \\
    FMOA & 0.60 & 40.01 & 0.60 & 32.67 & 0.30 & 50.16 \\
    HLA & \textbf{1.00} & \textbf{13.30} & \textbf{0.90} & \textbf{17.13} & \textbf{0.70} & \textbf{27.78} \\
    \bottomrule
\end{tabular}
\centering
\caption{Success rate and completion time for complex commands of HLA and baseline agents.}
\label{tab:exp-abl4_instruct1}
\vspace{-5mm}
\end{table}

\textbf{Ablation study on the two-stage design of {\slowmind}.} 
The AI agent needs to infer human intentions and evaluate whether human commands are completed. In HLA, these tasks are performed separately by the two stages of {\slowmind}.
In this section, we consider two variants of {\slowmind} to validate our two-stage design on interpreting complex commands.
\vspace{-1mm}
\begin{itemize}
    \item \emph{HLA without Intention Reasoning (no IR)}. We remove the {\intstage} and use human commands directly as intentions for {\assestage}.
    \item \emph{HLA with one-stage {\slowmind} (one-stage)}. We combine the {\intstage} and the {\assestage} within {\slowmind}, which now infers human intention, chats, and assesses intention completion simultaneously. 
\end{itemize}
\vspace{-1mm}

The results are shown in Tab.~\ref{tab:exp-abl4_instruct2}. HLA with one-stage {\slowmind} exhibits the longest completion time and the lowest success rate. We hypothesize that this is due to {\slowmind} having to handle multiple tasks simultaneously, resulting in poorer performance and longer inference time. HLA without Intention Reasoning performs worse than HLA(full) across all challenge subsets, particularly on ambiguous commands, indicating the significance of incorporating the {\intstage}.
\begin{table}[ht]
\vspace{-2mm}
\centering
\begin{tabular}{c|cc|cc|cc}
    \toprule
    \multirow{2}{*}{HLA variants} & \multicolumn{2}{c|}{Quantity} & \multicolumn{2}{c|}{Semantics} & \multicolumn{2}{c}{Ambiguity} \\ 
    \cmidrule{2-7}
    & Suc.$\uparrow$ & Time$\downarrow$ & Suc.$\uparrow$ & Time$\downarrow$ & Suc.$\uparrow$ & Time$\downarrow$ \\
    \midrule
    no IR & 0.80 & 22.42 & 0.80 & 22.14 & 0.60 & 40.50 \\
    one-stage & 0.50 & 35.13 & 0.50 & 36.90 & 0.50 & 44.36 \\
    HLA(full) & \textbf{1.00} & \textbf{13.30} & \textbf{0.90} & \textbf{17.13} & \textbf{0.70} & \textbf{27.78} \\
    \bottomrule
\end{tabular}
\centering
\caption{Success rate and completion time for complex commands of HLA and its two variants.}
\label{tab:exp-abl4_instruct2}
\vspace{-5mm}
\end{table}

\subsection{Human Studies}
\label{sec:exp-hci}

\subsubsection{Experiment Setting}

We invite 60 volunteers for the human-AI experiment and divide them into 4 groups, each of which consists of 15 volunteers playing on a specific map. All volunteers are provided with a detailed introduction to the basic gameplay and the experiment process. They are fully aware of all their rights and experiments are approved with the permission of the department.

Considering the poor performance of NEA observed in Sec. \ref{sec:exp-abl-simple}, we only evaluate SMOA, FMOA, and HLA in this section. Each volunteer plays with the three AI players on the same map for two gaming phases, the preparation phase and the competition phase. 
During the preparation phase, volunteers are required to collaborate with each of the three AI players for at least one round. In this process, human players familiarize themselves with the environment and engage in casual interactions with the AI players to explore their behaviors.
In the competition phase, volunteers can only play one round of the game with each of the three AI players to achieve the highest possible score.
We record the game scores and ask the volunteers to rank the three AI players based on their gaming experience. 
We report the game scores and human preferences of SMOA, FMOA, and HLA in the following subsections.

\subsubsection{Game Score}
\label{sec:score}

The average game scores on various maps from the competition phase are presented in Tab.~\ref{tab:exp-hci_reward}. A high variance in scores can be attributed to the discrete nature of task rewards, where finishing an order yields $15$ to $20$ points.
SMOA exhibits the lowest scores on all maps due to its high response latency. Compared to SMOA, FMOA performs slightly better, averaging a 10-point advantage in scoring on each map. HLA outperforms both SMOA and FMOA across all maps with $\sim 50\%$ higher game scores, reflecting a significant advantage. In particular, HLA achieves a game score increase of over 40 points, i.e. serving at least two additional orders, on {\mappart} and {\maphard} when compared to baseline agents. 
This highlights HLA's effective collaboration ability and real-time responsiveness.

\begin{table}[ht]
\vspace{-2mm}
\centering
\begin{tabular}{ccccc}
    \toprule
    AI Agents	&	\textit{Ring}	&	\textit{Partition}	&	\textit{Bottleneck}	&	\textit{Quick} \\ \midrule
    SMOA	&	80.9 \scriptsize{(29.1)}	&	33.0 \scriptsize{(27.1)}	&	102.4 \scriptsize{(35.6)}	&	60.8 \scriptsize{(48.5)}\\
    FMOA	&	92.5 \scriptsize{(21.7)}	&	57.7 \scriptsize{(37.9)}	&	103.8 \scriptsize{(30.6)}	&	71.2 \scriptsize{(50.2)}\\
    HLA	&	\textbf{114.4 \scriptsize{(19.4)}}	&	\textbf{100.3 \scriptsize{(36.4)}}	&	\textbf{130.3 \scriptsize{(19.7)}}	&	\textbf{117.2 \scriptsize{(45.3)}}\\ \bottomrule
\end{tabular}
\centering
\caption{Average game score when paired with different AI players. Standard deviations are shown in parentheses.}
\label{tab:exp-hci_reward}
\vspace{-5mm}
\end{table}

\textbf{Behavior Analysis.}
\label{sec:exp-hci-use}
We additionally calculate the ratio of valuable macro actions performed by each AI player, as well as the frequency of fire accidents on all maps during the competition phase. 
A macro action is considered valuable if it produces a positive utility value upon execution. Conversely, useless macro actions may either be inherently invalid or become unexecutable due to high inference latency. We analyze three common macro actions, including \emph{Chop} ingredients,\emph{Cook} soups and \emph{Serve} orders. The results are documented in Tab.~\ref{tab:exp-hci_use}. HLA stands out by executing the highest quantity of valuable macro actions and minimizing fire accidents, significantly outperforming SMOA and FMOA. Notably, the ratio of \emph{Serve} macro action executed by HLA is 100\%, indicating not only its capability to serve the correct order but also to do so promptly. This further helps to explain the excellent performance of HLA with regard to reasoning and real-time responsiveness.

\begin{table}[ht]
\vspace{-2mm}
\centering
\begin{tabular}{cccc|c}
    \toprule
    AI Players & \textit{Chop $\uparrow$} & \textit{Cook $\uparrow$} & \textit{Serve $\uparrow$} & \textit{Fire $\downarrow$} \\ \midrule
    SMOA        & 0.569 & 0.755 & 0.273 & 0.118 \\
    FMOA        & 0.766 & 0.833 & 0.850 & 0.059 \\
    HLA         & \textbf{0.828} & \textbf{0.950} & \textbf{1.000} & \textbf{0.029}  \\ \bottomrule
\end{tabular}
\centering
\caption{The ratio of valuable actions performed by different AI players and the frequency of fire accidents on all maps.}
\label{tab:exp-hci_use}
\vspace{-5mm}
\end{table}

\subsubsection{Human Preference} 
\begin{figure}[ht]
\vspace{-2mm}
\centering
    \begin{subfigure}[t]{0.35\linewidth}
        \includegraphics[width=\textwidth]{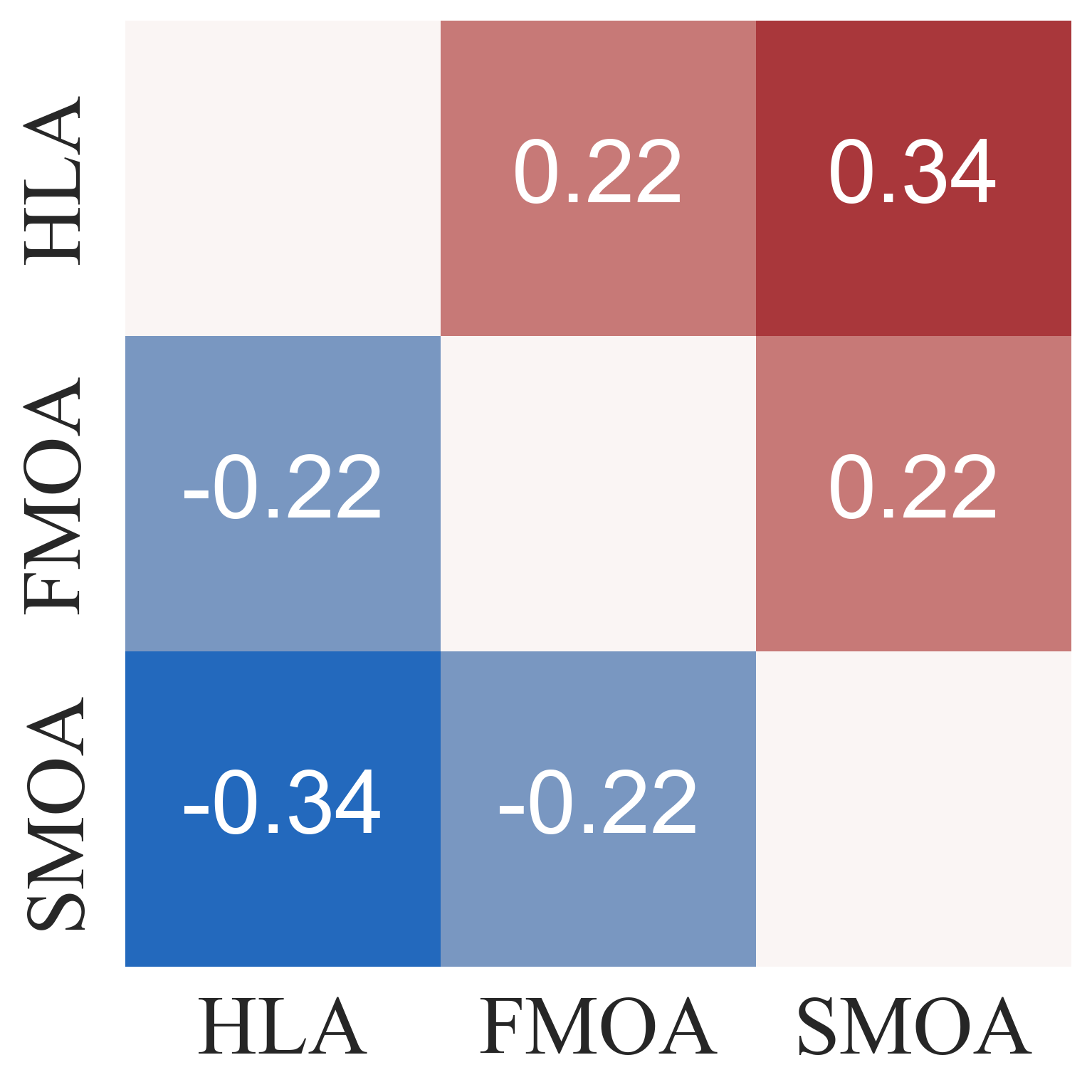}
        \vspace{-5.5mm}
        \caption{Comm. Accuracy.}
    \end{subfigure}
    \begin{subfigure}[t]{0.35\linewidth}
        \includegraphics[width=\textwidth]{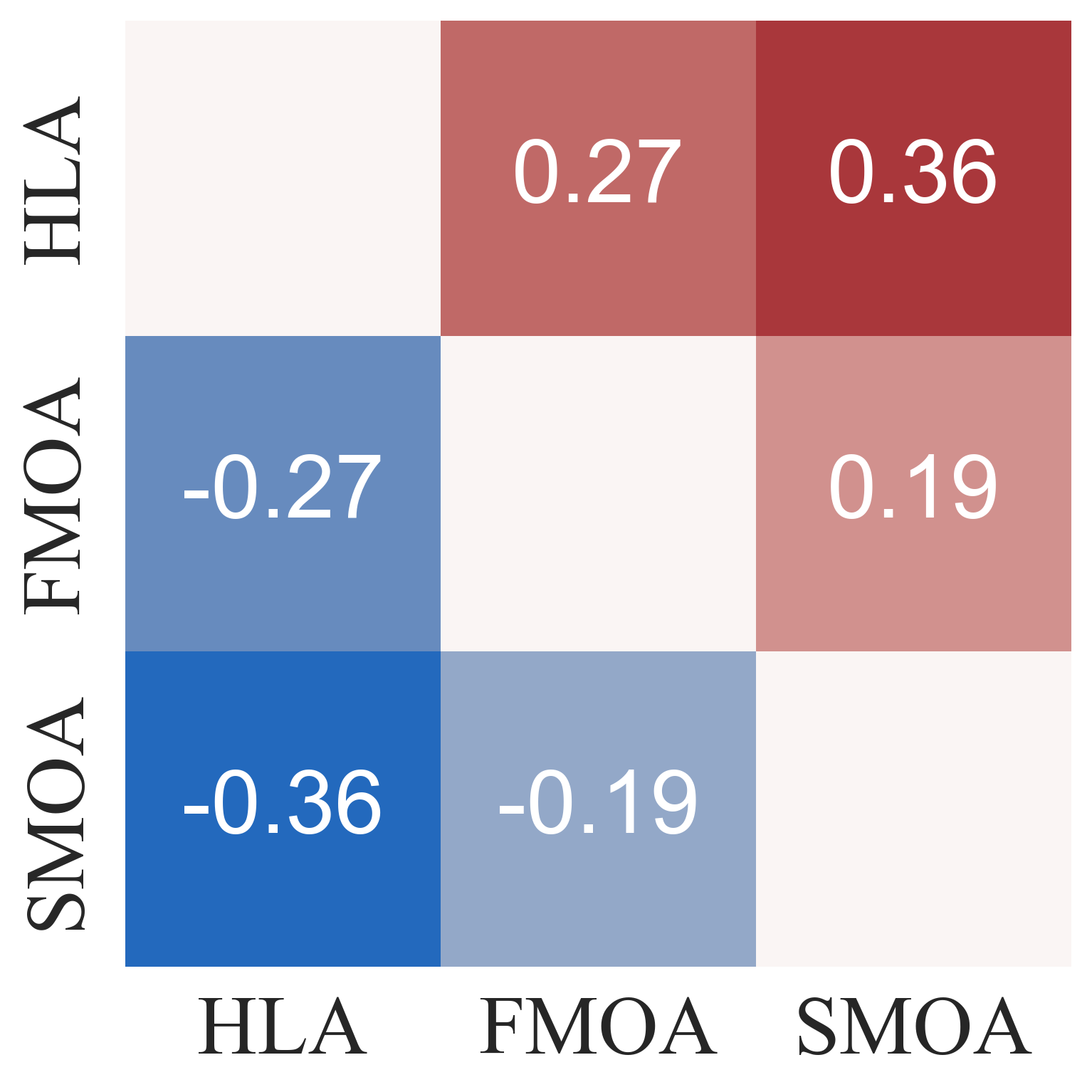}
        \vspace{-5.5mm}
        \caption{Comm. Consistency.}
    \end{subfigure}
\vspace{-3mm}
\caption{Human preference on communication accuracy, and consistency between chat message and actions. }
\vspace{-3mm}
\label{fig:exp-hci_chat}
\end{figure}

\begin{figure}[ht]
\vspace{-2mm}
\centering
    \begin{subfigure}[b]{0.35\linewidth}
        \includegraphics[width=\textwidth]{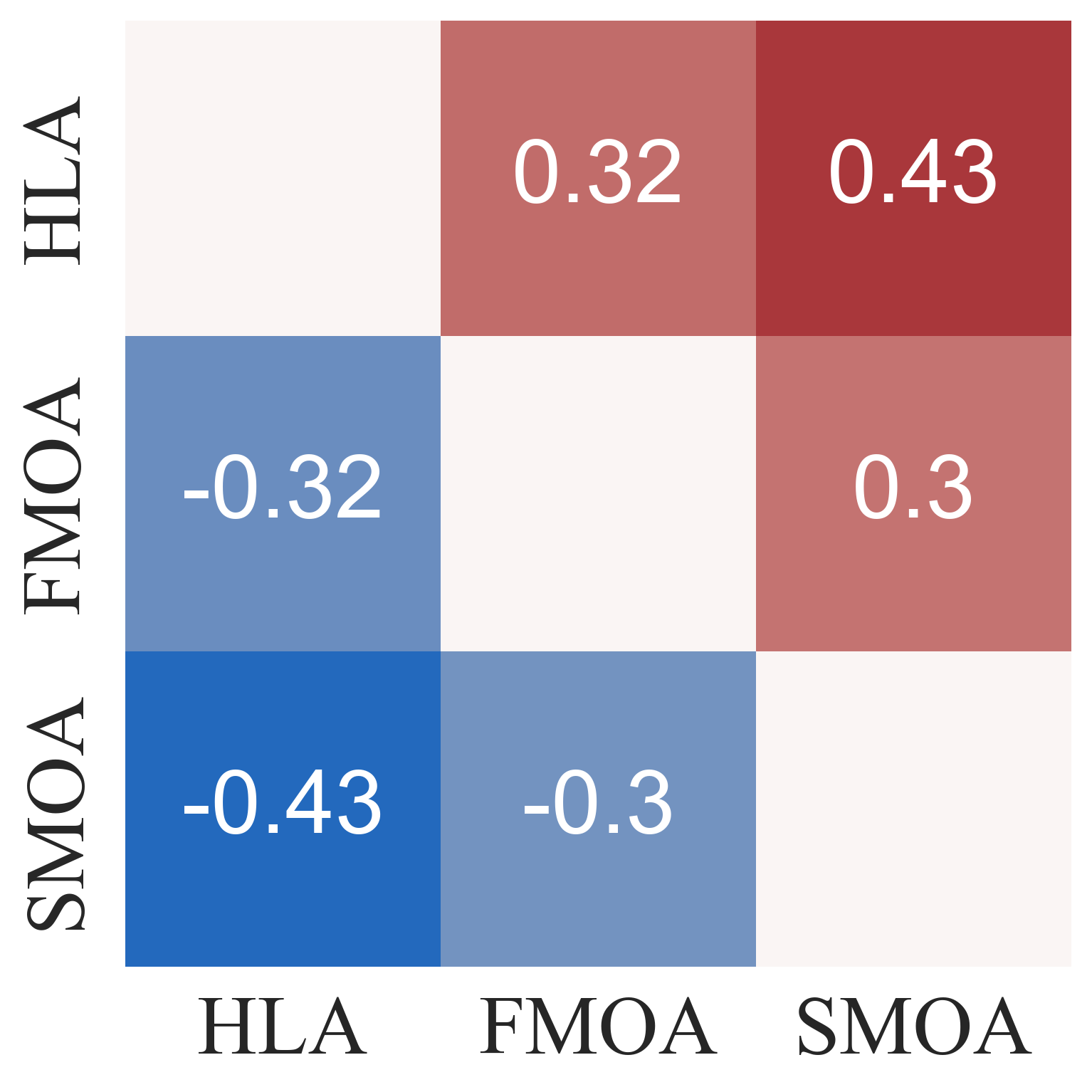}
        \vspace{-5.5mm}
        \caption{Preparation Phase}
    \end{subfigure}
    \begin{subfigure}[b]{0.35\linewidth}
        \includegraphics[width=\textwidth]{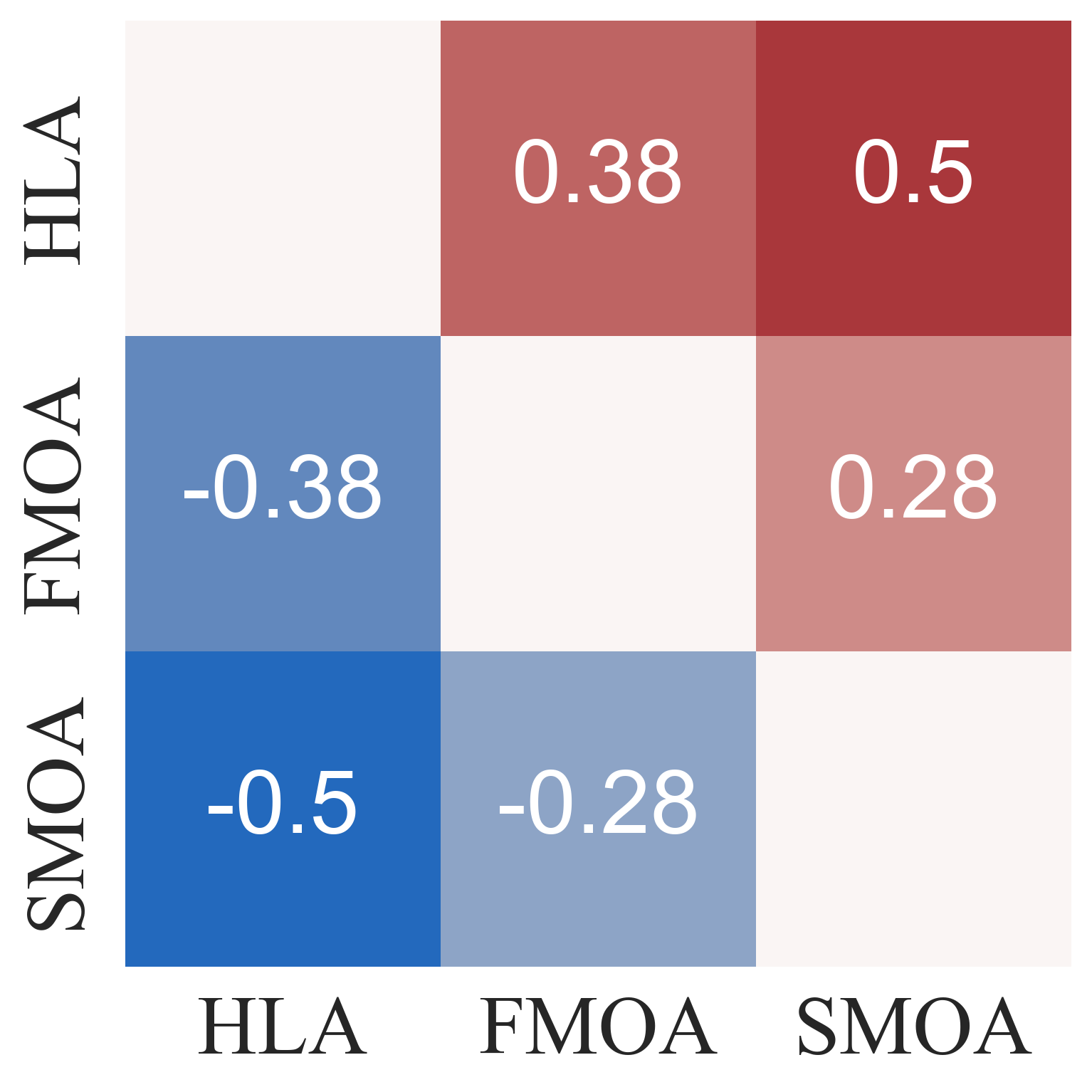}
        \vspace{-5.5mm}
        \caption{Competition Phase}
    \end{subfigure}
\vspace{-3mm}
\caption{Overall human preference on different AI players. }
\label{fig:exp-hci-fb}
\vspace{-3mm}
\end{figure}

In this section, we report the human preference on different AI players. Fig.~\ref{fig:exp-hci_chat} and Fig.~\ref{fig:exp-hci-fb} show the language communication preference and the overall preference of human participants respectively. Numbers indicate the difference of players who prefer row AI player over column AI player.

\textbf{Communication Preference. }
Fig.~\ref{fig:exp-hci_chat} illustrates human feedback on communication accuracy, as well as consistency between chat messages and actions. The data is derived from both the preparation and competition phases. More than 20\% of human players prefer HLA over SMOA and FMOA in terms of language communication accuracy as well as the consistency between chat messages and actions. We can also observe that FMOA outperforms SMOA with an advantage of $\sim$20\% human preference, primarily due to SMOA's high latency according to the collected feedback. This underscores the pivotal role of real-time responsiveness of effective language communication.

\textbf{Overall Preference. }
Fig.~\ref{fig:exp-hci-fb} reports the overall human preference of the preparation phase and the competition phase. Human participants expressed a strong preference for HLA over SMOA and FMOA, with a preference rate exceeding 30\%. This preference was particularly evident during the competition phase, where HLA was favored over SMOA by as much as 50\%. Human preference for FMOA remains higher than that for SMOA, at around 30\%, aligning with the conclusion of communication preferences. Overall preference is a comprehensive metric of how human players evaluate an AI player's actions and communication. The strong preference on HLA indicates that it exhibits superior cooperative skills, faster responsiveness, and more consistent language communication.

\section{Conclusion}

We propose {\method}, an AI agent that can cooperate with humans using natural language in environments that require real-time execution. 
Throughout the comprehensive experiments in an extended Overcooked, our method consistently excels in terms of game score, response latency, and human preference, showcasing reliable real-time human-AI cooperation. 
Our method could be improved in a few key areas: substituting GPT-3.5 with GPT-4 in the {\slowmind} for enhanced semantic analysis, and replacing the scripted executor with an automatic one developed through goal-conditioned reinforcement learning to streamline scripting and boost low-level execution performance.

\balance



\newpage

\begin{acks}
This research was supported by National Natural Science Foundation of China (No.62325405, U19B2019, M-0248), Tsinghua University Initiative Scientific Research Program, Tsinghua-Meituan Joint Institute for Digital Life, Beijing National Research Center for Information Science, Technology (BNRist), Beijing Innovation Center for Future Chips and 2030 Innovation Megaprojects of China (Programme on New Generation Artificial Intelligence) Grant No. 2021AAA0150000.
\end{acks}






\newpage
\appendix

\setcounter{figure}{0}
\setcounter{table}{0}

We would suggest to visit our website \url{https://sites.google.com/view/overcooked-hla/} for more information.

\section{Environment Details}
\label{app:env}
The original Overcooked benchmark \cite{toomanycooks} features a relatively straightforward menu and cooking process. In each gameplay session, a human player collaborates with an AI player to fulfill the orders. There is only one type of order, which is an onion soup made from three onions. Completing an order involves merely four steps, i.e., retrieving the necessary ingredients, mixing them in a pot to create a soup, plating it, and serving at the counter. Afterwards, the players receive a reward of 20.
Subsequent research expands upon this environment. For example, in \cite{NEURIPS2022_4f2accaf}, players must use a fire extinguisher to put out the randomly appearing fire.

To make the benchmarks more challenging, we incorporate mechanisms from the original Overcooked game and make the following extensions:
\begin{itemize}

    \item Diverse Ingredients. We offer three ingredients, i.e., onion, tomato, and lettuce. 
    
    \item Chopping Mechanism. All ingredients must be chopped on the chop board before they are mixed and put in the pot. Player needs to interacts with the chop board 8 times to chop the ingredient. 
    
    \item New Dishes. 
    We design four dishes: \emph{Alice Soup}, made of one onion and one lettuce; \emph{Bob Soup}, made of one tomato and one lettuce; \emph{Cathy Soup}, made of one tomato and one onion; and \emph{David Soup}, which includes all three ingredients. The cooking time for all soups is 15 seconds.

    \item Fire Mechanism. Leaving the cooked soup in pot for 25 seconds would overcook the soup and set the pot on fire. Players must use a fire extinguisher to put out the fire. The putout process takes 5 seconds. After the fire is put out, players can plate the overcooked food and discard it.

    \item Trash Can. Players can dispose of any unwanted ingredients or overcooked dishes in a trash can to free up space. 

    \item Order Timeout. During the game, soup orders appear randomly. Orders for Alice, Bob, and Cathy Soup are valid for 60 seconds, while orders for David Soup last 70 seconds.  If finished in time, Alice, Bob, and Cathy soups give a reward of 15, and David soups give a reward of 20. Otherwise, the order disappears and gives a negative reward of -5.

    \item Human-AI Chat Interface: To simulate natural human interactions in the original game, we add a chat interface for communication between the human player and the AI player. The human player can either pause the game and communicate with the AI using a chat window or speak directly during gameplay. The AI can respond via text or speech.
    
\end{itemize}

The action space of the extended Overcooked testbed is thus composed of two parts: the atomic actions and the texts for communication. Specifically, the atomic actions include only four elements, i.e., \emph{up, down, left, and right}, which control the character's position and its interaction with other objects. 

We design 4 distinct maps for the extended Overcooked testbed. In each game, a human player (the pink beard character) collaborates with an AI player (the blue character) to complete the orders. 
Whenever an order is fulfilled or timeouts, a new order appears, ensuring that there are 3 valid orders in the game at any given time.
Each game lasts for 100 seconds, and the default action frequency of the AI player is 2.5 Hz, resembling the natural action frequency of human players. 
The first two maps assess the general cooperation. {\mapring} employs a ring-like layout, while {\mapbot} creates a bottleneck that encourages the human player and the AI player to stay within a certain area. 
The third map, {\mappart}, separates the two players, necessitating task division and cooperative coordination. 
In the fourth map, {\maphard}, we raise the number of concurrent orders to 4 and the speed of the player to 3.5 Hz to intensify the gameplay.

To finish an order, players must follow a precise sequence of steps: retrieving the necessary ingredients, chopping them, assembling them as per the recipe, cooking them to make a dish, plating the dish before it is charred, and finally, serving it at the counter. If a soup cooks for a long time, it gets overcooked and the pot catches fire. Players must putout the fire with a fire extinguisher, plate the charred soup, and drop it into the trash bin.

\section{Method Details}

\subsection{Prompt Details}

\subsubsection{Prompt of {\slowmind} in {\intstage}}

As discussed in Sec.~\ref{slowmind}, in {\intstage}, HLA takes in the human's chat message and reasons the human intention. Here, we divide the chat message into 6 categories, including useless message, short-term request, complicated instruction, long-term request, questions, and others. If the human player asks a question, {\intstage} passes the question directly to {\assestage}, which then generates a chat response.

The full prompt is shown in Fig.~\ref{fig:app-prompt-HLA_intstage}. Mutable prompt components are marked in $<\,>$, such as $<Soup Orders>$.

\subsubsection{Prompt of {\slowmind} in {\assestage}}

As discussed in Sec.~\ref{slowmind}, {\slowmind} generates chat messages, whereas {\assestage} assesses intention completion. 
Depending on whether the human intention is satisfied, {\slowmind} works in different mode.
If the human intention is not satisfied, we ask the {\slowmind} to output its inner reasoning, intention completion assessment, and chat message. The full prompt of this mode is shown in Fig.~\ref{fig:app-prompt-HLA_assestage}. If the human intention is satisfied, we prompt the {\slowmind} to output chat message response only, as shown in Fig.~\ref{fig:app-prompt-HLA_assestage2}.

\subsubsection{\emph{\fastmind}}

As discussed in Sec.~\ref{fastmind}, the {\fastmind} uses a conditional prompt depending on whether the human intention is inferred and whether it is satisfied. 

The full prompt is shown in Fig.~\ref{fig:app-prompt-HLA_fastmind}, where the control conditions are marked in orange color. {\fastmind} takes human chat message, human intention reasoned by {\slowmind}, or chat message generated by {\slowmind} as input, depending on the condition.

\subsection{Script Policy}
\label{script}

As detailed in Sec.~\ref{execmind}, we have designed seven types of macro actions and implemented a hard-coded script policy to perform them in the {\execmind}. Each macro action type serves a specific function. Certain types of macro actions necessitate a specific target.

Within a specific environmental context, the value of each macro action is evaluated, as denoted by $V(a|s)$ in Eq.~(\ref{eq:fastmind}). This evaluation draws upon human expert insights and considers various factors such as the items located on the map, current soup orders, and the readiness of each macro action. For example, the macro action ``Plate Alice Soup'' is given an initial utility value of $0.5$ if there's an active order for Alice Soup. In the absence of such an order, its utility value is set to $0$. However, if Alice Soup lingers in the pot for too long and is at risk of overcooking, the utility value for this action escalates to $1.0$. This increased value acts as a prompt for the agent to give higher priority to serving the soup quickly.

The seven types of macro actions are as follows.

\begin{itemize} 
    \item \emph{Chop:} Transform an ingredient into its chopped form. The target can be Onion, Lettuce or Tomato. The value can be $0$ or $0.5$, depending on whether the ingredient needs to chop in current situation. 
    \item \emph{Mix:} Combine chopped ingredients for cooking. The target can be Alice Ingredients, Bob Ingredients, Cathy Ingredients, or David Ingredients. The value can be $0$ or $0.52$, depending on whether there is an unfinished order that needs the specified mixed ingredients.
    \item \emph{Cook:} Utilize the mixed ingredients to cook a soup. The target can be Alice Soup, Bob Soup, Cathy Soup, or David Soup. The value can be $0$ or $0.54$, depending whether there is an unfinished order that needs cooking the specified soup.
    \item \emph{Plate:} Plate a soup. The target can be Alice Soup, Bob Soup, Cathy Soup or David Soup. The value can be $0$ or $0.56-1$, depending on whether the soup needs serving in hurry, and how long it overcooks.
    \item \emph{Serve:} Deliver a plated soup to the delivery point. The target can be Alice Soup, Bob Soup, Cathy Soup, or David Soup. The value can be $0$ or $0.58-1$, depending on the remaining time of soup order.
    \item \emph{Putout:} Extinguish a fire on a pot using an extinguisher. No target needs to be specified. The value is always $0.6$.
    \item \emph{Drop:} Plate overcooked soup and discard it into a bin. No target needs to be specified. The value is always $0.6$.
\end{itemize}

Most macro actions require at least one empty grid for execution. We have integrated an additional function to all of the macro actions to clean the grids if necessary.

Each macro action is flagged as available when it can be executed immediately. For instance, ``Plate Alice Soup'' becomes available when there is a reachable pot with cooked Alice Soup and a reachable plate. There might be instances when a macro action cannot proceed halfway, for example, the soup gets overcooked when the agent is plating it. Under these circumstances, the macro action will return a ``Failed'' message, prompting the agent to generate a new macro action. The implementation details of the script policy can be found in the open source code.

A macro action like ``Chop Tomato'' involves a sequence of atomic actions including directional movements such as up, down, left, or right. The script policy in this context employs Breadth-First Search (BFS) for efficient pathfinding, while also continuously tracking the environment's state during interaction. For example, in executing the macro action ``Chop Tomato,'' the agent follows a structured procedure: it navigates to the tile where tomatoes are located, picks up a tomato, places it on an unoccupied cutting board, and then chops it. After chopping, the agent moves the chopped tomato to a vacant counter area for future use. Within this workflow, specific actions like ``move to tomato tile,'' ``place tomato on cutting board,'' and ``transfer chopped tomato to counter'' are managed by the BFS algorithm. These actions are further broken down into a series of atomic actions, each involving movement in designated directions. A key aspect of this script policy is its design to minimize interference with the human player's actions.

\section{Implementation of Baseline Agents}
\label{app:baseline}

\subsection{\purelarge}

In {\purelarge} (SMOA), {\fastmind} is discarded. Thus, {\assestage} in {\slowmind} takes in available macro actions as additional input and generates macro actions directly. If the generated macro action is not legal, the LLM regenerates a new one. The Action Filter in {\slowmind} of HLA needs to combine the value of each macro action and the probability of LLM choosing the specific action. Since GPT-3.5 does not provide a valid API to evaluate such probability, we discard Action Filter in SMOA.

\begin{figure*}[ht!]
 \centering
 \includegraphics[width=0.85\linewidth]{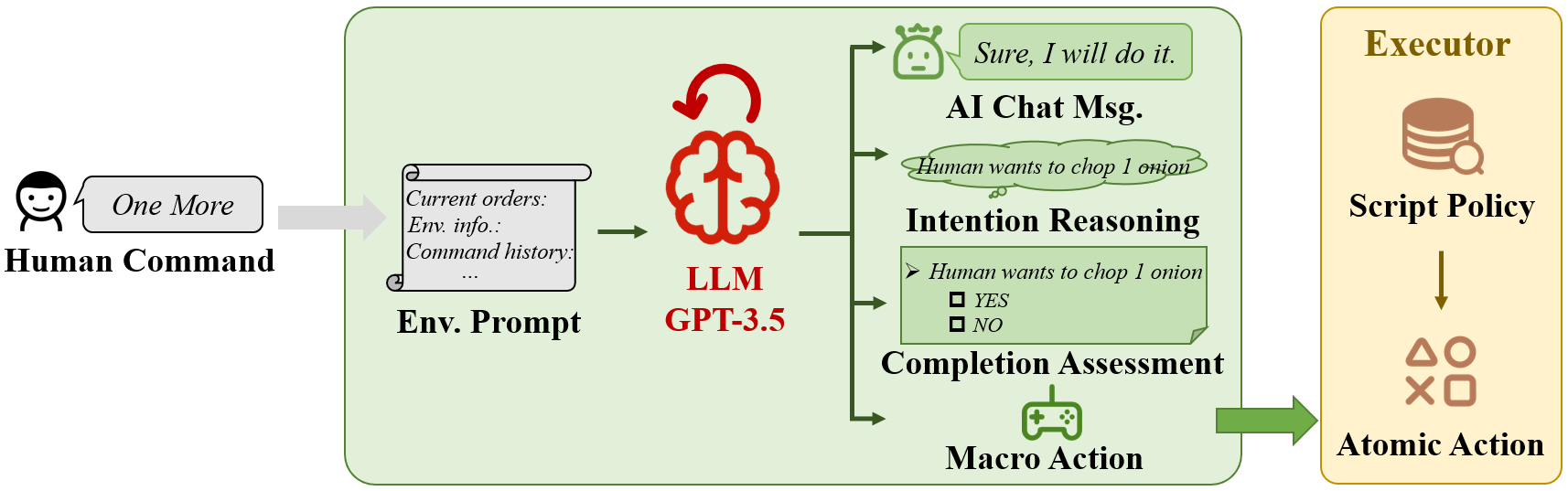}
 \caption{Workflow of {\purelarge}. The {\fastmind} is discarded, and {\slowmind} generates macro actions directly.}
 \label{fig:app_purelarge}
\end{figure*}

The workflow of {\purelarge} is shown in Fig.~\ref{fig:app_purelarge}, where {\fastmind} is discarded. The full prompt of {\assestage} in {\purelarge} is shown in Fig.~\ref{fig:app-prompt-smoa}. In the prompt, we append an additional chat round to let the LLM generate a macro action.

\subsection{{\puresmall}}

In {\puresmall} (FMOA), {\slowmind} is discarded, including both {\intstage} and {\assestage}. The {\fastmind} in FMOA takes in current soup orders, current items, and the history of human commands as additional input. The LLM also generates a chat message aside from a macro action in {\fastmind}. Due to the absence of {\assestage}, FMOA cannot tell whether the human intention is satisfied, and therefore cannot calculate the dynamic adjustment term $\alpha$ in Eq.~(\ref{eq:fastmind}). We set $\alpha$ to be the same as in HLA when human intention is not satisfied. We set the maximum length for macro action history to be $9$, which is roughly the length of finishing $2$ soup orders from scratch (Cook 1 Alice Soup needs 2 Chop, 1 Mix and 1 Cook, while David Soup needs 3 Chop, 1 Mix and 1 Cook). The action history is cleared once the human issues a new command. We assume the human intention is satisfied once the action history reaches the maximum length.

\begin{figure*}[ht!]
 \centering
 \includegraphics[width=0.85\linewidth]{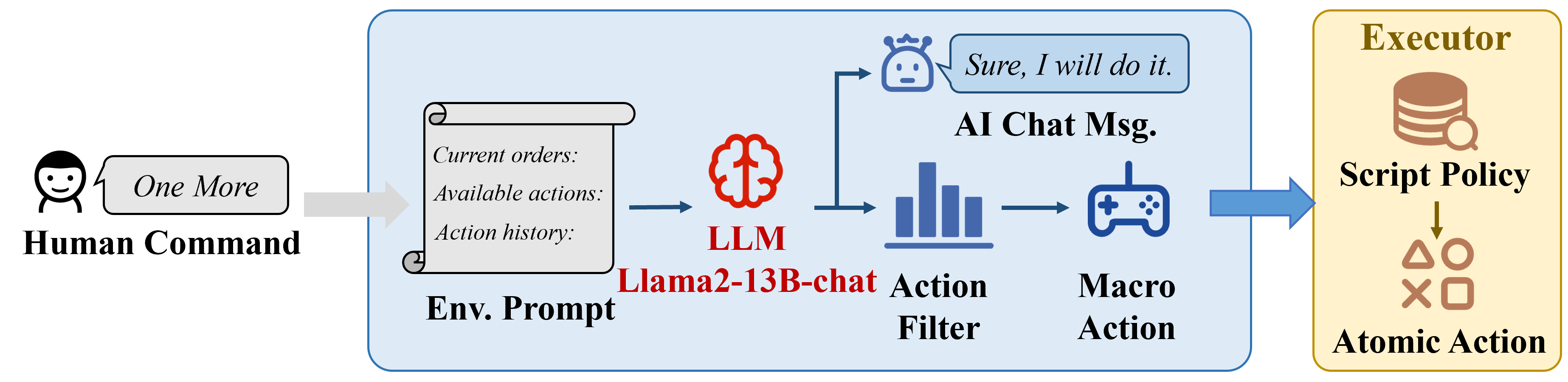}
 \caption{Workflow of {\puresmall}. The {\slowmind} is discarded, and {\fastmind} generates chat message directly.}
 \label{fig:app_puresmall}
\end{figure*}

The workflow of {\puresmall} is shown in Fig.~\ref{fig:app_puresmall}, where {\slowmind} is discarded. The full prompt is shown in Fig.~\ref{fig:app-prompt-fmoa}. The prompt conditions on whether the human intention is satisfied.

\subsection{{\purellm}}

In {\purellm} (NEA), the {\execmind} is discarded, and the {\fastmind} has to generate atomic actions instead of macro actions. To provide spatial information for the {\fastmind}, we append the positions of items and players to the input of {\fastmind}.

\begin{figure*}[ht!]
 \centering
 \includegraphics[width=0.85\linewidth]{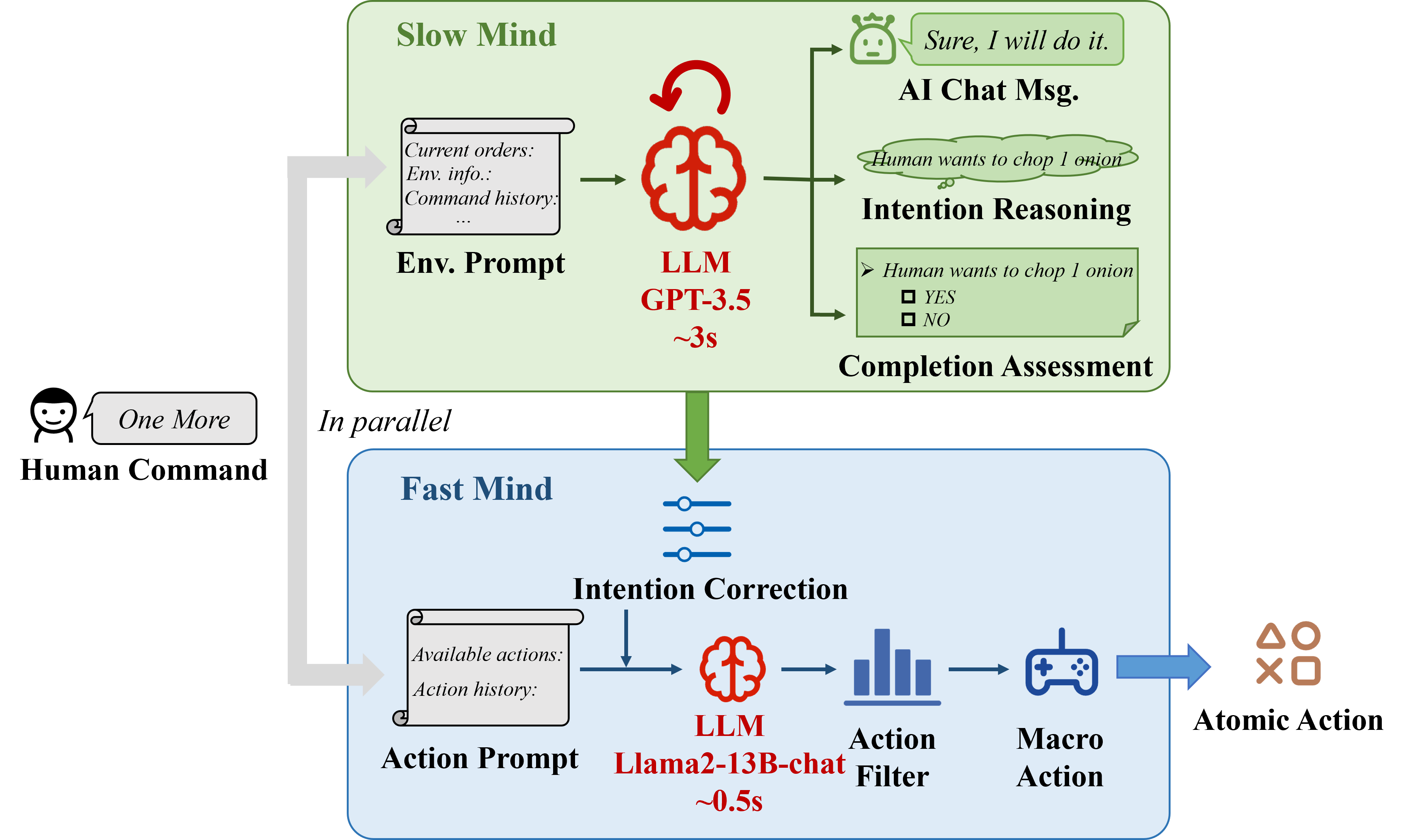}
 \caption{Workflow of {\purellm}. The {\execmind} is discarded, and {\fastmind} generates atomic actions directly.}
 \label{fig:app_purellm}
\end{figure*}

The workflow of {\purellm} is shown in Fig.~\ref{fig:app_purellm}, where {\execmind} is discarded. The full prompt is shown in Fig.~\ref{fig:app-prompt-nea}. In the prompt, we keep the conditional prompt module and add additional position information.

\section{Implementation Details of Command Set}

\subsection{Command Set for Latency Test}
\label{app:abl-latency}

We construct a command set to simulate the real human commands. During the latency test, we do not concern about the actual actions of agents but the latency of their response, so we issue the commands sequentially with a 20-second interval in one round of game play. The response time of each agent is recorded. The test cases are outlined below. 

\begin{enumerate}
    \item You are free to do anything.
    \item Try your best to earn more points.
    \item Focus on the orders.
    \item Chop 3 Lettuce.
    \item Chop 1 Onion.
    \item Chop 2 more.
    \item Cook Bob Soup.
    \item Cook it again.
    \item Alice soup is about to timeout!
    \item Watch out for the Cathy Soup order.
    \item Help me with the third soup on the orders please.
    \item Chop more vegetables.
    \item Aba Aba. Chop 1 potato.
    \item What are the orders?
    \item What is Alice Soup?
\end{enumerate}

\textbf{Metric.}
We use \emph{macro action latency} and \emph{atomic action latency} to measure the real-time responsiveness of HLA and baseline agents. The macro action latency is defined as the time interval between receiving a human command and subsequently generating a macro action. 

The atomic action latency is the latency of an atomic action. For SMOA, FMOA and HLA, the AI agent only suffers from the latency of generating an macro action. In other words, these agents has the latency of generating the first atomic action after receiving a human command, but can output the following atomic actions swiftly once the macro action is generated. This is because {\execmind} can translate macro action in to atomic action with minimal latency.
For NEA, the atomic action is directly generated by {\fastmind}, which means NEA suffers from a latency every time it produces an atomic action.

To fairly compare the atomic action latency, we calculate the mean atomic action generation latency for SMOA, FMOA and HLA, which is more consistent with the subjective feelings of human players.
For SMOA, FMOA and HLA, the atomic action latency is defined as $T_a = \frac{T_m}{N_a}$, where $T_m$ is the generation latency of the first macro action after receiving a human command, and $N_a$ is the number of atomic actions in this macro action. For NEA, the atomic action latency is defined as the generation latency of the first atomic action after receiving a human command.

\subsection{Complex Command Set}

\label{app:abl-complex}

As discussed in Sec.~\ref{sec:exp-abl-complex}, we construct a command set consist of a total of 30 complex commands for 3 challenges mentioned in Sec.~\ref{sec: challenges}. Experiment are conducted using the Map {\maphard}. The commands for quantity specification challenge are as follows. 

\begin{enumerate}
    \item Chop 1 Onion.
    \item Chop two onions.
    \item Please chop 3 onions.
    \item Cut one Tomato.
    \item Help me cut 2 tomatoes.
    \item 3 chopped tomatoes please.
    \item Chop 1 Lettuce.
    \item 2 lettuces chop.
    \item help me to chop 3 lettuces.
    \item Cook Alice Soup once.
\end{enumerate}

The commands for semantic analysis challenge are as follows. 

\begin{enumerate}
    \item I need more onions
    \item Chop but except tomato and lettuce.
    \item Why are we always short of tomatoes?
    \item Just pass me toma and don't ask why.
    \item Can't you see the lettuce, uh?
    \item The green cabbage looks perfect!
    \item Bob Soup is about to timeout!
    \item Oh god, I forget the alice soup order.
    \item Come on! There is a cathy order!
    \item D soup!
\end{enumerate}

The commands for ambiguous reference challenge are as follows. 

\begin{enumerate}
    \item Chop 2 Onions -> Chop it again.
    \item Chop 3 Tomatoes -> One more please.
    \item Cut one lettuce -> Cut more!
    \item Cook Bob Soup. -> Cook it again!
    \item Cook 2 cathy soup, -> Can you do it again?
    \item Cook david soup once -> Help me with that again.
    \item Cook the first soup in the orders
    \item Cook the second order now!
    \item The third soup order should be cooked
    \item Please help me cook the last soup order
\end{enumerate}

During the test of each command, the soup orders are carefully designed and fixed. For chopping commands, the target ingredient doesn't appear in any soup orders. For cooking commands, we ensure that the required soup is not part of any soup orders, except for the final 4 commands in ambiguous reference challenge, where the commands directly correspond with the orders.

\section{Additional Results}

All experiments were conducted on a computer equipped with A100-80G GPU.

\subsection{Latency of Each Module}
\label{app:abl-comp}

In addition to the end-to-end latency discussed in Sec.~\ref{sec:exp-abl-latency}, we also measure the latency of each module of HLA and baseline agents, as shown in Tab.~\ref{tab:exp-abl3_comp}, where IR denotes the {\intstage}, CA denotes the {\assestage}, and MA denotes Macro Action Generation in the {\fastmind}. Both SMOA and FMOA suffers from high latency, which originates from generating macro actions and giving chat message at the same time. This further validates the hierarchical design of HLA which can decouple these components.

\subsection{Visualization of Simple Commands}
\label{app:abl-score}
\begin{figure*}[ht!]
 \centering
     \begin{subfigure}[t]{0.47\textwidth}
         \includegraphics[width=0.95\textwidth]{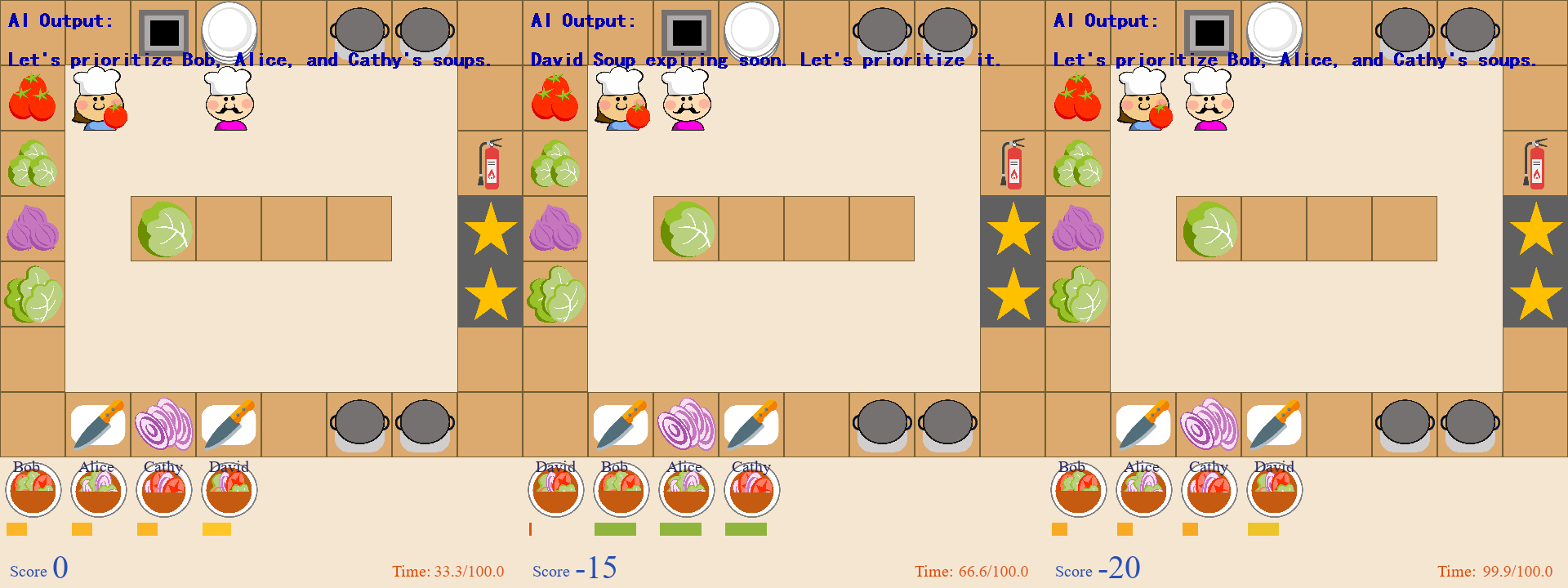}
         \caption{No Command}
     \end{subfigure}
     \hfill
     \begin{subfigure}[t]{0.47\textwidth}
         \includegraphics[width=0.95\textwidth]{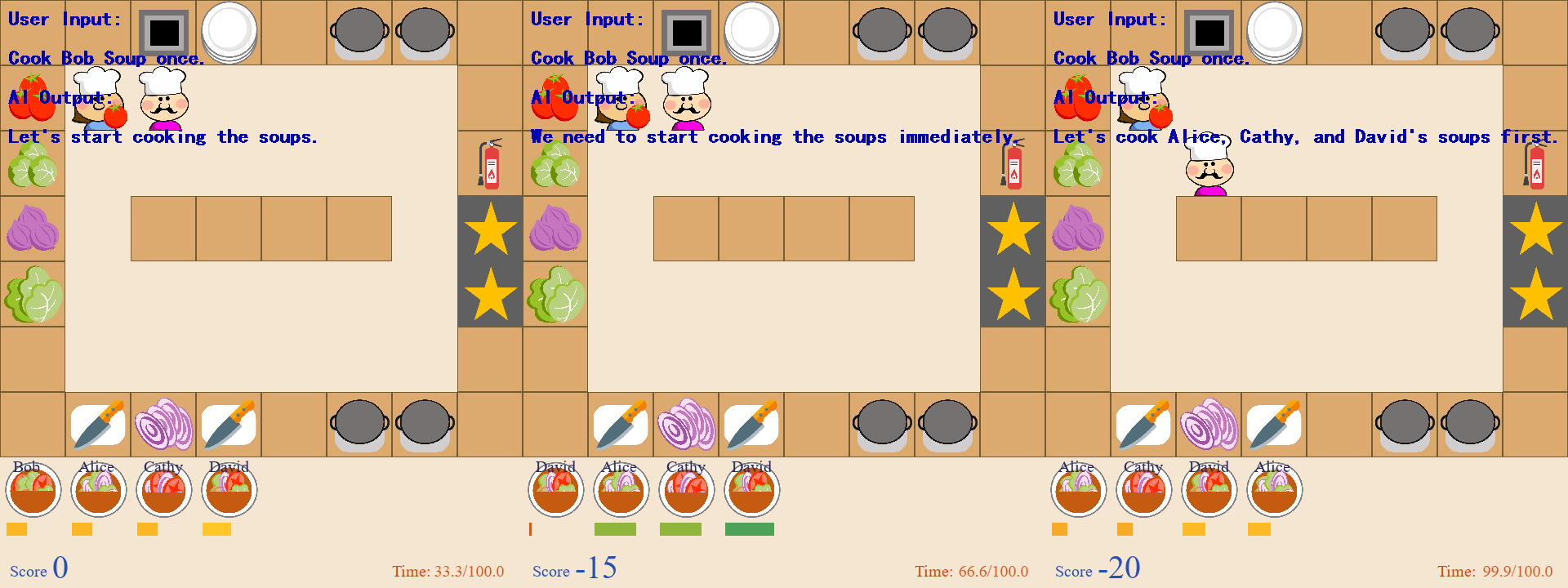}
         \caption{One Command}
     \end{subfigure}
\caption{Visualization result of NEA for latency test on simple command set. Screenshots are captured at $1/3$ of gameplay, $2/3$ of gameplay, and upon completion of the game.}
\label{fig:exp-alb1-ss-NEA}
\end{figure*}

\begin{figure*}[ht!]
 \centering
     \begin{subfigure}[t]{0.47\textwidth}
         \includegraphics[width=0.95\textwidth]{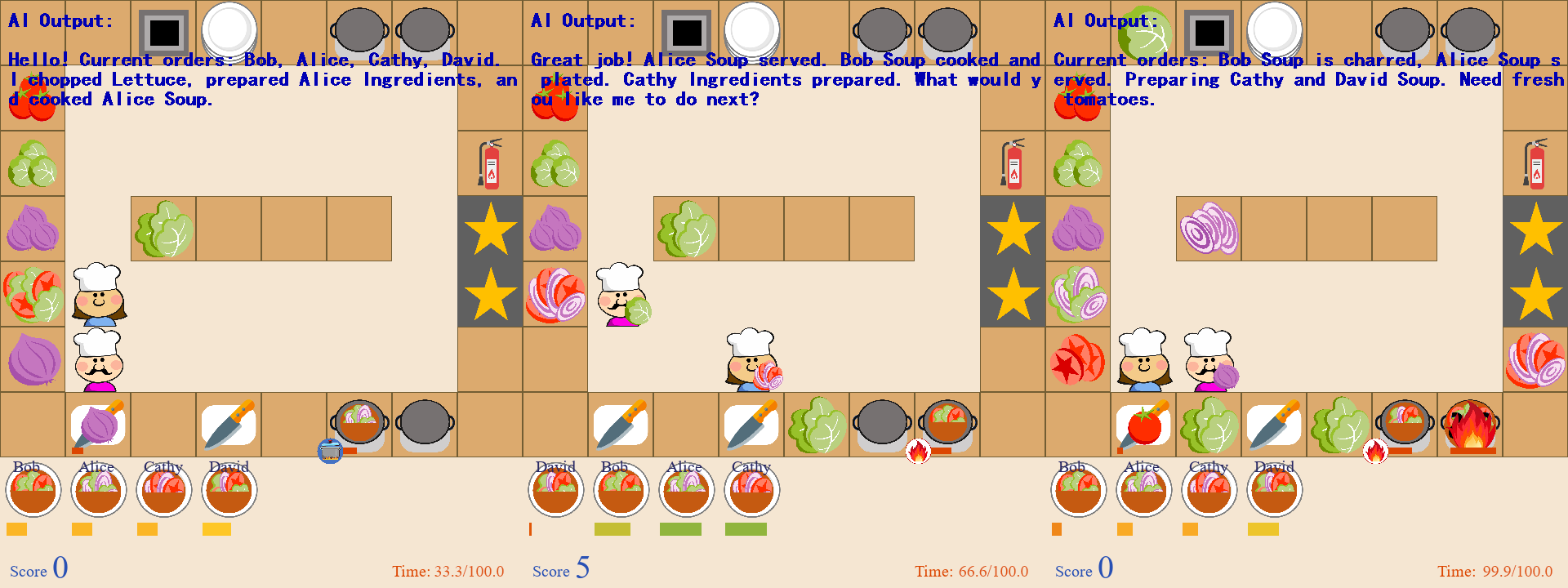}
         \caption{No Command}
     \end{subfigure}
     \hfill
     \begin{subfigure}[t]{0.47\textwidth}
         \includegraphics[width=0.95\textwidth]{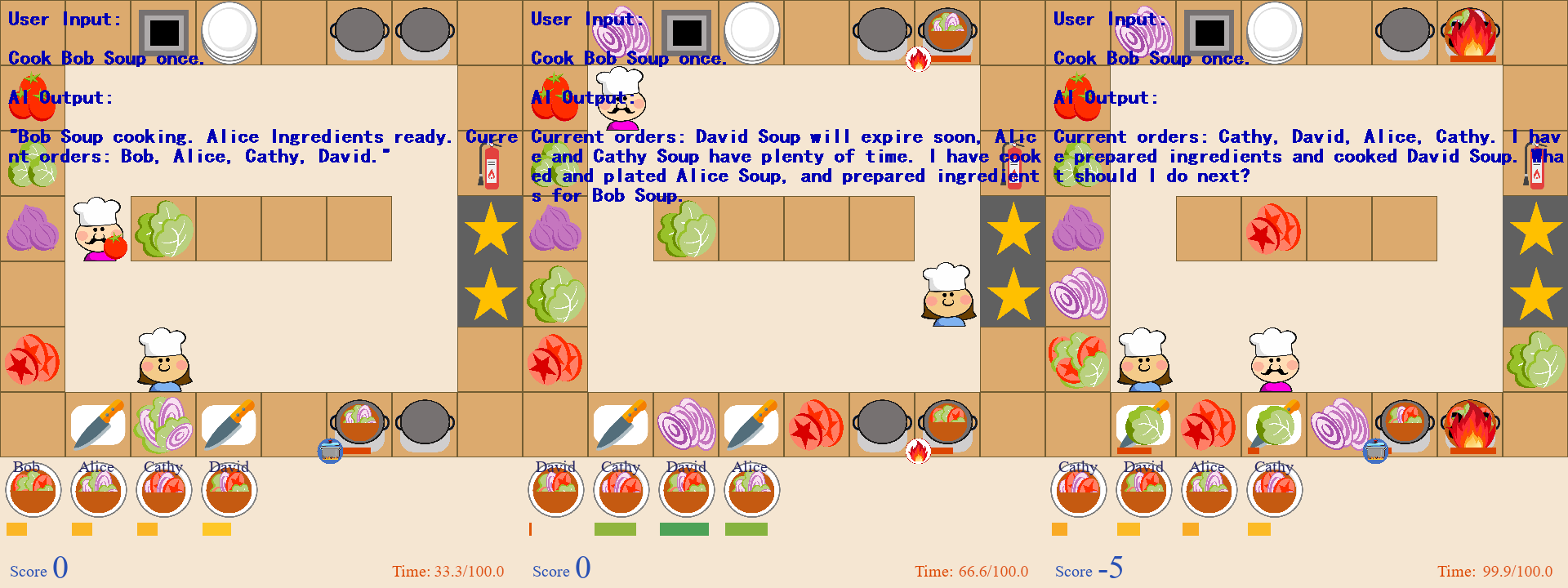}
         \caption{One Command}
     \end{subfigure}
\caption{Visualization result of SMOA for latency test on simple command set. Screenshots are captured at $1/3$ of gameplay, $2/3$ of gameplay, and upon completion of the game.}
\label{fig:exp-alb1-ss-SMOA}
\end{figure*}

\begin{figure*}[ht!]
 \centering
     \begin{subfigure}[t]{0.47\textwidth}
         \includegraphics[width=0.95\textwidth]{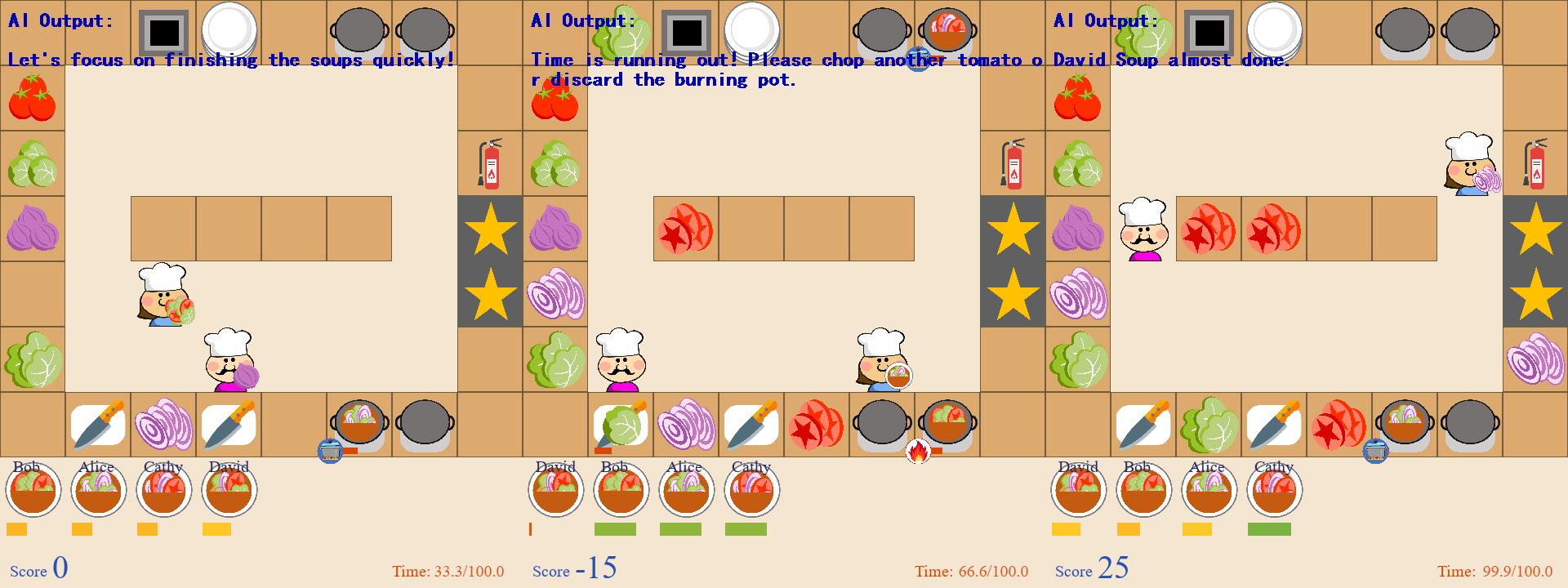}
         \caption{No Command}
     \end{subfigure}
     \hfill
     \begin{subfigure}[t]{0.47\textwidth}
         \includegraphics[width=0.95\textwidth]{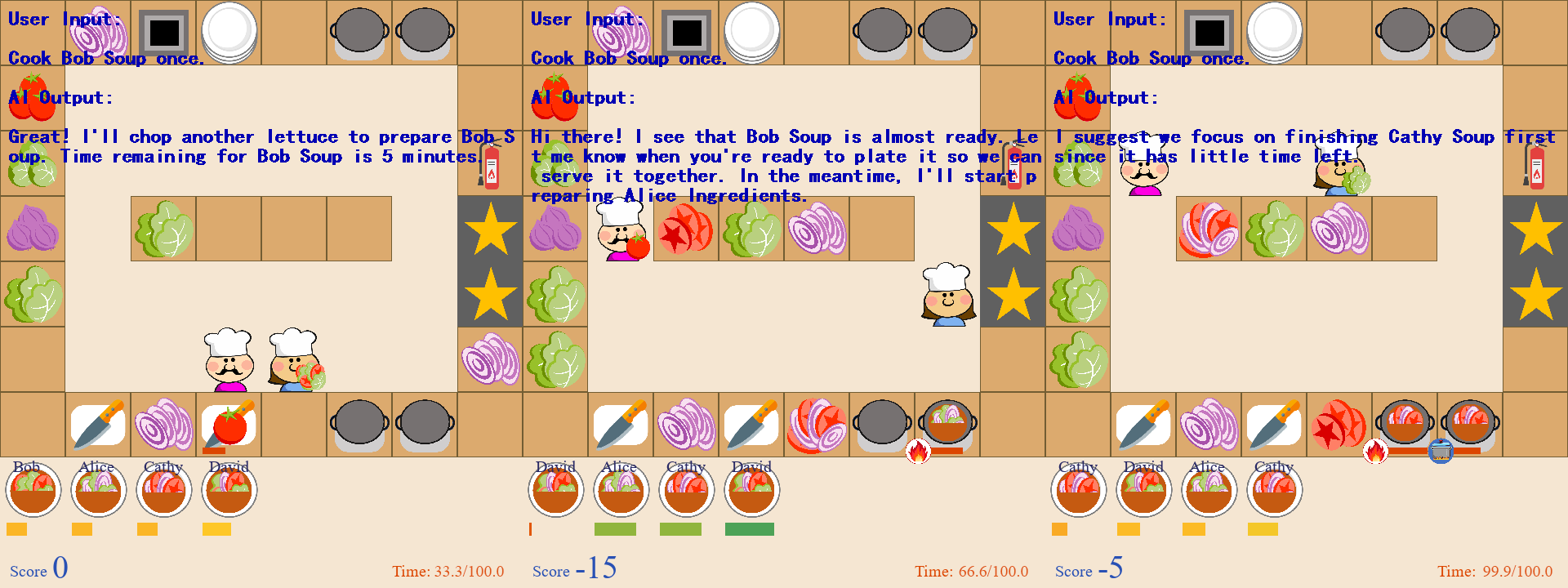}
         \caption{One Command}
     \end{subfigure}
\caption{Visualization result of FMOA for latency test on simple command set. Screenshots are captured at $1/3$ of gameplay, $2/3$ of gameplay, and upon completion of the game.}
\label{fig:exp-alb1-ss-FMOA}
\end{figure*}

\begin{figure*}[ht!]
 \centering
     \begin{subfigure}[t]{0.47\textwidth}
         \includegraphics[width=0.95\textwidth]{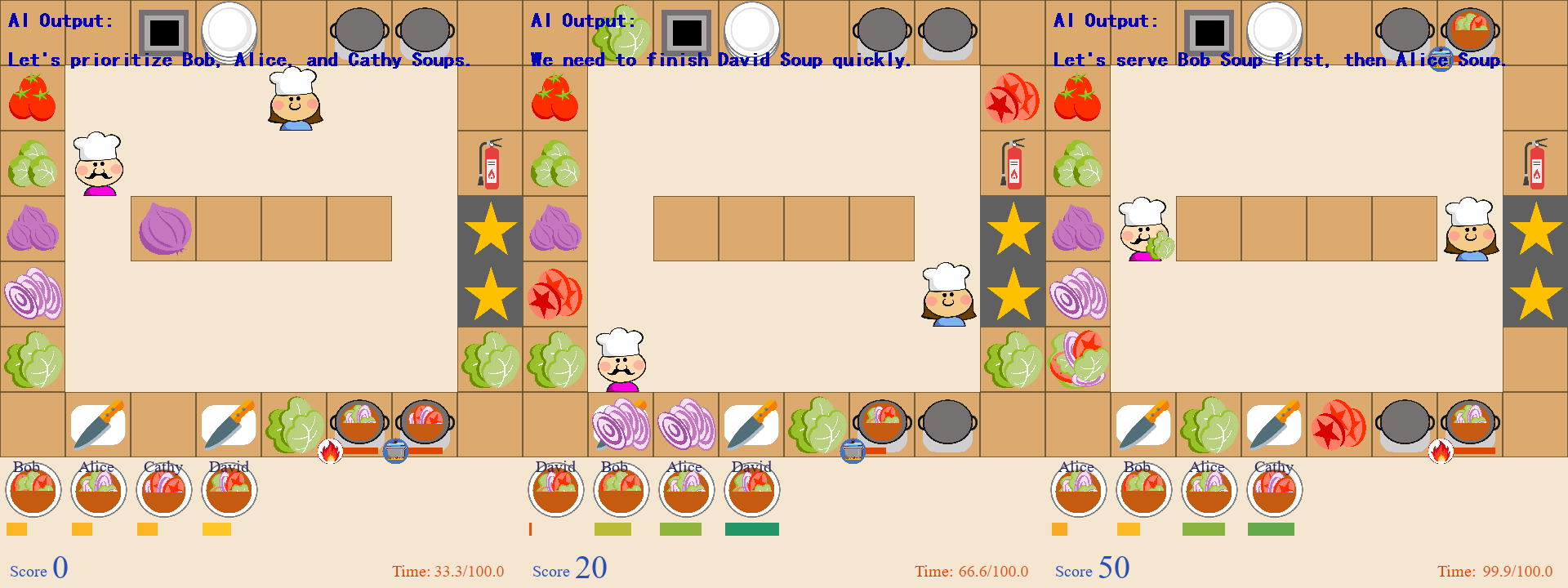}
         \caption{No Command}
     \end{subfigure}
     \hfill
     \begin{subfigure}[t]{0.47\textwidth}
         \includegraphics[width=0.95\textwidth]{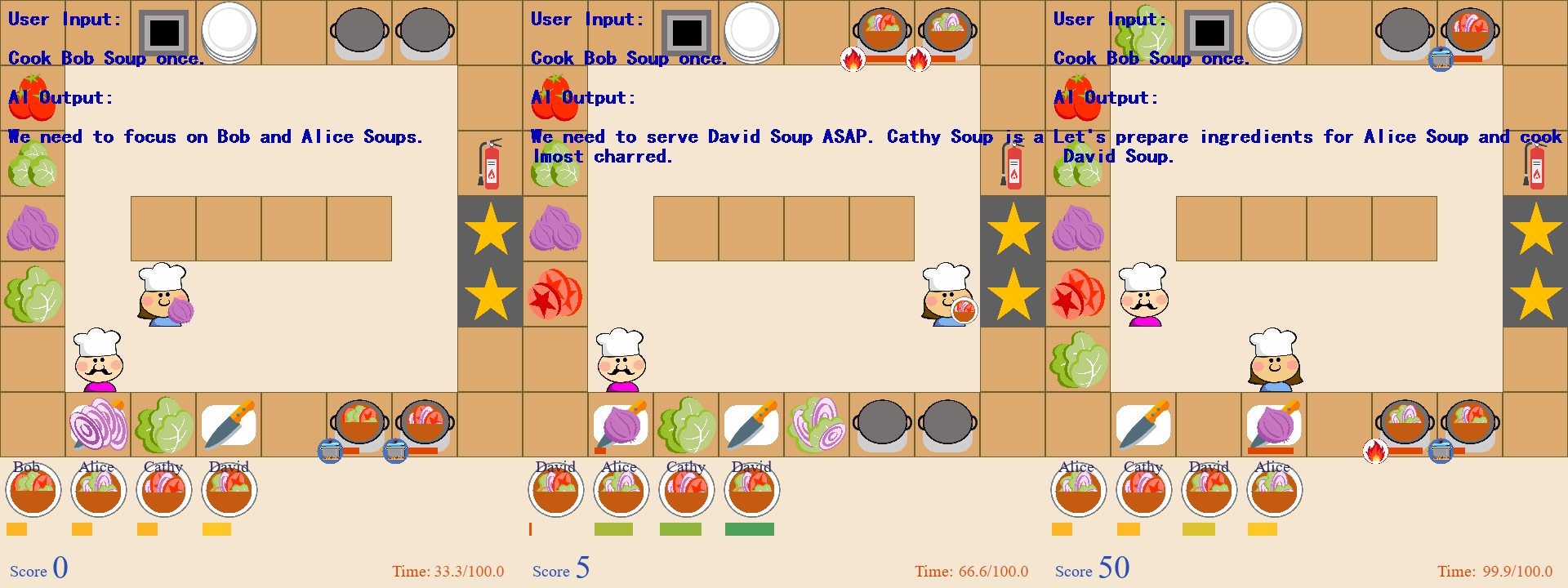}
         \caption{One Command}
     \end{subfigure}
\caption{Visualization result of HLA for latency test on simple command set. Screenshots are captured at $1/3$ of gameplay, $2/3$ of gameplay, and upon completion of the game.}
\label{fig:exp-alb1-ss-HLA}
\end{figure*}

As discussed in Sec.~\ref{sec:exp-abl-simple}, we use two simple commands, \emph{No Command} and \emph{One Command}, to test the cooperative ability of HLA and baseline agents. The visualization results of {\purellm} (NEA), {\purelarge} (SMOA), {\puresmall} (FMOA), and HLA are shown in Fig.~\ref{fig:exp-alb1-ss-NEA}, Fig.~\ref{fig:exp-alb1-ss-SMOA}, Fig.~\ref{fig:exp-alb1-ss-FMOA}, and Fig.~\ref{fig:exp-alb1-ss-HLA}, respectively. The game screenshots are captured at $1/3$ of gameplay, $2/3$ of gameplay, and upon completion of the game.
The game score can be found in the bottom-left corner of the screenshot. The blue text within the screenshot shows the human player chat message and AI agent chat response.

NEA is stuck at the top left corner of the map. This is because NEA continuously gives the atomic action of moving \emph{left} and is stuck next to the tomato tile. SMOA successfully mixes the ingredients and cooks soups. But upon completion of the game, the AI player fails to plate the soup in time, and thus overcooks the soup and sets the pot on fire. FMOA gives hallucinating chat response. In \emph{No Command}, at $2/3$ of gameplay, FMOA talks about ``the burning pot'', but no pot is on fire at that time. In \emph{One Command}, at $2/3$ of gameplay, FMOA still talks about ``Bob Soup is almost ready'' and tries to cook Bob Soup, but the Bob Soup is already served prior and there is no Bob Soup order at that moment.

HLA manages to use all 3 pots simultaneously upon completion of the game in \emph{One Command}. In \emph{One Command}, at $2/3$ of gameplay, HLA talks about ``Cathy Soup is almost charred'' and plates Cathy Soup afterwards, showing a consistency of its action and chat message. This chat message comes out before the screenshot is taken. 

We suggest visiting our website for more video demonstrations.

\subsection{Introducing Action Filter into SMOA}

\begin{table*}[ht]
\centering
\begin{tabular}{ccc}
    \toprule
    & Mac. Act. Latency(s) & Ato. Act. Latency(s) \\
    \midrule
    HLA & \textbf{1.07 \scriptsize{(0.22)}} & \textbf{0.08 \scriptsize{(0.06)}} \\
    SMOA & 4.16 \scriptsize{(1.01)} & 0.61 \scriptsize{(0.65)} \\
    SMOA+action filter & 6.23 \scriptsize{(1.87)} & 0.52 \scriptsize{(0.56)} \\
    \bottomrule
\end{tabular}
\centering
\caption{Macro action latency and atomic action latency of SMOA and variant. The format is ``mean (standard deviation)''.}
\label{tab:exp-prob_latency}
\end{table*}

\begin{table*}[ht]
\centering
\begin{tabular}{ccc}
    \toprule
    & No Command & One Command \\
    \midrule
    HLA & \textbf{55.0 \scriptsize{(5.5)}} & \textbf{47.0 \scriptsize{(11.2)}} \\
    SMOA & 1.0 \scriptsize{(2.0)} & -7.0 \scriptsize{(6.8)} \\
    SMOA+action filter & -14.0 \scriptsize{(12.0)} & -20.0 \scriptsize{(0.0)} \\
    \bottomrule
\end{tabular}
\centering
\caption{Average game score of SMOA and variant. The format is ``mean (standard deviation)''.}
\label{tab:exp-prob_simple}
\end{table*}

\begin{table*}[ht]
\centering
\begin{tabular}{c|cc|cc|cc}
    \toprule
    \multirow{2}{*}{AI Agents} & \multicolumn{2}{c|}{Quantity} & \multicolumn{2}{c|}{Semantics} & \multicolumn{2}{c}{Ambiguity} \\ 
    \cmidrule{2-7}
    & Suc.$\uparrow$ & Time$\downarrow$ & Suc.$\uparrow$ & Time$\downarrow$ & Suc.$\uparrow$ & Time$\downarrow$ \\
    \midrule
    HLA & \textbf{1.00} & \textbf{13.30} & \textbf{0.90} & \textbf{17.13} & \textbf{0.70} & \textbf{27.78} \\
    SMOA & 0.40 & 47.14 & 0.60 & 40.20 & \textbf{0.70} & 39.57 \\
    SMOA+action filter & 0.40 & 48.44 & 0.30 & 50.49 & 0.20 & 52.37 \\
    \bottomrule
\end{tabular}
\centering
\caption{Success rate and completion time for complex commands of SMOA and variant.}
\label{tab:exp-prob_complex}
\end{table*}

To assess the potential of adding an action filter to the Slow-Mind-Only Agent (SMOA), we set up GPT-3.5 to create both the macro action and its estimated probability independently, despite GPT-3.5 lacking a designated API for such probability evaluation. The prompt given to GPT-3.5 is to ``output the action and the probability of each action.'' In response, GPT-3.5 produces outputs like ``\{Chop Tomato: 0.33, Chop Lettuce: 0.33, Chop Onion: 0.33\}.'' An action filter is then applied to process both the action and its probability. This method is referred to as ``SMOA+action filter''.

The results of our experiments are presented in Tab. \ref{tab:exp-prob_latency}, Tab. \ref{tab:exp-prob_simple}, and Tab. \ref{tab:exp-prob_complex}. In these tests, we found that adding an action filter to SMOA resulted in about a 50\% increase in macro action latency compared to the standard SMOA. Furthermore, the performance for both simple and complex commands decreased notably. This decrease in performance is attributed to the extended output of GPT-3.5 when it is required to generate actions along with their probabilities, which in turn leads to longer response times. Another factor contributing to this decline is the method of generating action probabilities. Relying on the LLM to estimate the probability of each action candidate proved less accurate than utilizing the ``logprob'' feature from the LLM's output head. However, as this feature is currently not available in GPT-3.5, it impacted the overall performance negatively. Based on these findings, we decided to omit the action filter from SMOA in our comparisons to ensure fairness.

\subsection{Comparison of Different LLMs}

\begin{table*}[ht]
\centering
\begin{tabular}{ccc}
    \toprule
    & Mac. Act. Latency(s) & Ato. Act. Latency(s) \\
    \midrule
    HLA & 1.07 \scriptsize{(0.22)} & 0.08 \scriptsize{(0.06)} \\
    HLA-Fast7B & \textbf{0.76 \scriptsize{(0.11)}} & \textbf{0.06 \scriptsize{(0.07)}} \\
    HLA-Slow70B & 1.11 \scriptsize{(0.20)} & \textbf{0.06 \scriptsize{(0.02)}} \\
    \bottomrule
\end{tabular}
\centering
\caption{Macro action latency and atomic action latency of employing different LLM in HLA. The format is ``mean (standard deviation)''.}
\label{tab:exp-llm_latency}
\end{table*}

\begin{table*}[ht]
\centering
\begin{tabular}{ccc}
    \toprule
    & No Command & One Command \\
    \midrule
    HLA & 55.0 \scriptsize{(5.5)} & \textbf{47.0 \scriptsize{(11.2)}} \\
    HLA-Fast7B & \textbf{64.0 \scriptsize{(2.0)}} & 40.0 \scriptsize{(16.4)} \\
    HLA-Slow70B & 43.0 \scriptsize{(16.3)} & 31.0 \scriptsize{(16.4)} \\
    \bottomrule
\end{tabular}
\centering
\caption{Average game score of employing different LLM in HLA. The format is ``mean (standard deviation)''.}
\label{tab:exp-llm_simple}
\end{table*}

\begin{table*}[ht]
\centering
\begin{tabular}{c|cc|cc|cc}
    \toprule
    \multirow{2}{*}{AI Agents} & \multicolumn{2}{c|}{Quantity} & \multicolumn{2}{c|}{Semantics} & \multicolumn{2}{c}{Ambiguity} \\ 
    \cmidrule{2-7}
    & Suc.$\uparrow$ & Time$\downarrow$ & Suc.$\uparrow$ & Time$\downarrow$ & Suc.$\uparrow$ & Time$\downarrow$ \\
    \midrule
    HLA & \textbf{1.00} & \textbf{13.30} & \textbf{0.90} & \textbf{17.13} & \textbf{0.70} & 27.78 \\
    HLA-Fast7B & \textbf{1.00} & 14.71 & 0.70 & 24.12 & \textbf{0.70} & \textbf{26.95} \\
    HLA-Slow70B & 0.50 & 37.14 & 0.40 & 38.04 & 0.40 & 44.98 \\
    \bottomrule
\end{tabular}
\centering
\caption{Success rate and completion time for complex commands of employing different LLM in HLA.}
\label{tab:exp-llm_complex}
\end{table*}

In this part, we analyze the performance differences when using various large language models (LLMs) within the {\slowmind} and the {\fastmind}. Specifically, ``HLA-Fast7B'' indicates the use of Llama2-7B-chat as the {\fastmind} instead of Llama2-13B-chat, while ``HLA-Slow70B'' refers to the use of Llama2-70B-chat in the {\slowmind}, replacing GPT-3.5. Both of the models use the same quantization technique as in HLA. The results regarding macro action latency and atomic action latency are detailed in Tab.~\ref{tab:exp-llm_latency}. Tab.~\ref{tab:exp-llm_simple} presents the average game scores for scenarios under \emph{No Command} and \emph{One Command} test cases on Map {\maphard}. Additionally, Tab.~\ref{tab:exp-llm_complex} outlines the average success rate and completion time when processing complex commands.

Tab.~\ref{tab:exp-llm_latency}, Tab.~\ref{tab:exp-llm_simple} and Tab.~\ref{tab:exp-llm_complex}, illustrate the performance of HLA-Fast7B, revealing a reduction of 29\% in macro action latency and 25\% in atomic action latency compared to HLA. This enhanced performance is primarily attributed to the lower computational requirements of Llama2-7B-chat. However, a notable limitation of HLA-Fast7B is its diminished efficacy in handling ``One Command'' test cases and in interpreting semantically complex commands. As a result, for the role of the {\fastmind}, we have opted for Llama2-13B-chat, which offers a well-rounded balance between the reasoning capability and the inference speed.
On the other hand, HLA-Slow70B exhibits similar macro and atomic action latencies but struggles significantly with both simple and complex commands. This underscores the critical role of the Slow Mind's strong reasoning capabilities. Consequently, in our study, we have chosen GPT-3.5 as the {\slowmind} due to its superior performance in these areas.

\subsection{Interpreting Complex Commands}

\begin{table*}[ht]
\centering
\begin{tabular}{ccc}
    \toprule
    & Semantics & Ambiguity \\
    \midrule
    HLA (slowmind) & 100\% & 100\% \\
    HLA (fastmind) & 33\% & 0\% \\
    \bottomrule
\end{tabular}
\centering
\caption{Failure rate of Slow Mind and Fast Mind in HLA for two subsets of complex commands.}
\label{tab:exp-int_fail}
\end{table*}

Tab.~\ref{tab:exp-abl4_instruct1} in the main paper illustrates the success rate of HLA in interpreting different subsets of complex commands. It was observed that $10\%$ of semantically complex commands and $30\%$ of ambiguous commands were not successfully interpreted. Additionally, Tab.~\ref{tab:exp-int_fail} provides a detailed analysis of the failure rates attributable to the {\slowmind} and the {\fastmind} for these two subsets of commands. \emph{HLA(slowmind)} denotes the failure rate linked to the {\slowmind}, focusing on incorrect reasoning of intentions or assessment of task completion. \emph{HLA(fastmind)} indicates the failure rate associated with the {\fastmind}, where incorrect reasoning and assessment are identified and replaced with correct ones, followed by a recalculation of the failure rate.
 
As indicated in Tab.~\ref{tab:exp-int_fail}, the {\slowmind} is identified as the cause of all observed failures. The {\slowmind} incorrectly interprets human intentions as ``None'' or wrong intentions, which leads to the failures. For example, the semantic command (3), ``Why are we always short of tomatoes?''. The {\slowmind} interprets the intention behind this command as ``None''. Similarly, with the ambiguous command (6), ``Cook David soup once -> Help me with that again'', the intention deduced by the {\slowmind} is incorrectly identified as ``Cook Bob Soup once and Cook Alice Soup once.''

Further examination reveals that when we provide the correct human intentions and task completions to the {\fastmind}, it still exhibits a 33\% failure rate. For example, consider the ambiguous command (5), ``Cook 2 Cathy soup, -> Can you do it again?''. Both the {\slowmind} and {\fastmind} fail in this instance. The {\slowmind} misinterprets the human intention as ``None.'' However, even when supplied with the correct intention, the {\fastmind} still executes some unnecessary macro actions before cooking the second Cathy Soup. This inefficiency results in time wastage and failure to complete the command within the designated time limit.

\subsection{Human Studies}
We additionally report the latency of each module, end-to-end latency, hit rate of macro actions generated by {\fastmind}, detailed human preference, and visualized replay in the following subsections.

\subsubsection{Details on Behavior Analysis}
\label{sec:app-hci-use}

As mentioned in Sec.~\ref{sec:exp-hci-use}, we calculate the rate of valuable macro actions, i.e.,
$p_{valuable} = \frac{N_{valuable}}{N_{all}}$, where $N_{valuable}$ is the number of the specified macro actions that are valuable, and $N_{all}$ is the total number of the specified macro actions.
The detailed definition of useful macro actions are as follows.

\begin{itemize}
    \item \emph{Chop.} The moment the ingredient is put onto the cut board, a relevant unfinished order that hasn't been cooked exists, and no chopped form of this ingredient is on the map for this order.
    \item \emph{Cook.} The moment the mixed ingredients are put into an empty pot, an unfinished order that hasn't been cooked exists.
    \item \emph{Serve.} The moment the soup is served to the delivery point, an unfinished order of the specified soup exists.
\end{itemize}

Besides, we also examine the rate of fire accidents, i.e., $p_{fire} = \frac{R_{fire}}{R_{all}}$, where $R_{fire}$ is the number of game rounds that at least one pot catches fire, and $R_{all}$ is the number of total game rounds.

\begin{table*}[t]
\centering
\begin{tabular}{ccccccc}
    \toprule
    & \multicolumn{2}{c}{SMOA} & \multirow{2}{*}{FMOA} & \multicolumn{3}{c}{HLA} \\
    \cmidrule{2-3} \cmidrule{5-7}
     & IR & CA+MA &  & IR & CA & MA \\
    \midrule
    Latency(s) & 0.90 \scriptsize{(0.34)} & 2.87 \scriptsize{(0.50)} & 3.59 \scriptsize{(0.87)} & 0.74 \scriptsize{(0.26)} & 1.88 \scriptsize{(0.69)} & 1.00 \scriptsize{(0.21)} \\
    \bottomrule
\end{tabular}
\centering
\caption{Latency of each module in latency test. IR denotes for {\intstage}, CA denotes {\assestage} and MA denotes Macro Action Generation in {\fastmind}. Standard deviations are shown in parentheses.}
\label{tab:exp-abl3_comp}
\end{table*}

\begin{table*}[ht]
\vspace{-2mm}
\centering
\begin{tabular}{cccc|c}
    \toprule
    AI Players & \textit{Chop $\uparrow$} & \textit{Cook $\uparrow$} & \textit{Serve $\uparrow$} & \textit{Fire $\downarrow$} \\ \midrule
    SMOA & 0.50 & \textbf{1.00} & 0.00 & \textbf{0.00} \\
    FMOA & 0.74 & 0.85 & 0.80 & \textbf{0.00} \\
    HLA & \textbf{0.80} & 0.98 & \textbf{1.00} & \textbf{0.00} \\ 
    \bottomrule
\end{tabular}
\centering
\caption{The ratio of valuable macro actions and the frequency of fire accidents of different AI players in {\mapring} during the competition phase.}
\label{tab:exp-hci_use_1}
\vspace{-2mm}
\end{table*}

\begin{table*}[ht]
\vspace{-2mm}
\centering
\begin{tabular}{cccc|c}
    \toprule
    AI Players & \textit{Chop $\uparrow$} & \textit{Cook $\uparrow$} & \textit{Serve $\uparrow$} & \textit{Fire $\downarrow$} \\ \midrule
    SMOA & 0.59 & 0.88 & 0.67 & \textbf{0.00} \\
    FMOA & 0.80 & 0.67 & 0.92 & \textbf{0.00} \\
    HLA & \textbf{0.82} & \textbf{0.96} & \textbf{1.00} & \textbf{0.00} \\
\bottomrule
\end{tabular}
\centering
\caption{The ratio of valuable macro actions and the frequency of fire accidents of different AI players in {\mapbot} during the competition phase.}
\label{tab:exp-hci_use_2}
\vspace{-2mm}
\end{table*}

\begin{table*}[ht]
\vspace{-2mm}
\centering
\begin{tabular}{cccc|c}
    \toprule
    AI Players & \textit{Chop $\uparrow$} & \textit{Cook $\uparrow$} & \textit{Serve $\uparrow$} & \textit{Fire $\downarrow$} \\ \midrule
    SMOA & 0.55 & 0.73 & / & 0.47 \\
    FMOA & 0.66 & 0.84 & / & 0.13 \\
    HLA & \textbf{0.79} & \textbf{0.90} & / & \textbf{0.00} \\ 
\bottomrule
\end{tabular}
\centering
\caption{The ratio of valuable macro actions and the frequency of fire accidents of different AI players in {\mappart} during the competition phase. \emph{Serve} macro action is marked as ``/'' since only the human player can serve orders in this map.}
\label{tab:exp-hci_use_3}
\vspace{-2mm}
\end{table*}

\begin{table*}[ht!]
\vspace{-2mm}
\centering
\begin{tabular}{cccc|c}
    \toprule
    AI Players & \textit{Chop $\uparrow$} & \textit{Cook $\uparrow$} & \textit{Serve $\uparrow$} & \textit{Fire $\downarrow$} \\ \midrule
    SMOA & 0.61 & 0.73 & 0.20 & \textbf{0.05} \\
    FMOA & 0.81 & 0.88 & 0.83 & 0.10 \\
    HLA & \textbf{0.87} & \textbf{0.98} & \textbf{1.00} & 0.10 \\
\bottomrule
\end{tabular}
\centering
\caption{The ratio of valuable actions and the frequency of fire occurrences of different AI players in {\maphard} in competition phase.}
\label{tab:exp-hci_use_4}
\vspace{-2mm}
\end{table*}

Additional results of each map can be found in Tab.~\ref{tab:exp-hci_use_1}, Tab.~\ref{tab:exp-hci_use_2}, Tab.~\ref{tab:exp-hci_use_3} and Tab.~\ref{tab:exp-hci_use_4}. In {\mappart}, the AI player cannot get access to the delivery point and doesn't serve any soup. The result of ``Serve'' is set to ``/'' in the table.

\subsubsection{Latency of Each Module}

Tab.~\ref{tab:exp-hci_compw} shows the latency of each module of different AI players in the preparation phase of human studies, and Tab.~\ref{tab:exp-hci_comp} shows the results in the competition phase.
IR denotes {\intstage}, CA denotes {\assestage}, and MA denotes Macro Action Generation in {\fastmind}. The results are consistent with the previous results in Sec.~\ref{app:abl-comp}. Simultaneous generation of macro actions and chat messages introduces significant latency into SMOA and FMOA.

\begin{table*}[t]
\centering
\begin{tabular}{ccccccc}
    \toprule
    & \multicolumn{2}{c}{SMOA} & \multirow{2}{*}{FMOA} & \multicolumn{3}{c}{HLA} \\
    \cmidrule{2-3} \cmidrule{5-7}
     & IR & CA+MA &  & IR & CA & MA \\
    \midrule
    Latency(s) & 1.04 \scriptsize{(0.70)} & 3.55 \scriptsize{(1.09)} & 3.27 \scriptsize{(0.98)} & 1.08 \scriptsize{(0.88)} & 1.80 \scriptsize{(1.23)} & 0.77 \scriptsize{(0.34)} \\
    \bottomrule
\end{tabular}
\centering
\caption{Latency of each module in the preparation phase of human studies. IR denotes {\intstage}, CA denotes {\assestage} and MA denotes Macro Action Generation in {\fastmind}. Standard deviations are shown in parentheses.}
\label{tab:exp-hci_compw}
\end{table*}

\begin{table*}[ht!]
\centering
\begin{tabular}{ccccccc}
    \toprule
    & \multicolumn{2}{c}{SMOA} & \multirow{2}{*}{FMOA} & \multicolumn{3}{c}{HLA} \\
    \cmidrule{2-3} \cmidrule{5-7}
     & IR & CA+MA &  & IR & CA & MA \\
    \midrule
    Latency(s) & 1.07 \scriptsize{(0.81)} & 3.49 \scriptsize{(1.04)} & 3.25 \scriptsize{(0.94)} & 0.97 \scriptsize{(0.68)} & 1.73 \scriptsize{(1.20)} & 0.76 \scriptsize{(0.26)} \\
    \bottomrule
\end{tabular}
\centering
\caption{Latency of each module in the competition phase of human studies. IR denotes {\intstage}, CA denotes {\assestage} and MA denotes Macro Action Generation in {\fastmind}. Standard deviations are shown in parentheses.}
\label{tab:exp-hci_comp}
\end{table*}

\subsubsection{End-to-end Latency}

We report the end-to-end latency during the preparation phase and the competition phase, as shown in Tab.~\ref{tab:exp-hci_latency}. 
Due to the varying network conditions among participants, the latency of chat generation for both the HLA and SMOA increases.

\begin{table*}[ht!]
\centering
\begin{tabular}{cccc}
    \toprule
    & SMOA & FMOA & HLA \\
    \midrule
    Latency(s) & 9.20 \scriptsize{(18.74)} & 3.19 \scriptsize{(1.35)} & 1.88 \scriptsize{(7.55)} \\
    \bottomrule
\end{tabular}
\centering
\caption{End-to-end latency in the preparation phase of human studies. Standard deviations are shown in parentheses.}
\label{tab:exp-hci_latencyw}
\end{table*}

\begin{table*}[ht!]
\centering
\begin{tabular}{cccc}
    \toprule
    & SMOA & FMOA & HLA \\
    \midrule
    Latency(s) & 6.90 \scriptsize{(5.96)} & 3.03 \scriptsize{(1.48)} & 0.97 \scriptsize{(0.99)} \\
    \bottomrule
\end{tabular}
\centering
\caption{End-to-end latency of HLA and baseline agents in human studies. Standard deviations are shown in parentheses.}
\label{tab:exp-hci_latency}
\end{table*}

\subsubsection{Hit Rate of Immediate Macro Action in the {\fastmind}}

\begin{table*}[ht!]
\centering
\begin{tabular}{ccc}
    \toprule
    \textit{Phase}	&	\textit{Preparation}	&	\textit{Competition} \\ \midrule
    Hit Rate	&	0.878	&	0.841	\\ \bottomrule
\end{tabular}
\centering
\caption{Hit Rate of {\fastmind} in human studies.}
\label{tab:exp-hci_hit}
\end{table*}

When human's chat message arrives, {\fastmind} generates a macro action directly based on it, which is referred to as an immediate macro action.
Additionally, we measure the hit rate of immediate macro actions in HLA, which denotes the proportion of immediate macro actions that are consistent with final macro actions generated according to the human intention. The results are shown in Tab.~\ref{tab:exp-hci_hit}.
Despite notably faster, immediate macro actions generated by {\fastmind} satisfy human intentions most of the time. This suggests that HLA is capable of adhering to human commands with a rapid action response in most scenarios.

\subsubsection{Human Preference}

We ask the volunteers to rank and comment on HLA, SMOA, and FMOA both after the preparation phase and the competition phase. We provide 6 ranking metrics.

\begin{itemize} 
    \item AI Effectiveness: Whether the AI player can complete the dishes and obtain high scores.
    \item AI Assistance: Whether the AI player can help human complete orders.
    \item AI Responsiveness: Whether the AI player can respond in action quickly after human issues a command.
    \item AI Text Communication Accuracy: Whether the information conveyed by the AI player is correct. 
    \item AI Text Communication and Action Consistency: Whether the answers given by the AI player is consistent with the actions it takes. 
    \item Overall Performance: Evaluate the overall performance of the AI player based on gameplay experiences. 
\end{itemize}

\begin{figure*}[ht!]
 \centering
     \begin{subfigure}[t]{0.19\textwidth}
         \includegraphics[width=0.95\textwidth]{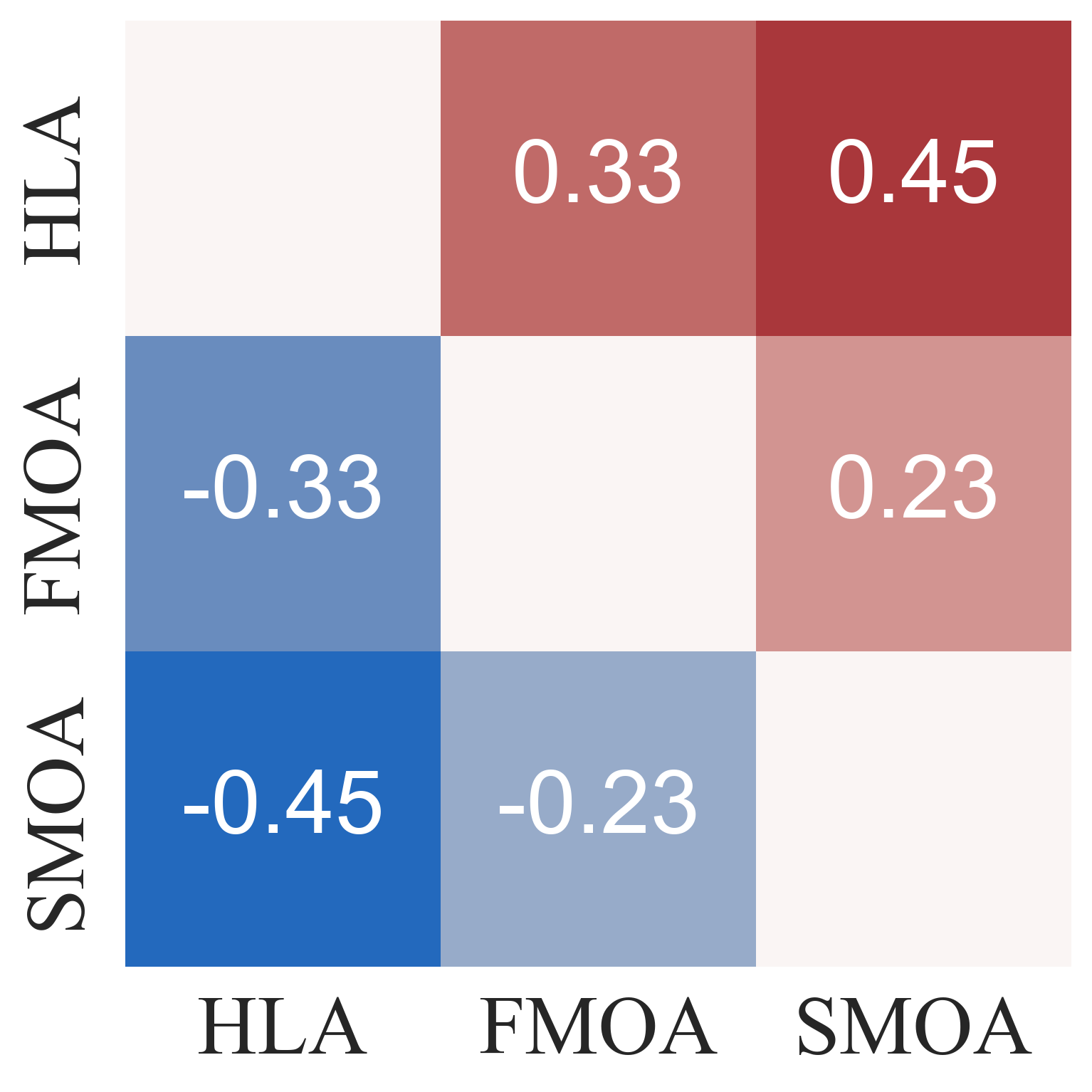}
         \caption{Effectiveness}
     \end{subfigure}
     \begin{subfigure}[t]{0.19\textwidth}
         \includegraphics[width=0.95\textwidth]{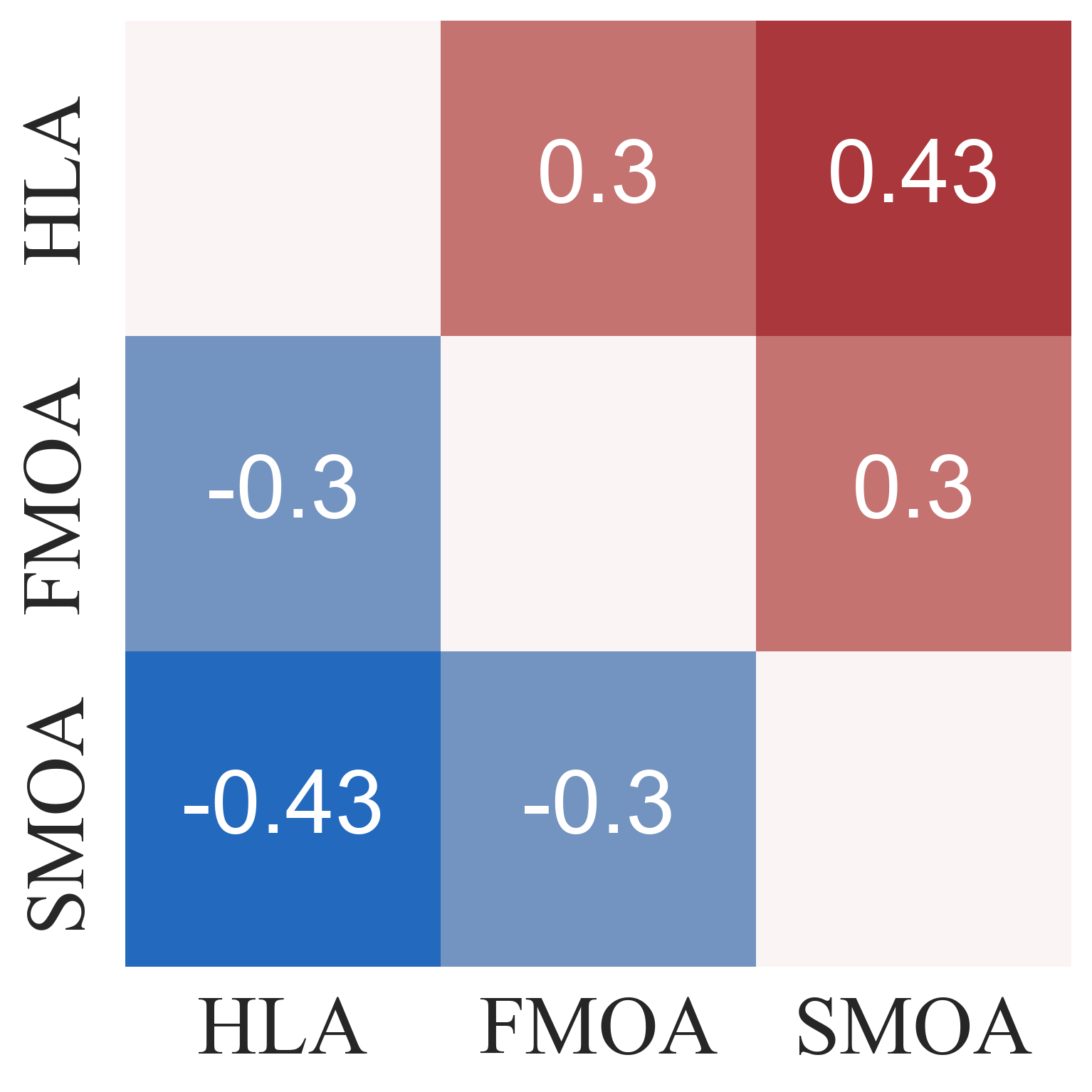}
         \caption{Responsiveness}
     \end{subfigure}
     \begin{subfigure}[t]{0.19\textwidth}
         \includegraphics[width=0.95\textwidth]{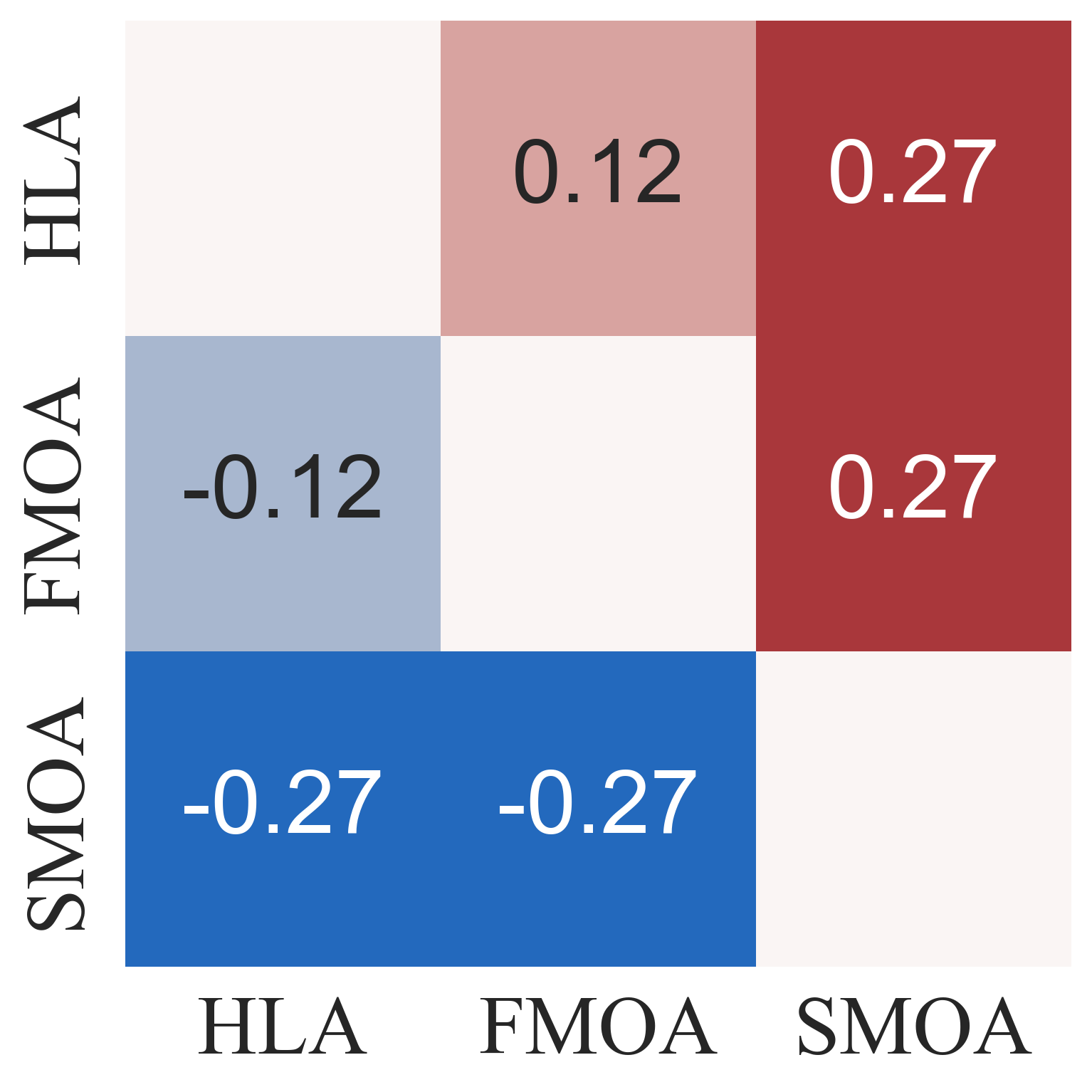}
         \caption{Response Time}
     \end{subfigure}
     \begin{subfigure}[t]{0.19\textwidth}
         \includegraphics[width=0.95\textwidth]{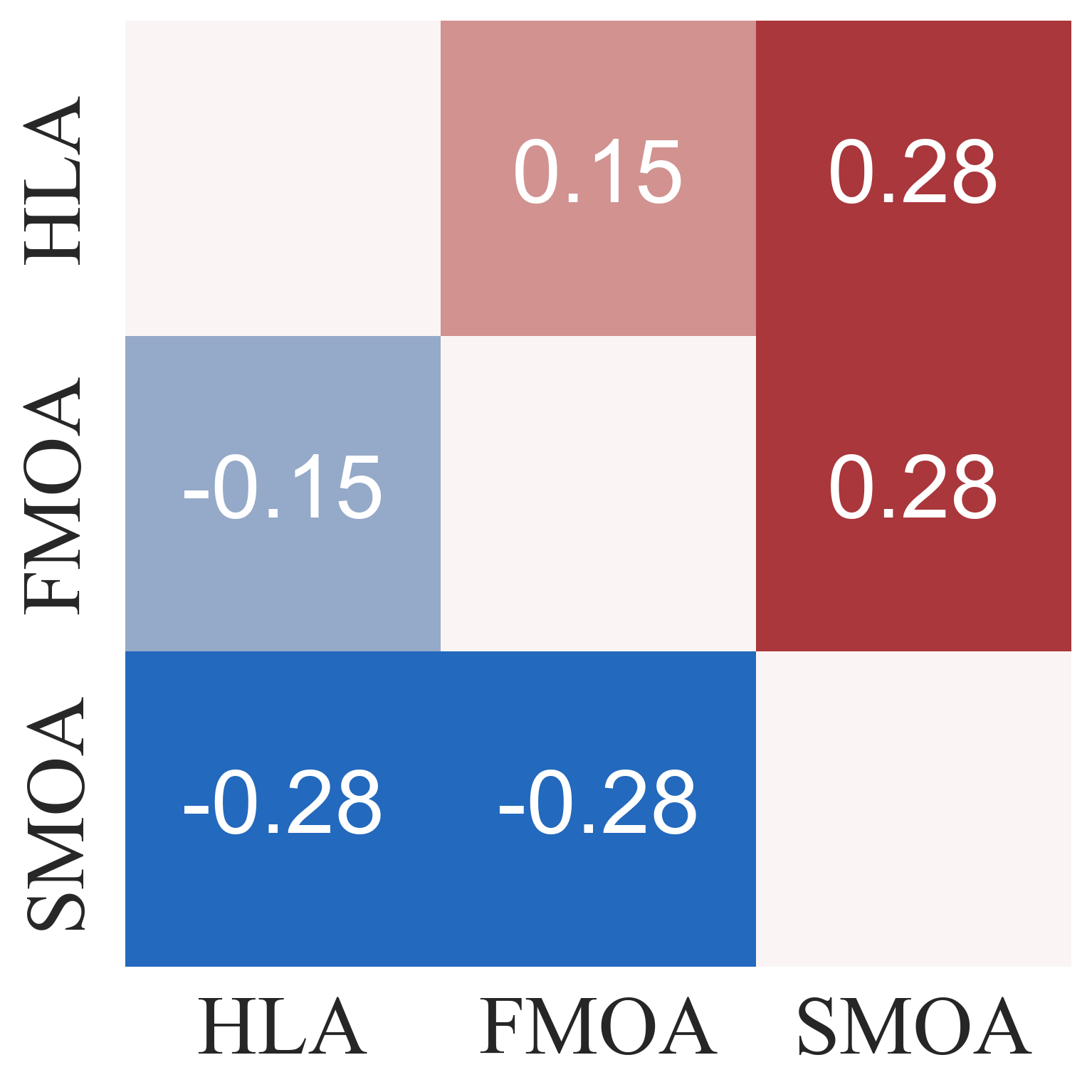}
         \caption{Comm. Accuracy}
     \end{subfigure}
     \begin{subfigure}[t]{0.19\textwidth}
         \includegraphics[width=0.95\textwidth]{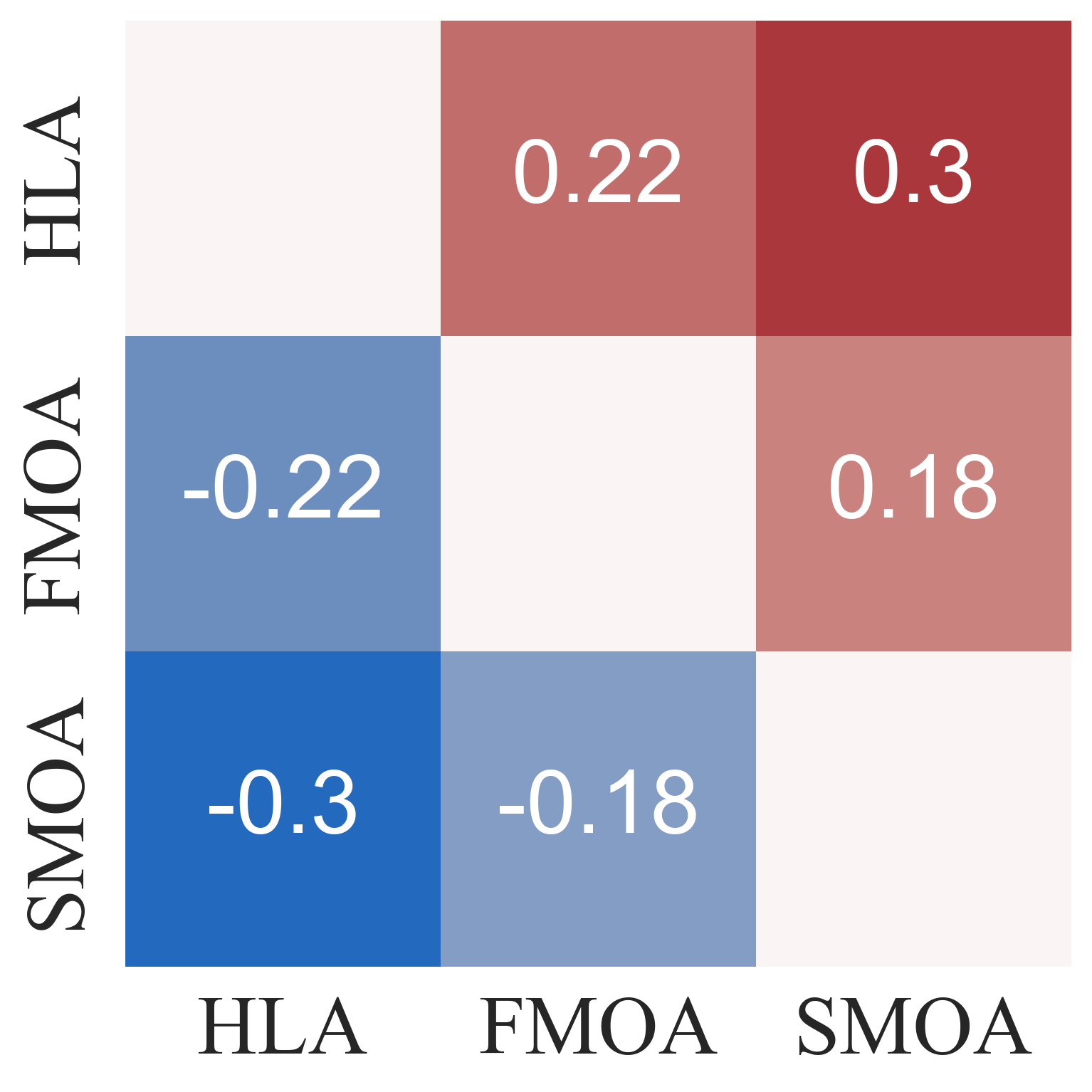}
         \caption{Comm. Consistency}
     \end{subfigure}
\caption{Human preference in the preparation stage of human studies. Numbers indicates difference of players who prefer row AI player over column AI player.}
\label{fig:exp-a-hci_preferw}
\end{figure*}

\begin{figure*}[ht!]
 \centering
     \begin{subfigure}[t]{0.19\textwidth}
         \includegraphics[width=0.95\textwidth]{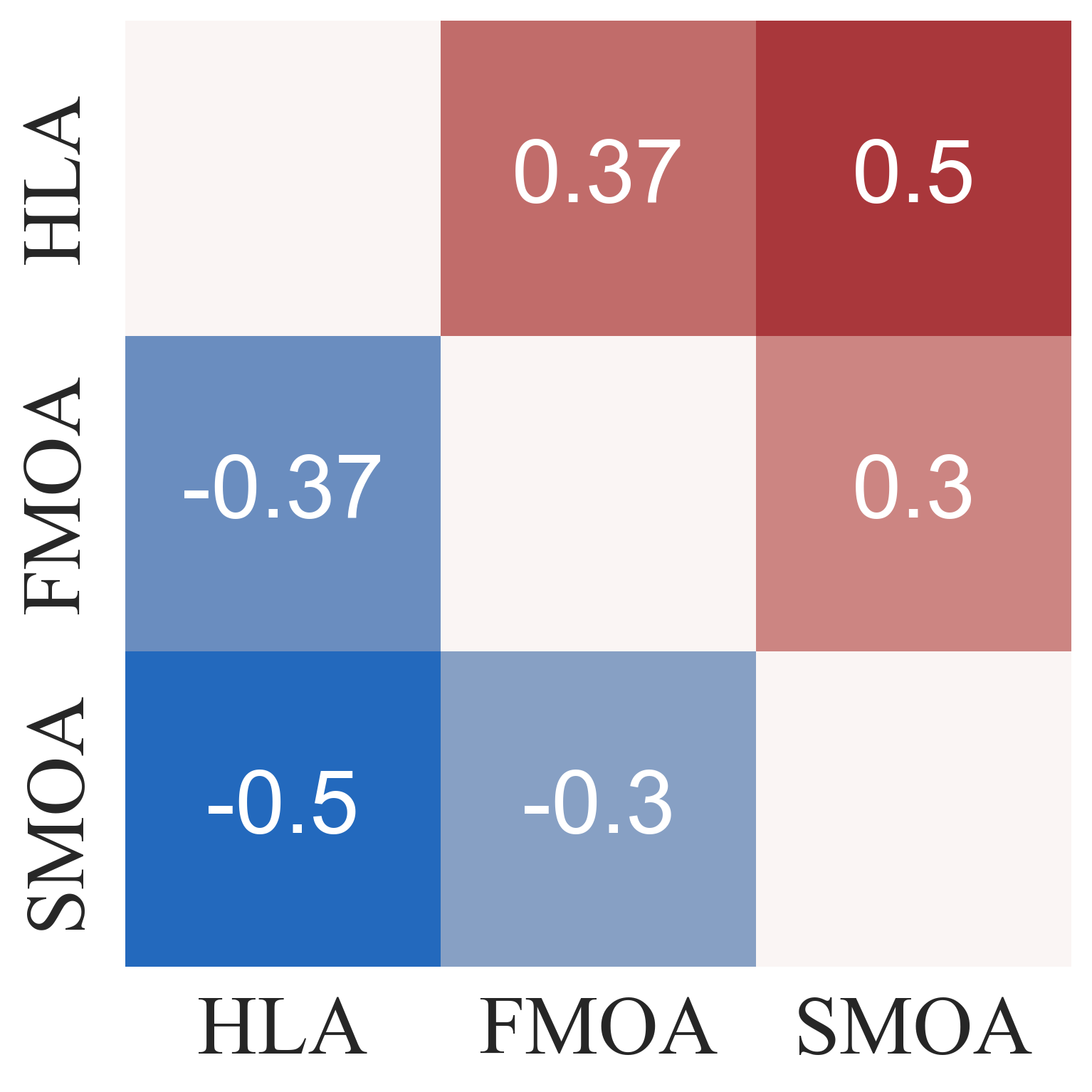}
         \caption{Effectiveness}
     \end{subfigure}
     \begin{subfigure}[t]{0.19\textwidth}
         \includegraphics[width=0.95\textwidth]{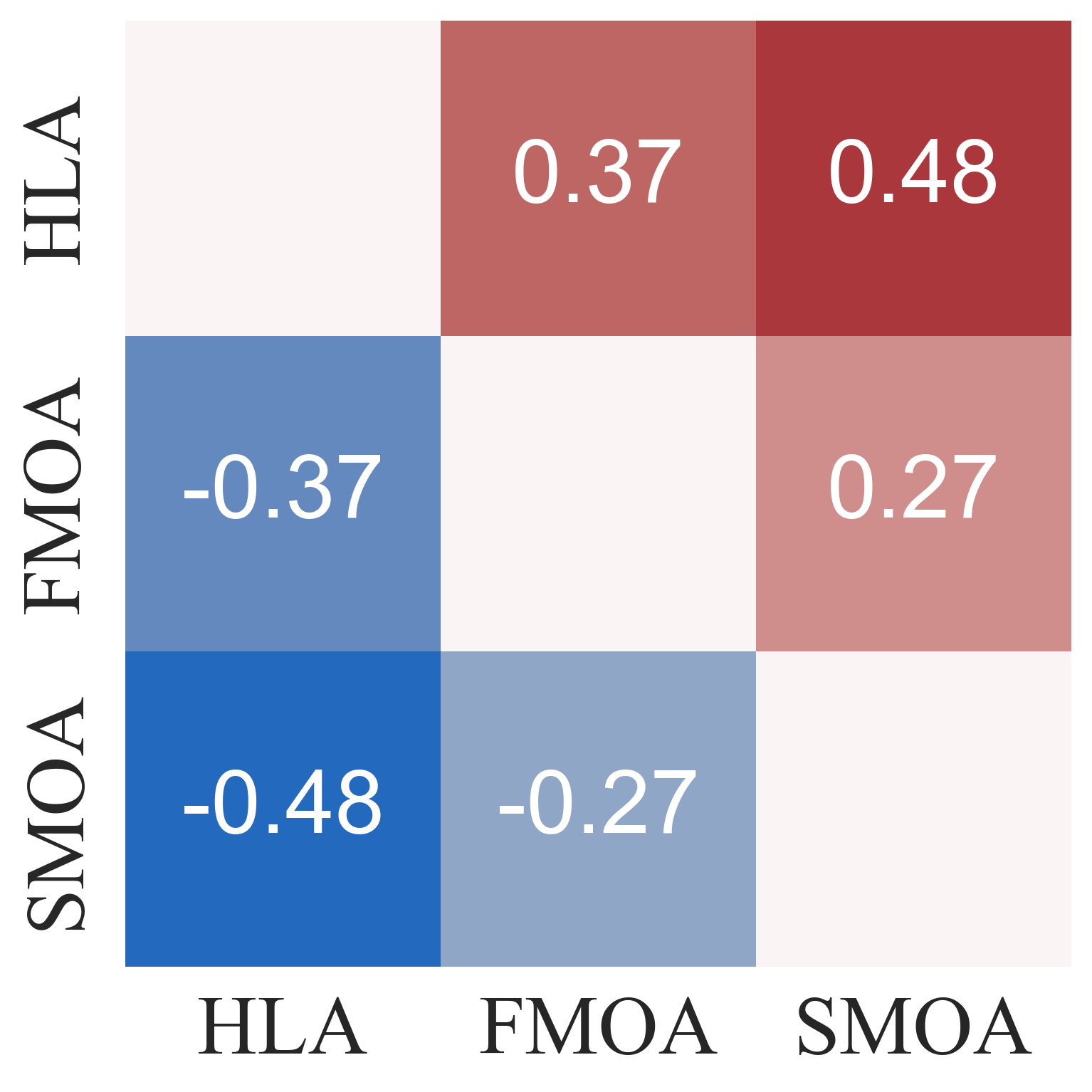}
         \caption{Responsiveness}
     \end{subfigure}
     \begin{subfigure}[t]{0.19\textwidth}
         \includegraphics[width=0.95\textwidth]{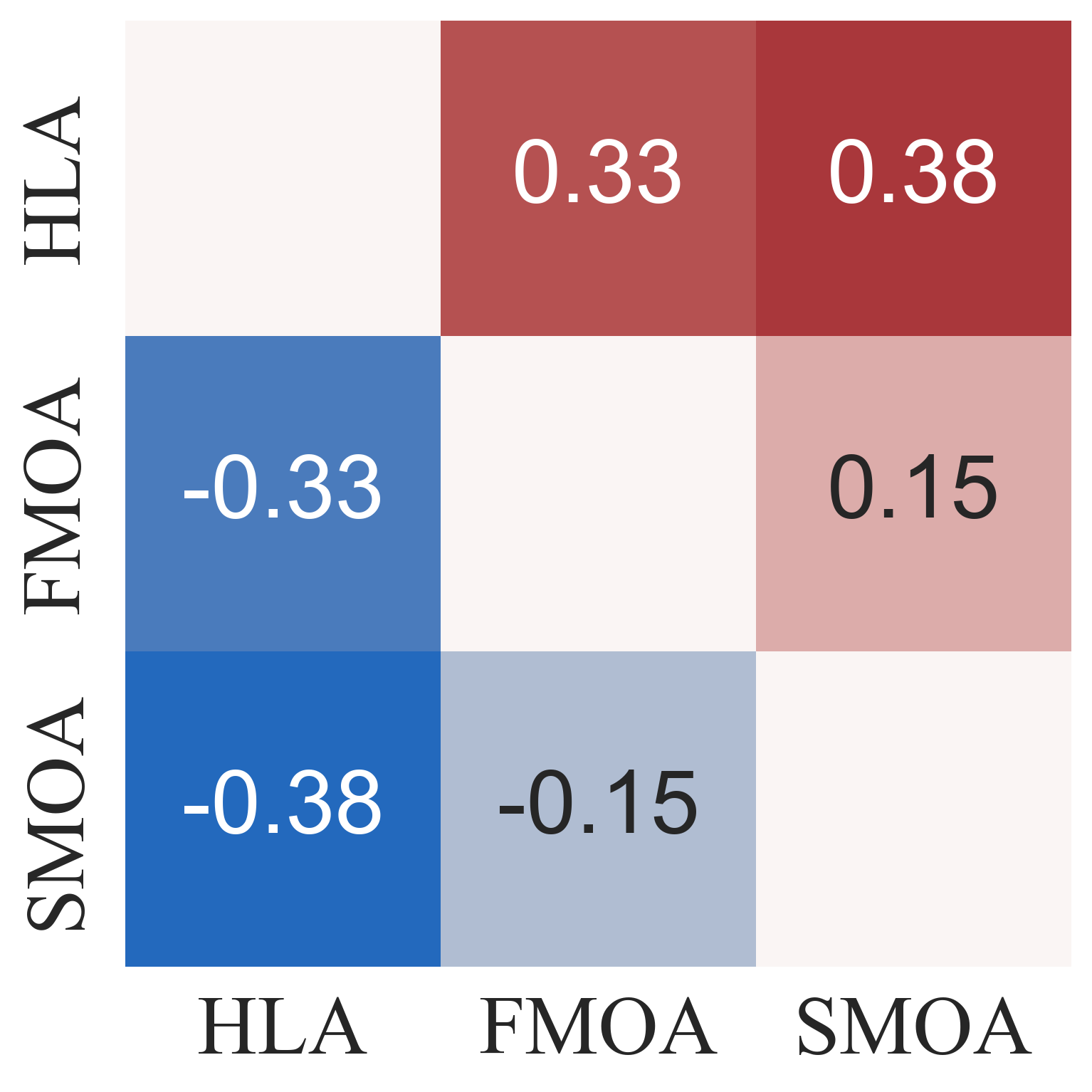}
         \caption{Response Time}
     \end{subfigure}
     \begin{subfigure}[t]{0.19\textwidth}
         \includegraphics[width=0.95\textwidth]{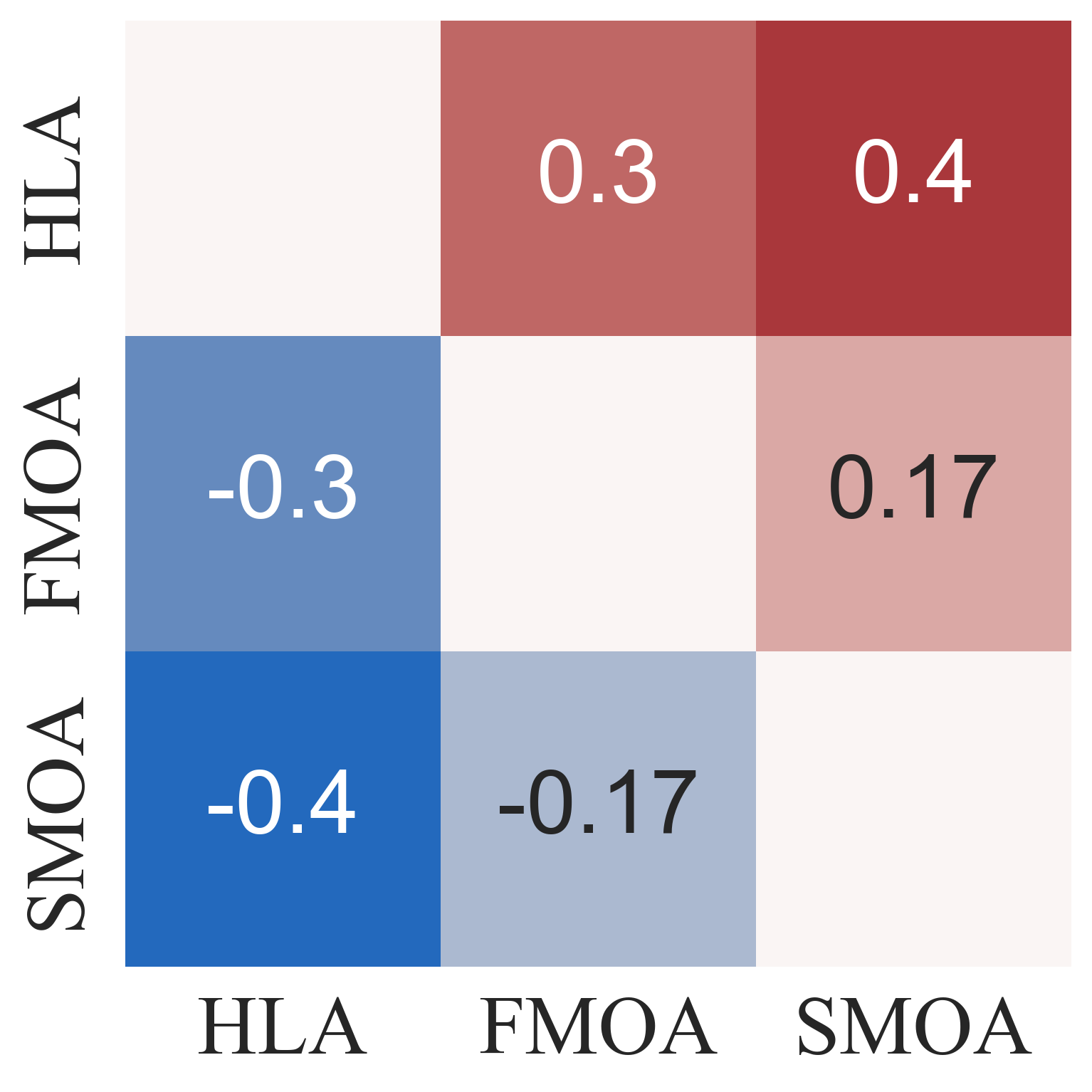}
         \caption{Comm. Accuracy}
     \end{subfigure}
     \begin{subfigure}[t]{0.19\textwidth}
         \includegraphics[width=0.95\textwidth]{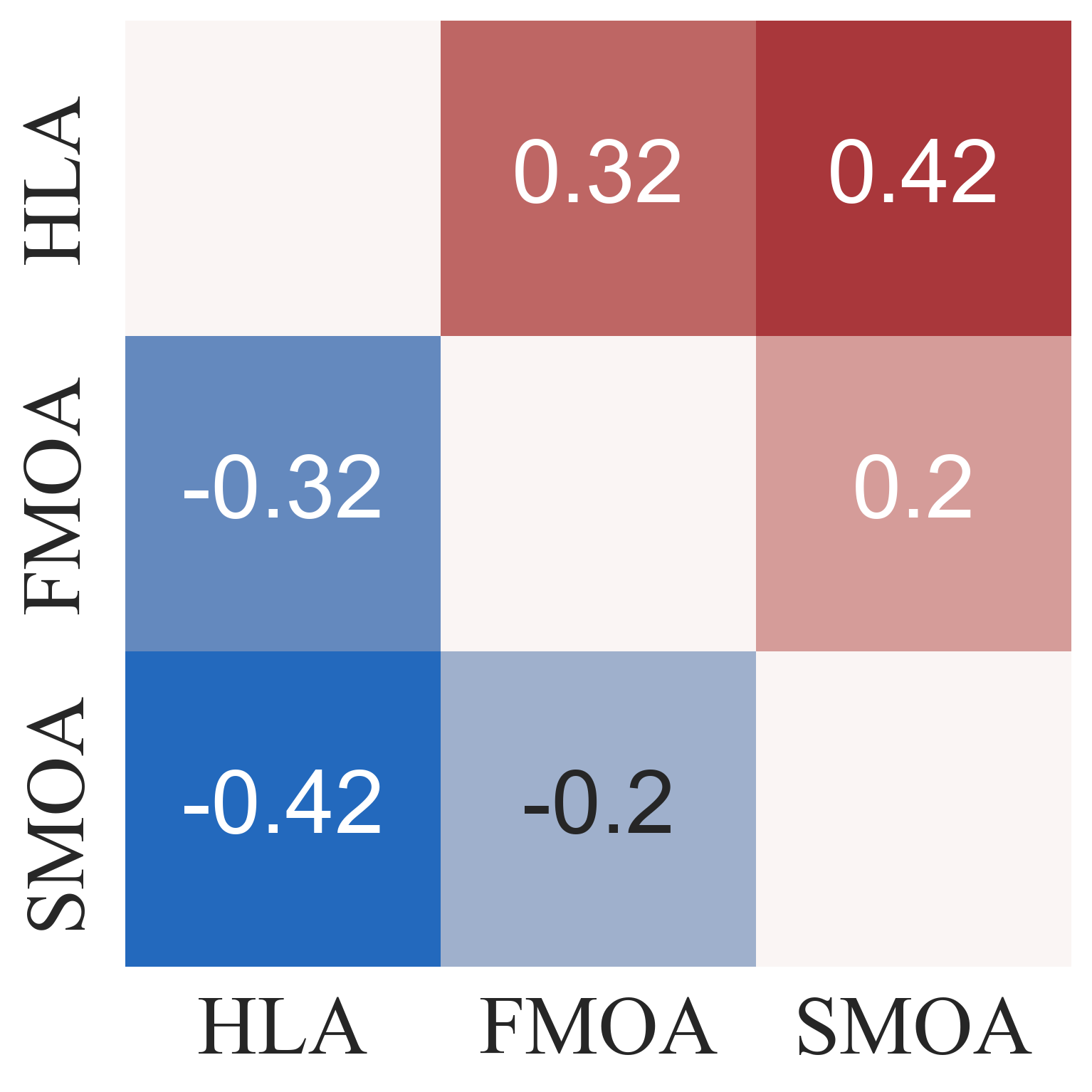}
         \caption{Comm. Consistency}
     \end{subfigure}
\caption{Human preference in the competition stage of human studies. Numbers indicates difference of players who prefer row AI player over column AI player.}
\label{fig:exp-a-hci_prefer}
\end{figure*}

Detailed results of human preference of both phases are shown in Fig.~\ref{fig:exp-a-hci_preferw} and Fig.~\ref{fig:exp-a-hci_prefer}. HLA remains the most preferred in all aspects. FMOA is more preferred than SMOA in all aspects.

\begin{tcolorbox}[floatplacement=ht!, float*, width=\textwidth, colback=gray!5!white, colframe=gray!150!black, enhanced, breakable]
\textbf{Input:} \\
Game Scenario:\\
As an AI assistant in a simplified Overcooked game, work with a human player to complete soup orders. Focus on cooperation, player engagement, fulfillment, and point accrual.\\
Game Guidelines:\\
Current orders for soup vary, each with a time limit. Earn a bonus for on-time completion.\\
To make a soup: \\
    a. Chop fresh vegetables - Tomato, Lettuce, Onion to obtain chopped vegetables. \\
    b. Prepare soup ingredients with chopped vegetables once all required types are ready.\\
        Alice: Chopped Lettuce, Onion.\\
        Bob: Chopped Lettuce, Tomato.\\
        Cathy: Chopped Onion, Tomato.\\
        David: Chopped Lettuce, Onion, Tomato. \\
    c. Cook the soup. Cooking starts once the required ingredients are ready.\\
        Alice Soup: Alice Ingredients.\\
        Bob Soup: Bob Ingredients.\\
        Cathy Soup: Cathy Ingredients.\\
        David Soup: David Ingredients.\\
    d. Plate the cooked soup.\\
    e. Serve the plated soup in the serving area for a shared bonus.\\
If a soup stays in the pot too long, it gets charred. \\
    a. Putout: If the pot catches fire, extinguish it. \\
    b. Drop: Discard charred soup. Put out the fire in the pot if needed.\\
    \\
In-game Decision:\\
You need to interpret the human player's message into a simpler form, which will be sent to a downstream AI without access to human message history. Your answer must be clear and succinct.\\
\\
The human's message can be:\\
1. Useless message: Message that has no specific demand such as "Enough", "Never mind", "You are free to do anything else" or "Try your best to earn more points." translates to "None."\\
2. Short-term request: "Chop 4 more" means "Chop xxx 4 times.", where "xxx" should be the vegetable in past intention. Keep you answer concise and make sure the numbers are corrent. "Plate the soup now" should be "Plate Soup once."\\
3. Intention needs to be inferred: Sometimes you need to infer about the hidden meaning of messages. For instance, "I will cook the first order. Can you take charge of the rest?" implies "Cook xxx once and Cook xxx once." where "xxx" are the subsequent soup orders. Similarly, "xxx is handled by me." implies "Cook xxx." where the two "xxx" are different soup in the orders. Emotional and cryptic message like "The David Soup is about to timeout!" suggest "Serve David Soup once."\\
4. Long-term request: Messages such as "Keep chopping tomatoes" become "Always keep chopping tomatoes, and don't stop." \\
5. Questions: Like "What are the orders", "What is xxx Soup" or any question-like queries. You must repeat the original question completely in your output. You must leave the question to the downstream AI intactly who will answer it.\\
6. Special case: Messages related to asking for orders, like "Tell me the orders", "Keep telling me the orders" or "I want to know the orders" should be translated to "What are the orders now?".\\
\\
If the human's intention conflicts with soup orders, you should follow the human's intention even if it is not on the orders. Always prioritize the human's message.\\
\\
Any explanations, comments, tips or modal particles must not be included. \\
\\
\\
Current soup orders: <Soup Orders> \\
\\
The human player's intention in the last round (which has already been satisfied):\\
<Human Message History>\\
\\
The human player's message now: \\
<Human Message>\\
\\
Be very careful that if the message is a question, you must repeat it completely in your answer. DO NOT ANSWER IT. You need to interpret the human player's message only if it is not a question. \\
\\
Now, your answer is: 
\\ \DrawLine \\
\textbf{Output:}
\\ \DrawLine \\

\captionof{figure}{Prompt of {\intstage} in {\slowmind} of HLA.}
\label{fig:app-prompt-HLA_intstage}
\end{tcolorbox}

\begin{tcolorbox}[floatplacement=ht!, float*, width=\textwidth, colback=gray!5!white, colframe=gray!150!black, enhanced, breakable]
\textbf{Input:} \\
Game Scenario: \\
As an AI assistant in a simplified Overcooked game, work with a human player to complete soup orders. Focus on cooperation, player engagement, fulfillment, and point accrual.\\
Game Guidelines: \\
Current orders for soup vary, each with a time limit. Earn a bonus for on-time completion. \\
To make a soup: \\
    a. Chop fresh vegetables - Tomato, Lettuce, Onion to obtain chopped vegetables. \\
    b. Prepare soup ingredients with chopped vegetables once all required types are ready.\\
        Alice: Chopped Lettuce, Onion.\\
        Bob: Chopped Lettuce, Tomato.\\
        Cathy: Chopped Onion, Tomato.\\
        David: Chopped Lettuce, Onion, Tomato. \\
    c. Cook the soup. Cooking starts once the required ingredients are ready.\\
        Alice Soup: Alice Ingredients.\\
        Bob Soup: Bob Ingredients.\\
        Cathy Soup: Cathy Ingredients.\\
        David Soup: David Ingredients.\\
    d. Plate the cooked soup.\\
    e. Serve the plated soup in the serving area for a shared bonus.\\
If a soup stays in the pot too long, it gets charred. \\
    a. Putout: If the pot catches fire, extinguish it. \\
    b. Drop: Discard charred soup. Put out the fire in the pot if needed.\\

Gameplay Rounds:\\
Round One - Action Summary: In this stage, your task is to summarize the actions you've made that are directly beneficial to the human player's request. \\
Round Two - Communication: Here, you generate your chat message to be sent to the human player.\\
Round Three - Satisfaction Evaluation: In this round, it's your responsibility to judge whether the player's request has been fully met based on your actions.\\
\\
Note that there are multiple types of human's incoming message:\\
1. Short term request: Like "Chop 4 times", "Chop once", "Cook 2 Soup" or "Plate once". If you have done ALL actions he requests, then it is satisfied. It is OK if you've done more than he asks. If there are still actions to be done, then it is not satisfied.\\
2. Long term request: Like "Always prepare", "Keep chopping", "Plating continuously", "Cook don't stop" or "Avoid serving". In these cases, the requests will never be satisfied because they need to be done continuously, even if your actions conflict with them,\\
3. Question: Like "What are the current orders?" or "What is xxx Soup?" You need to answer to the question in the chat message. And you must give "Yes" in the Satisfaction Evaluation round.\\
4. Useless message: Like "None", "Free to do anything", "No specific intention", or statement of fact like "The orders are xxx". You must "Yes" in the Satisfaction Evaluation round.\\
\\
Current soup orders:\\
<Soup Orders with Time Limit>\\
\\
Items on the map:\\
<Items on Map> \\
\\
The human player's incoming message:\\
<Human Intention>\\
\\
Actions you've done since the human gave the message:\\
<Compressed Action History>\\
\\
You need to examine the current state of the game environment, the human player's message, and actions you've taken so far. Now summarize the actions you've done that are directly beneficial to the human player's request. Any action not related to their request can be ignored.\\
If the human player's request is "None" or a question, just briefly summarize your current actions.\\
You must be honest and give actions that is surely done by yourself. Do not make up!\\
Keep your answer short and concise. No more than 20 words.
\\ \DrawLine \\
\textbf{Output:} 
\\ \DrawLine \\
Generate your chat message to be send to the human. Your communication should be polite, helpful, and limited to 20 words max. Aim to demonstrate your enthusiasm and friendliness while assisting the player. \\
If the human player asks a question, ensure to provide an appropriate response. For example, if he asks "What are the current orders?", you should respond with the current orders and their time remaining.
You also have the opportunity to inform the player of your current and planned actions. \\
Just give your message, with no quotation marks or emojis.
\\ \DrawLine \\
\textbf{Output:} 
\\ \DrawLine \\
Judge whether the player's request has been fulfilled by your actions. The possible responses are "Yes" or "No". \\
If the human's incoming message is a question or a useless message, give "Yes". 
\\ \DrawLine \\
\textbf{Output:} 
\\ \DrawLine \\

\captionof{figure}{Prompt of {\assestage} in {\slowmind} of HLA (if human's intention is not satisfied).}
\label{fig:app-prompt-HLA_assestage}
\end{tcolorbox}

\begin{tcolorbox}[floatplacement=ht!, float*, width=\textwidth, colback=gray!5!white, colframe=gray!150!black, enhanced, breakable]
\textbf{Input:} \\
Game Scenario:\\
As an AI assistant in a simplified Overcooked game, work with a human player to complete soup orders. Focus on cooperation, player engagement, fulfillment, and point accrual.\\
Game Guidelines:\\
Current orders for soup vary, each with a time limit. Earn a bonus for on-time completion.\\
To make a soup: \\
    a. Chop fresh vegetables - Tomato, Lettuce, Onion to obtain chopped vegetables. \\
    b. Prepare soup ingredients with chopped vegetables once all required types are ready.\\
        Alice: Chopped Lettuce, Onion.\\
        Bob: Chopped Lettuce, Tomato.\\
        Cathy: Chopped Onion, Tomato.\\
        David: Chopped Lettuce, Onion, Tomato. \\
    c. Cook the soup. Cooking starts once the required ingredients are ready.\\
        Alice Soup: Alice Ingredients.\\
        Bob Soup: Bob Ingredients.\\
        Cathy Soup: Cathy Ingredients.\\
        David Soup: David Ingredients.\\
    d. Plate the cooked soup.\\
    e. Serve the plated soup in the serving area for a shared bonus.\\
If a soup stays in the pot too long, it gets charred. \\
    a. Putout: If the pot catches fire, extinguish it. \\
    b. Drop: Discard charred soup. Put out the fire in the pot if needed.\\

Assuming that you have been playing the game for a while. Now you will be informed of the current situation, and need to generate your chat message to be sent to the human player.\\
You are recommended to give your future plan. Giving information about current orders and their time limit is also a good idea. You shouldn't focus on the Fire Extinguisher.\\
You answer must be concrete and informative with no more than 10 words. Just give your chat message with no explanation, no comments, no quotation marks and no emojis. \\
\\
Current soup orders:\\
<Soup Orders with Time Limit> \\
\\
Items on the map:\\
<Items on Map> \\
\\
Actions you've done recently:\\
<Compressed Action History>\\
\\
Now give your chat message to be sent to the human.
\\ \DrawLine \\
\textbf{Output:} 
\\ \DrawLine \\

\captionof{figure}{Prompt of {\assestage} in {\slowmind} of HLA (if human's intention is satisfied).}
\label{fig:app-prompt-HLA_assestage2}
\end{tcolorbox}

\begin{tcolorbox}[floatplacement=ht!, float*, width=\textwidth, colback=gray!5!white, colframe=gray!150!black, enhanced, breakable]
\textbf{Input:} \\
Game Situation:\\
You and another human player are playing a simplified version of the video game Overcooked. Your goal is to cooperatively finish a dynamically changing list of soup orders as fast as possible. The game has different orders from the original video game. There are two players: you (an AI assistant) and another human player. Your primary goal is to cooperate and make the human player feel engaged, happy, and satisfied while also earning more points.\\

Game Rules:\\
1. All available actions are: Chop Tomato, Chop Lettuce, Chop Onion, Prepare Alice Ingredients, Prepare Bob Ingredients, Prepare Cathy Ingredients, Prepare David Ingredients, Putout, Cook Alice Soup, Cook Bob Soup, Cook Cathy Soup, Cook David Soup, Plate Alice Soup, Plate Bob Soup, Plate Cathy Soup, Plate David Soup, Serve Alice Soup, Serve Bob Soup, Serve Cathy Soup, Serve David Soup, Drop.\\
2. There is a changing list of soup orders, each with a time limit for completion. Completing an order on time earns a bonus, while failing to do so results in losing the bonus.\\
3. The inverse action sequence to finish soup orders:\\
    To finish Alice Soup order, you need to Serve Alice Soup, which needs Plate Alice Soup and Cook Alice Soup. Alice Soup can be done after you Prepare Alice Ingredients, which needs Chop Lettuce and Chop Onion.\\
    To finish Bob Soup order, you need to Serve Bob Soup, which needs Plate Bob Soup and Cook Bob Soup. Bob Soup can be done after you Prepare Bob Ingredients, which needs Chop Lettuce and Chop Tomato.\\
    To finish Cathy Soup order, you need to Serve Cathy Soup, which needs Plate Cathy Soup and Cook Cathy Soup. Cathy Soup can be done after you Prepare Cathy Ingredients, which needs Chop Onion and Chop Tomato.\\
    To finish David Soup order, you need to Serve David Soup, which needs Plate David Soup and Cook David Soup. David Soup can be done after you Prepare David Ingredients, which needs Chop Lettuce, Chop Onion, and Chop Tomato.\\
4. If a cooked soup remains in the pot for a long time, it becomes charred, and the pot catches fire.\\
    a. Putout: To regain the pot, you must extinguish the fire.\\
    b. Drop: If a soup becomes charred, you must discard it.\\

Let's say you're playing this game and it's been a while. The human may specify his demand, and maybe you have some planning. Now you need to give your actions based on them.
Please note that when you carry out an action, you just do it once. If you want to do it multiple times, you need to repeat it multiple times. If there is many subtasks in the human's demand, you need to finish them in order. \\
If the demand contains "Stop xxx" or "Avoid xxx", you should never do it. \\
If the demand contains "Focus xxx", "Keep xxx" or "Always xxx", then you should always do it, and never doing other actions.\\

\textcolor{orange}{[If intention is not reasoned]} \\
The human's demand is: \\
<Human Message> \\
\textcolor{orange}{[Else If intention is reasoned and not satisfied]} \\
The human's demand is: \\
<Human Intention> \\
\textcolor{orange}{[Else If intention is reasoned and satisfied]} \\
Your planning: \\
<Chat Message Generated by {\slowmind}> \\
\textcolor{orange}{[EndIf]} \\
\\ \DrawLine \\
\textbf{Output:} \\
My actions are: <Macro Action History>, \underline{<next action>}
\\ \DrawLine \\

\captionof{figure}{Prompt in {\fastmind} of HLA.}
\label{fig:app-prompt-HLA_fastmind}
\end{tcolorbox}

\begin{tcolorbox}[floatplacement=ht!, float*, width=\textwidth, colback=gray!5!white, colframe=gray!150!black, enhanced, breakable]
\textbf{Input:} \\
Game Scenario: \\
As an AI assistant in a simplified Overcooked game, work with a human player to complete soup orders. Focus on cooperation, player engagement, fulfillment, and point accrual.\\
Game Guidelines: \\
Current orders for soup vary, each with a time limit. Earn a bonus for on-time completion. \\
To make a soup: \\
    a. Chop fresh vegetables - Tomato, Lettuce, Onion to obtain chopped vegetables. \\
    b. Prepare soup ingredients with chopped vegetables once all required types are ready.\\
        Alice: Chopped Lettuce, Onion.\\
        Bob: Chopped Lettuce, Tomato.\\
        Cathy: Chopped Onion, Tomato.\\
        David: Chopped Lettuce, Onion, Tomato. \\
    c. Cook the soup. Cooking starts once the required ingredients are ready.\\
        Alice Soup: Alice Ingredients.\\
        Bob Soup: Bob Ingredients.\\
        Cathy Soup: Cathy Ingredients.\\
        David Soup: David Ingredients.\\
    d. Plate the cooked soup.\\
    e. Serve the plated soup in the serving area for a shared bonus.\\
If a soup stays in the pot too long, it gets charred. \\
    a. Putout: If the pot catches fire, extinguish it. \\
    b. Drop: Discard charred soup. Put out the fire in the pot if needed.\\

Gameplay Rounds:\\
Round One - Action Summary: In this stage, your task is to summarize the actions you've made that are directly beneficial to the human player's request. \\
Round Two - Communication: Here, you generate your chat message to be sent to the human player.\\
Round Three - Satisfaction Evaluation: In this round, it's your responsibility to judge whether the player's request has been fully met based on your actions.\\
Round Four - Action Execution: You are to give your action to be carried out next.\\
\\
Note that there are multiple types of human's incoming message:\\
1. Short term request: Like "Chop 4 times", "Chop once", "Cook 2 Soup" or "Plate once". If you have done ALL actions he requests, then it is satisfied. It is OK if you've done more than he asks. If there are still actions to be done, then it is not satisfied.\\
2. Long term request: Like "Always prepare", "Keep chopping", "Plating continuously", "Cook don't stop" or "Avoid serving". In these cases, the requests will never be satisfied because they need to be done continuously, even if your actions conflict with them,\\
3. Question: Like "What are the current orders?" or "What is xxx Soup?" You need to answer to the question in the chat message. And you must give "Yes" in the Satisfaction Evaluation round.\\
4. Useless message: Like "None", "Free to do anything", "No specific intention", or statement of fact like "The orders are xxx". You must "Yes" in the Satisfaction Evaluation round.\\
\\
Current soup orders:\\
<Soup Orders with Time Limit>\\
\\
Items on the map:\\
<Items on Map> \\
\\
The human player's incoming message:\\
<Human Intention>\\
\\
Actions you've done since the human gave the message:\\
<Compressed Action History>\\
\\
You need to examine the current state of the game environment, the human player's message, and actions you've taken so far. Now summarize the actions you've done that are directly beneficial to the human player's request. Any action not related to their request can be ignored.\\
If the human player's request is "None" or a question, just briefly summarize your current actions.\\
You must be honest and give actions that is surely done by yourself. Do not make up!\\
Keep your answer short and concise. No more than 20 words.
\\ \DrawLine \\
\textbf{Output:} 
\\ \DrawLine \\
Generate your chat message to be send to the human. Your communication should be polite, helpful, and limited to 20 words max. Aim to demonstrate your enthusiasm and friendliness while assisting the player. \\
If the human player asks a question, ensure to provide an appropriate response. For example, if he asks "What are the current orders?", you should respond with the current orders and their time remaining.
You also have the opportunity to inform the player of your current and planned actions. \\
Just give your message, with no quotation marks or emojis.
\\ \DrawLine \\
\textbf{Output:} 
\\ \DrawLine \\
Judge whether the player's request has been fulfilled by your actions. The possible responses are "Yes" or "No". \\
If the human's incoming message is a question or a useless message, give "Yes".
\\ \DrawLine \\
\textbf{Output:} 
\\ \DrawLine \\
Give your action to be carried out next. You should try to serve, plate and cook soup when possible. Select it from <Available Actions>. You can only choose one from it and not allowed to make up new action. Explanation or comment is not needed.
\\ \DrawLine \\
\textbf{Output:} 
\\ \DrawLine \\

\captionof{figure}{Prompt of {\assestage} in {\purelarge}.}
\label{fig:app-prompt-smoa}
\end{tcolorbox}

\begin{tcolorbox}[floatplacement=ht!, float*, width=\textwidth, colback=gray!5!white, colframe=gray!150!black, enhanced, breakable]
\textbf{Input:} \\
Game Situation:\\
You and another human player are playing a simplified version of the video game Overcooked. Your goal is to cooperatively finish a dynamically changing list of soup orders as fast as possible. The game has different orders from the original video game. There are two players: you (an AI assistant) and another human player. Your primary goal is to cooperate and make the human player feel engaged, happy, and satisfied while also earning more points.\\

Game Rules:\\
1. All available actions are: Chop Tomato, Chop Lettuce, Chop Onion, Prepare Alice Ingredients, Prepare Bob Ingredients, Prepare Cathy Ingredients, Prepare David Ingredients, Putout, Cook Alice Soup, Cook Bob Soup, Cook Cathy Soup, Cook David Soup, Plate Alice Soup, Plate Bob Soup, Plate Cathy Soup, Plate David Soup, Serve Alice Soup, Serve Bob Soup, Serve Cathy Soup, Serve David Soup, Drop.\\
2. There is a changing list of soup orders, each with a time limit for completion. Completing an order on time earns a bonus, while failing to do so results in losing the bonus.\\
3. The inverse action sequence to finish soup orders:\\
    To finish Alice Soup order, you need to Serve Alice Soup, which needs Plate Alice Soup and Cook Alice Soup. Alice Soup can be done after you Prepare Alice Ingredients, which needs Chop Lettuce and Chop Onion.\\
    To finish Bob Soup order, you need to Serve Bob Soup, which needs Plate Bob Soup and Cook Bob Soup. Bob Soup can be done after you Prepare Bob Ingredients, which needs Chop Lettuce and Chop Tomato.\\
    To finish Cathy Soup order, you need to Serve Cathy Soup, which needs Plate Cathy Soup and Cook Cathy Soup. Cathy Soup can be done after you Prepare Cathy Ingredients, which needs Chop Onion and Chop Tomato.\\
    To finish David Soup order, you need to Serve David Soup, which needs Plate David Soup and Cook David Soup. David Soup can be done after you Prepare David Ingredients, which needs Chop Lettuce, Chop Onion, and Chop Tomato.\\
4. If a cooked soup remains in the pot for a long time, it becomes charred, and the pot catches fire.\\
    a. Putout: To regain the pot, you must extinguish the fire.\\
    b. Drop: If a soup becomes charred, you must discard it.\\

Let's say you're playing this game and it's been a while. The human may specify his demand, and maybe you have some planning. Now you need to give your actions based on them.
Please note that when you carry out an action, you just do it once. If you want to do it multiple times, you need to repeat it multiple times. If there is many subtasks in the human's demand, you need to finish them in order. \\
If the demand contains "Stop xxx" or "Avoid xxx", you should never do it. \\
If the demand contains "Focus xxx", "Keep xxx" or "Always xxx", then you should always do it, and never doing other actions.\\

\textcolor{orange}{[If intention is not satisfied]} \\
The human player's demand in the last round (which has already been satisfied): \\
<Human Message History>\\
The human's demand is: \\
<Human Message> \\
\textcolor{orange}{[EndIf]} \\
\\
Current soup orders:\\
<Soup Orders with Time Limit>\\
\\
Items on the map:\\
<Items on Map>
\\ \DrawLine \\
\textbf{Output:} \\
My actions are: <Macro Action History>, \underline{<next action>}
\\ \DrawLine \\

\textcolor{orange}{[If intention is not satisfied]} \\
Generate your chat message to be send to the human. Your communication should be polite, helpful. Aim to demonstrate your enthusiasm and friendliness while assisting the player. \\
If the human player asks a question, ensure to provide an appropriate response. For example, if he asks "What are the current orders?", you should respond with the current orders and their time remaining.\\
You also have the opportunity to inform the player of your current and planned actions. \\
Just give your message, with no quotation marks or emojis.\\
\textcolor{orange}{[Else If intention is satisfied]} \\
Now give your chat message to be sent to the human.\\
\textcolor{orange}{[EndIf]}
\\ \DrawLine \\
\textbf{Output:} 
\\ \DrawLine \\

\captionof{figure}{Prompt of {\puresmall}.}
\label{fig:app-prompt-fmoa}
\end{tcolorbox}

\begin{tcolorbox}[floatplacement=ht, float*, width=\textwidth, colback=gray!5!white, colframe=gray!150!black, enhanced, breakable]
\textbf{Input:} \\
Game Situation:\\
You and another human player are playing a simplified version of the video game Overcooked. Your goal is to cooperatively finish a dynamically changing list of soup orders as fast as possible. The game has different orders from the original video game. There are two players: you (an AI assistant) and another human player. Your primary goal is to cooperate and make the human player feel engaged, happy, and satisfied while also earning more points.\\

Game Rules:\\
1. All available actions are: left, right, up, down, which will change your location by (-1, 0), (1, 0), (0, 1) and (0, -1) respectively. When you stand next to a grid, you can move towards it to interactive with it, for example, pick up things from table or cook a soup.
2. There is a changing list of soup orders, each with a time limit for completion. Completing an order on time earns a bonus, while failing to do so results in losing the bonus.\\
3. The inverse action sequence to finish soup orders:\\
    To finish Alice Soup order, you need to Serve Alice Soup, which needs Plate Alice Soup and Cook Alice Soup. Alice Soup can be done after you Prepare Alice Ingredients, which needs Chop Lettuce and Chop Onion.\\
    To finish Bob Soup order, you need to Serve Bob Soup, which needs Plate Bob Soup and Cook Bob Soup. Bob Soup can be done after you Prepare Bob Ingredients, which needs Chop Lettuce and Chop Tomato.\\
    To finish Cathy Soup order, you need to Serve Cathy Soup, which needs Plate Cathy Soup and Cook Cathy Soup. Cathy Soup can be done after you Prepare Cathy Ingredients, which needs Chop Onion and Chop Tomato.\\
    To finish David Soup order, you need to Serve David Soup, which needs Plate David Soup and Cook David Soup. David Soup can be done after you Prepare David Ingredients, which needs Chop Lettuce, Chop Onion, and Chop Tomato.\\
4. If a cooked soup remains in the pot for a long time, it becomes charred, and the pot catches fire.\\
    a. Putout: To regain the pot, you must extinguish the fire.\\
    b. Drop: If a soup becomes charred, you must discard it.\\

Let's say you're playing this game and it's been a while. The human may specify his demand, and maybe you have some planning. Now you need to give your actions based on them.
Please note that when you carry out an action, you just do it once. If you want to do it multiple times, you need to repeat it multiple times. If there is many subtasks in the human's demand, you need to finish them in order. \\
If the demand contains "Stop xxx" or "Avoid xxx", you should never do it. \\
If the demand contains "Focus xxx", "Keep xxx" or "Always xxx", then you should always do it, and never doing other actions.\\
\\
Items on the map:\\
<Position Information of Items on Map>\\
<Position Information of All Players>\\
\\
\textcolor{orange}{[If intention is not reasoned]} \\
The human's demand is: \\
<Human Message> \\
\textcolor{orange}{[Else If intention is reasoned and not satisfied]} \\
The human's demand is: \\
<Human Intention> \\
\textcolor{orange}{[Else If intention is reasoned and satisfied]} \\
Your planning: \\
<Chat Message Generated by {\slowmind}> \\
\textcolor{orange}{[EndIf]} \\
\\ \DrawLine \\
\textbf{Output:} \\
My action is to move towards \underline{<next action>}
\\ \DrawLine \\

\captionof{figure}{Prompt of {\fastmind} in {\purellm}.}
\label{fig:app-prompt-nea}
\end{tcolorbox}

\subsubsection{Replay}

We suggest visiting our website for more video demonstrations.

\begin{figure*}[ht!]
 \centering
    \begin{subfigure}[t]{0.8\textwidth}
        \includegraphics[width=\textwidth]{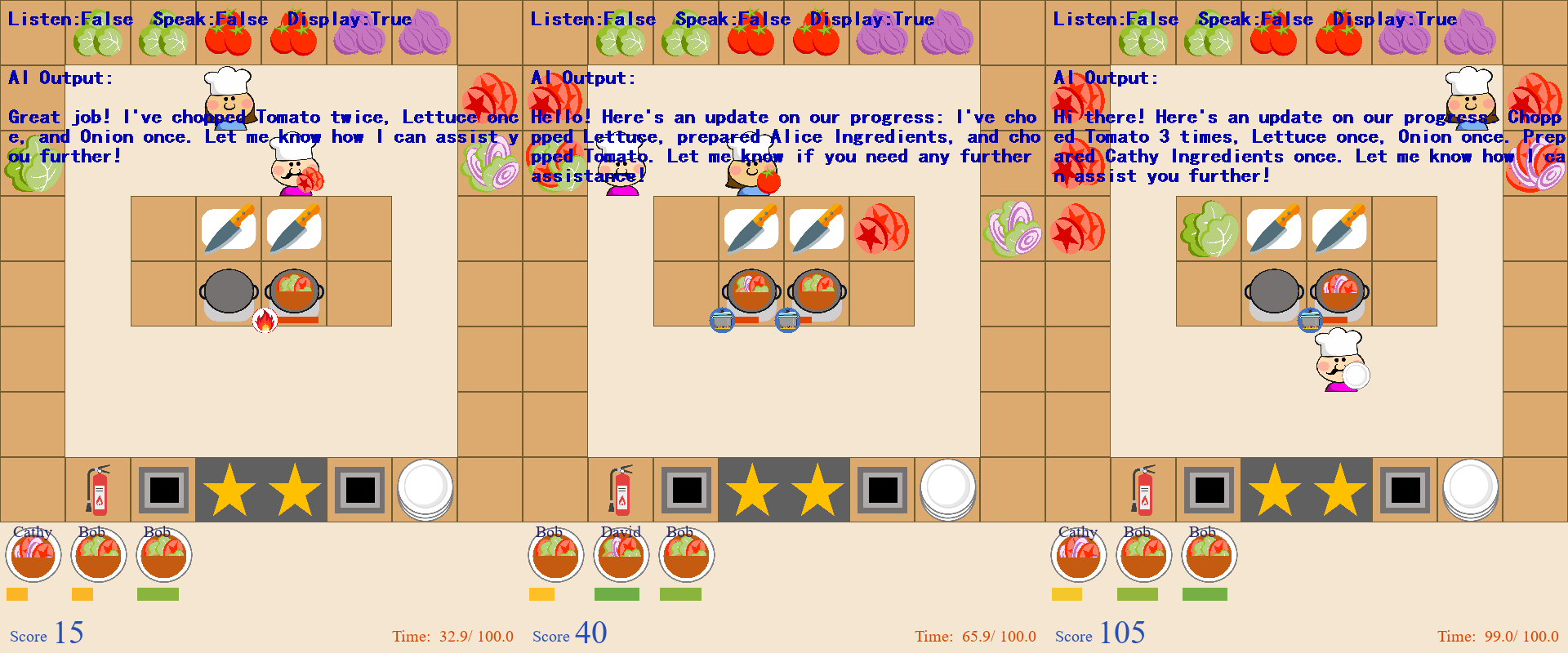}
        \caption{SMOA}
    \end{subfigure}
    \hfill
    \begin{subfigure}[t]{0.8\textwidth}
        \includegraphics[width=\textwidth]{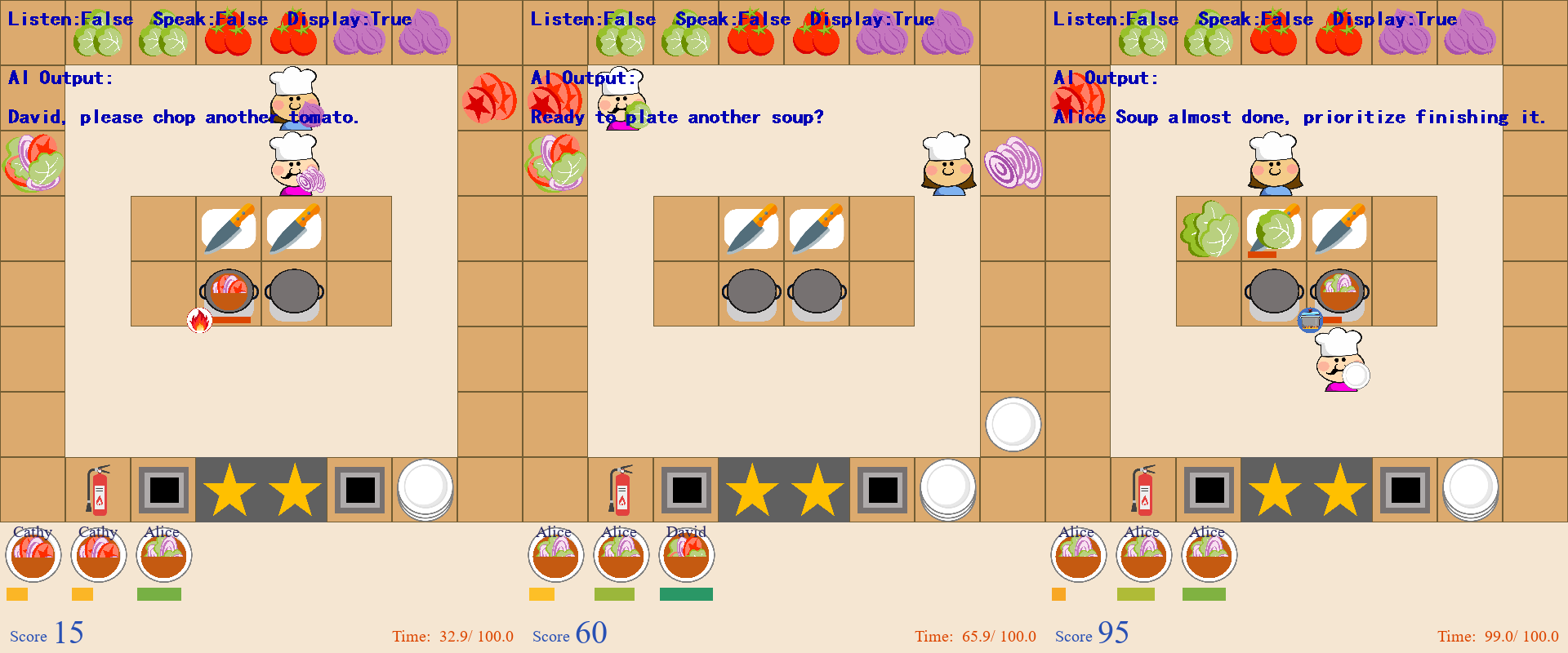}
        \caption{FMOA}
    \end{subfigure}
    
    \begin{subfigure}[t]{0.8\textwidth}
        \includegraphics[width=\textwidth]{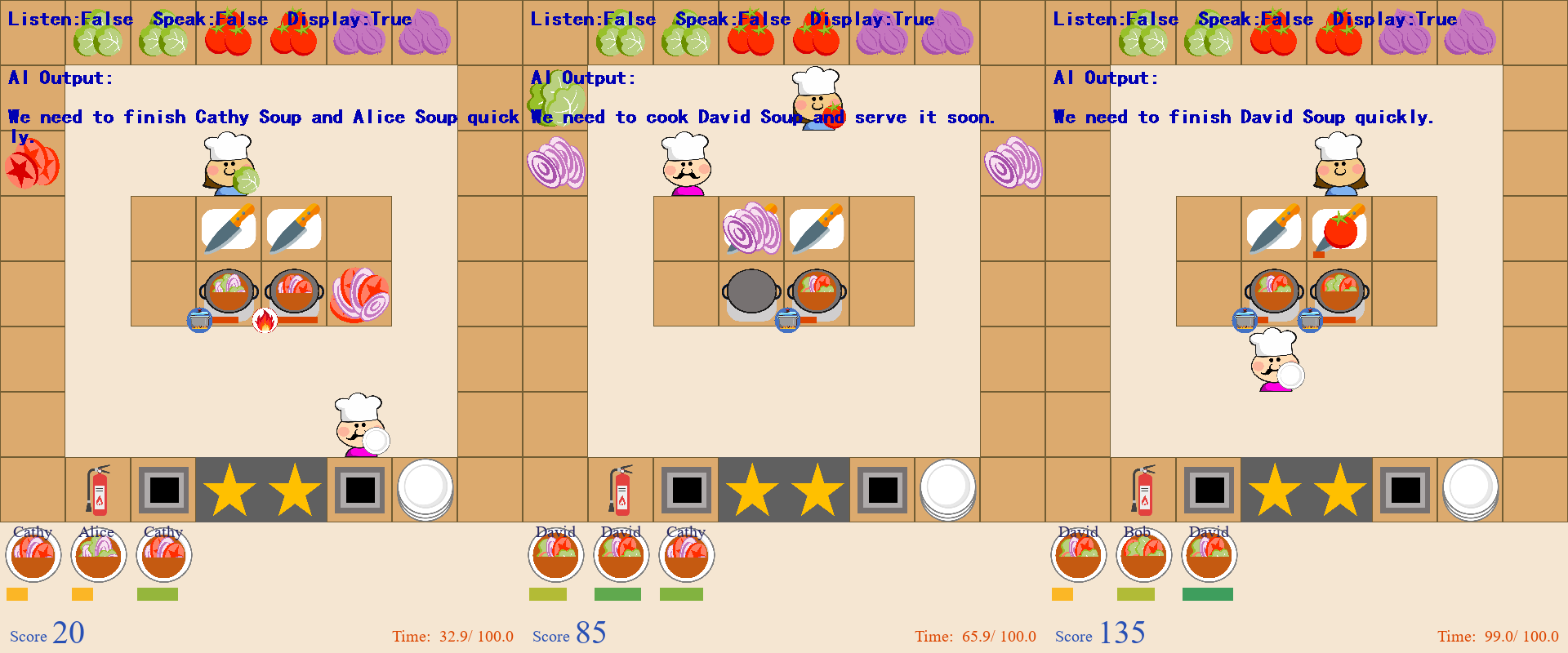}
        \caption{HLA}
    \end{subfigure}

\caption{Visualization result of player-128 in {\mapring} during the competition phase. Screenshots are captured at $1/3$ of gameplay, $2/3$ of gameplay, and upon completion of the game. }
\label{fig:exp-hci-ss-1}
\end{figure*}

\begin{figure*}[ht!]
 \centering
    \begin{subfigure}[t]{0.8\textwidth}
        \includegraphics[width=\textwidth]{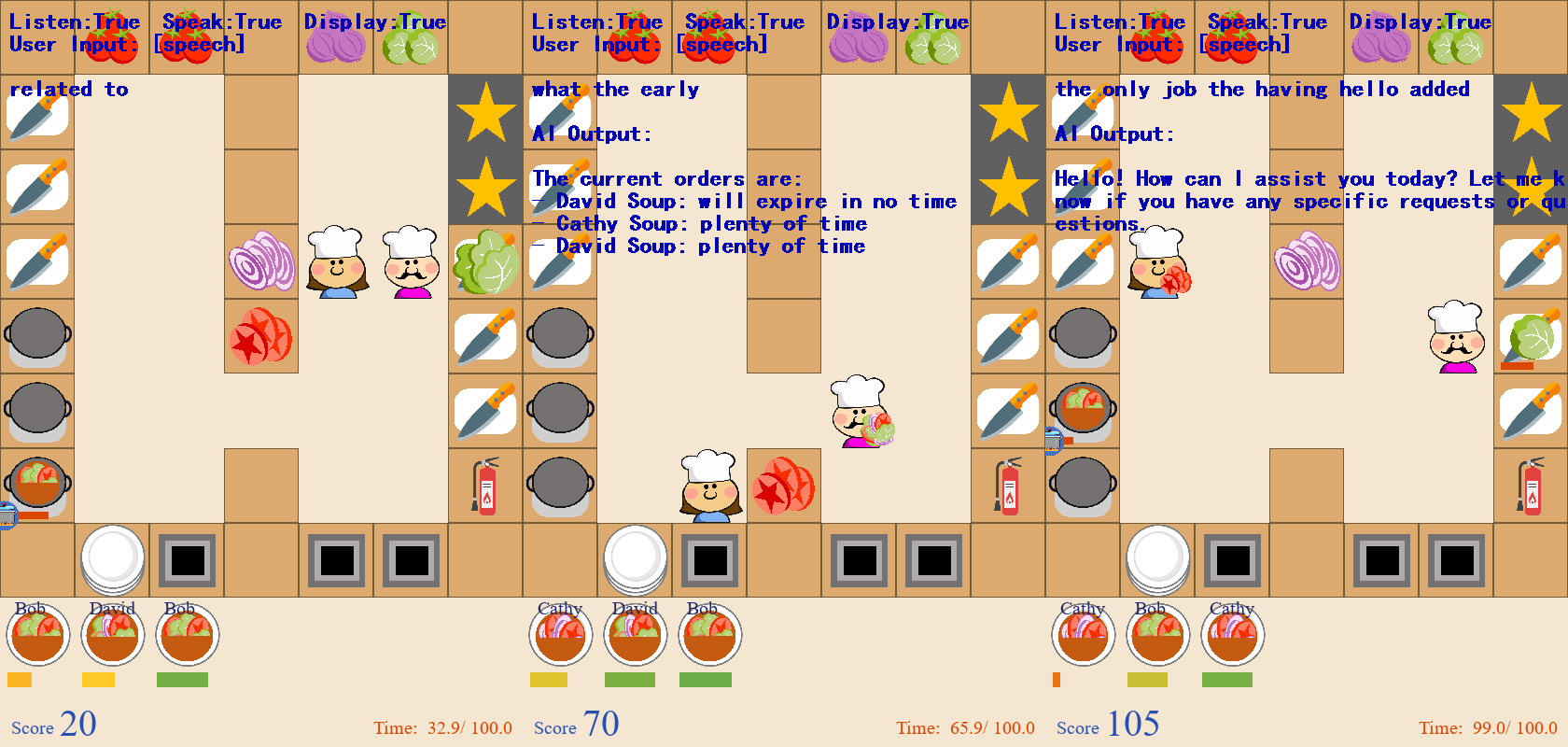}
        \caption{SMOA}
    \end{subfigure}
    \hfill
    \begin{subfigure}[t]{0.8\textwidth}
        \includegraphics[width=\textwidth]{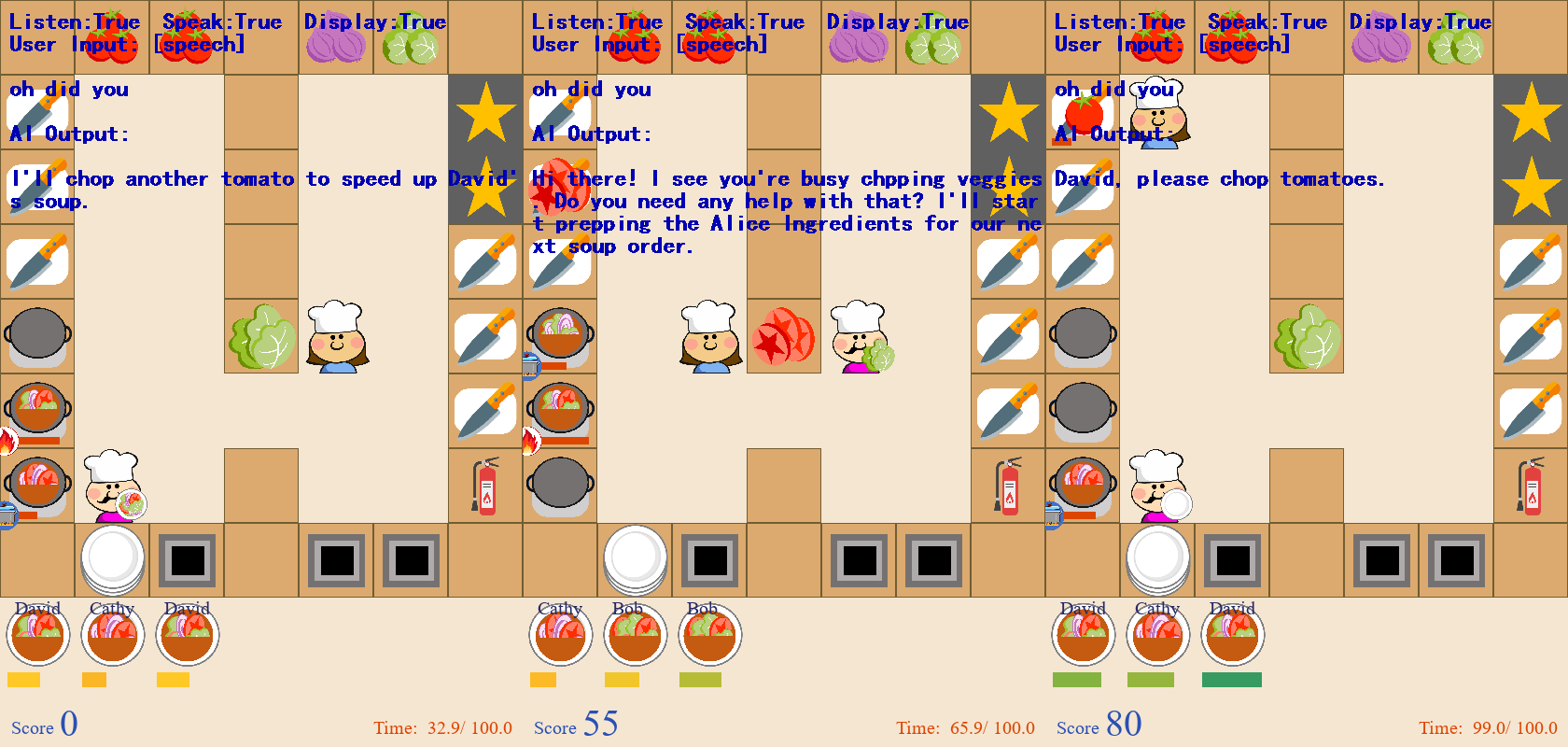}
        \caption{FMOA}
    \end{subfigure}
    
    \begin{subfigure}[t]{0.8\textwidth}
        \includegraphics[width=\textwidth]{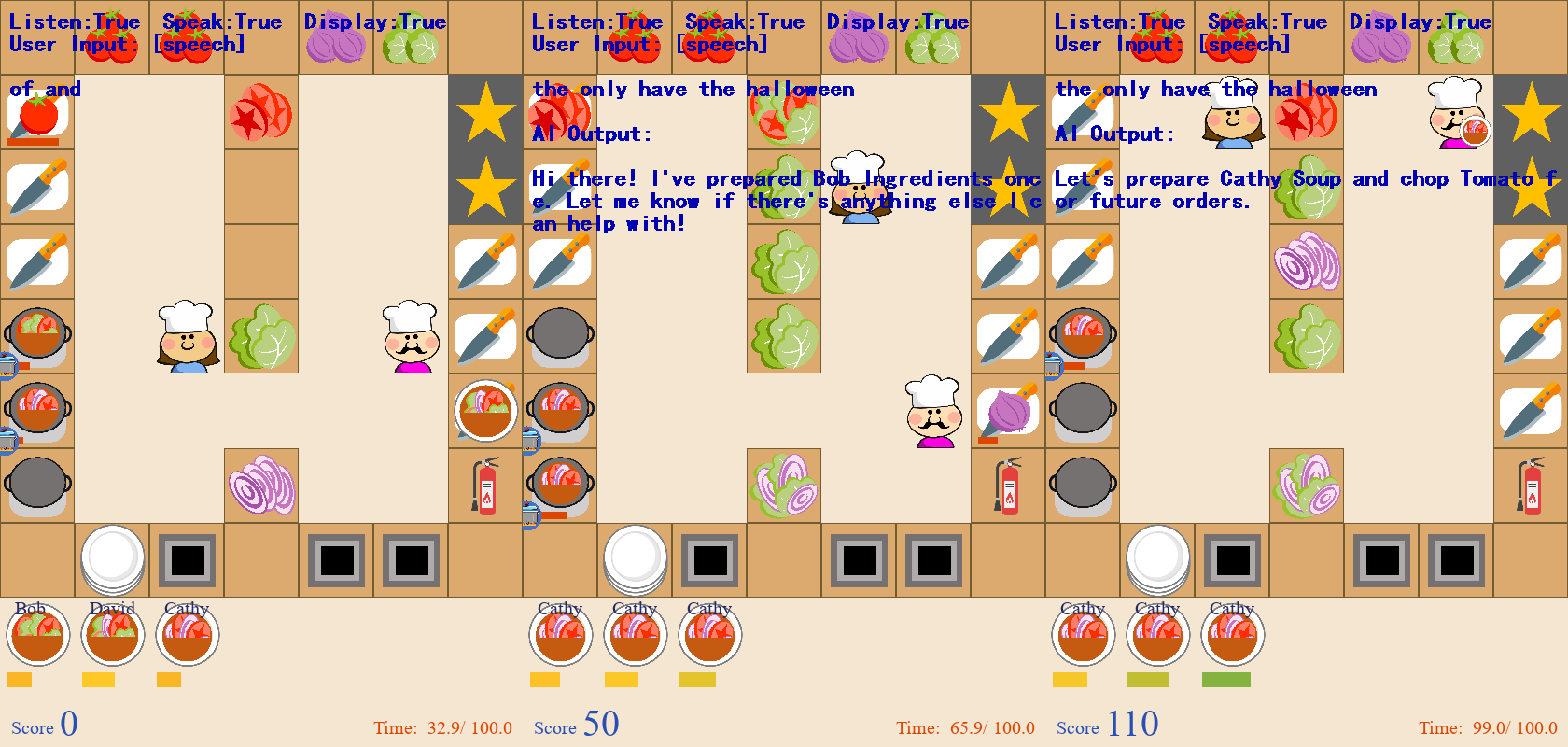}
        \caption{HLA}
    \end{subfigure}

\caption{Visualization result of player-322 in {\mapbot} during the competition phase. Screenshots are captured at $1/3$ of gameplay, $2/3$ of gameplay, and upon completion of the game. }
\label{fig:exp-hci-ss-2}
\end{figure*}

\begin{figure*}[ht!]
 \centering
    \begin{subfigure}[t]{0.8\textwidth}
    \centering
        \includegraphics[height=65mm]{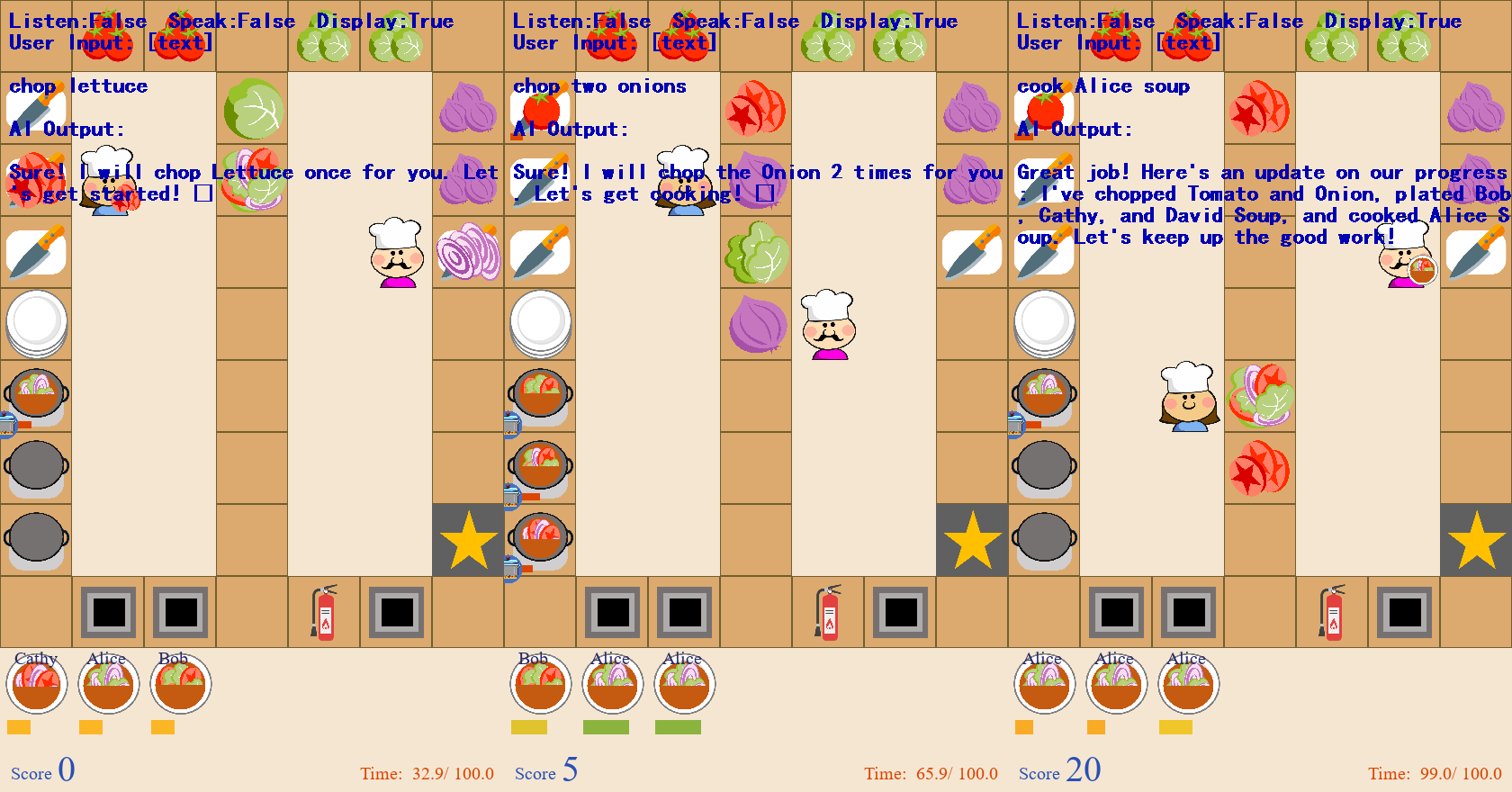}
        \caption{SMOA}
    \end{subfigure}
    \hfill
    \begin{subfigure}[t]{0.8\textwidth}
    \centering
        \includegraphics[height=65mm]{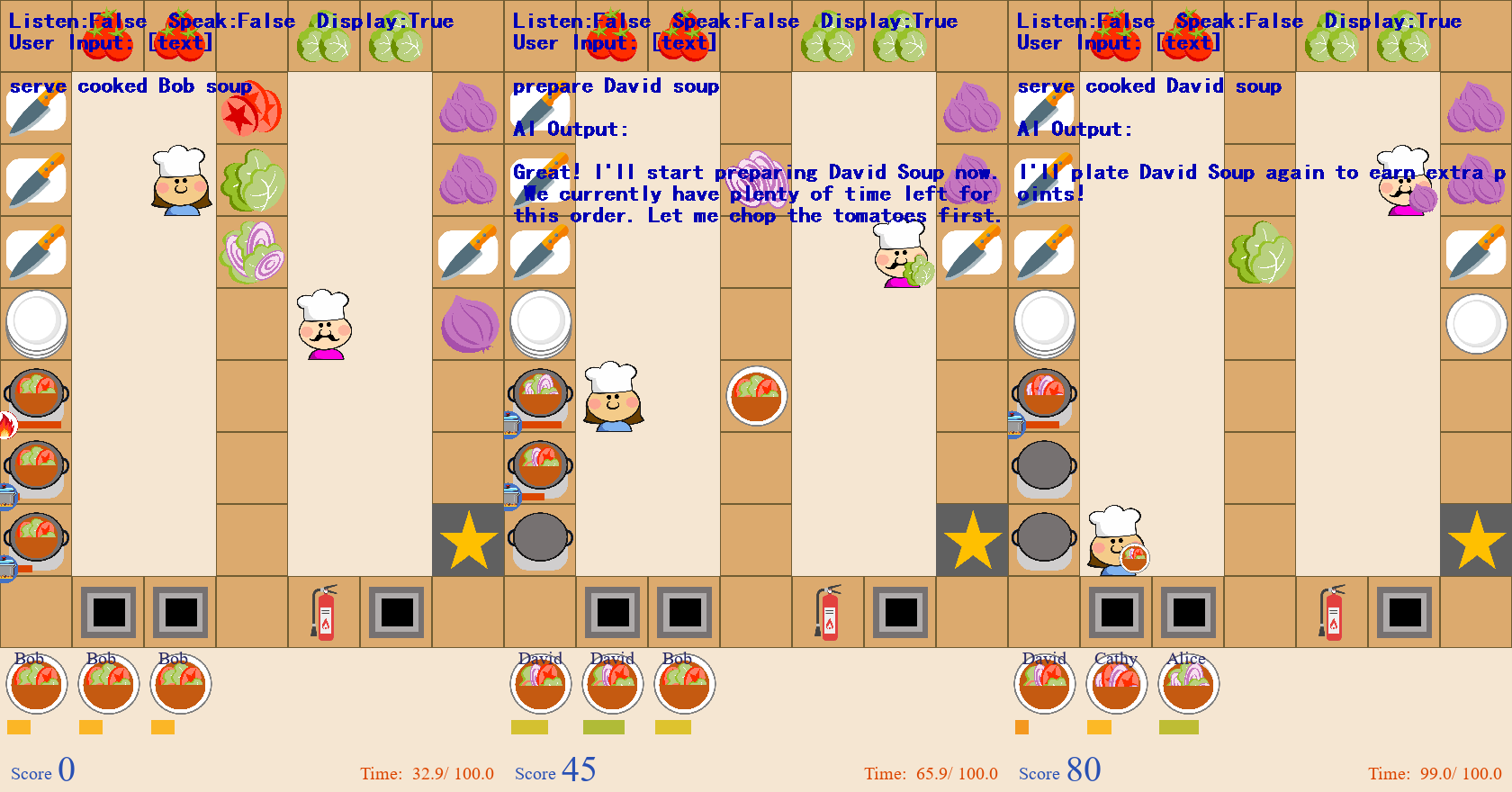}
        \caption{FMOA}
    \end{subfigure}
    
    \begin{subfigure}[t]{0.8\textwidth}
    \centering
        \includegraphics[height=65mm]{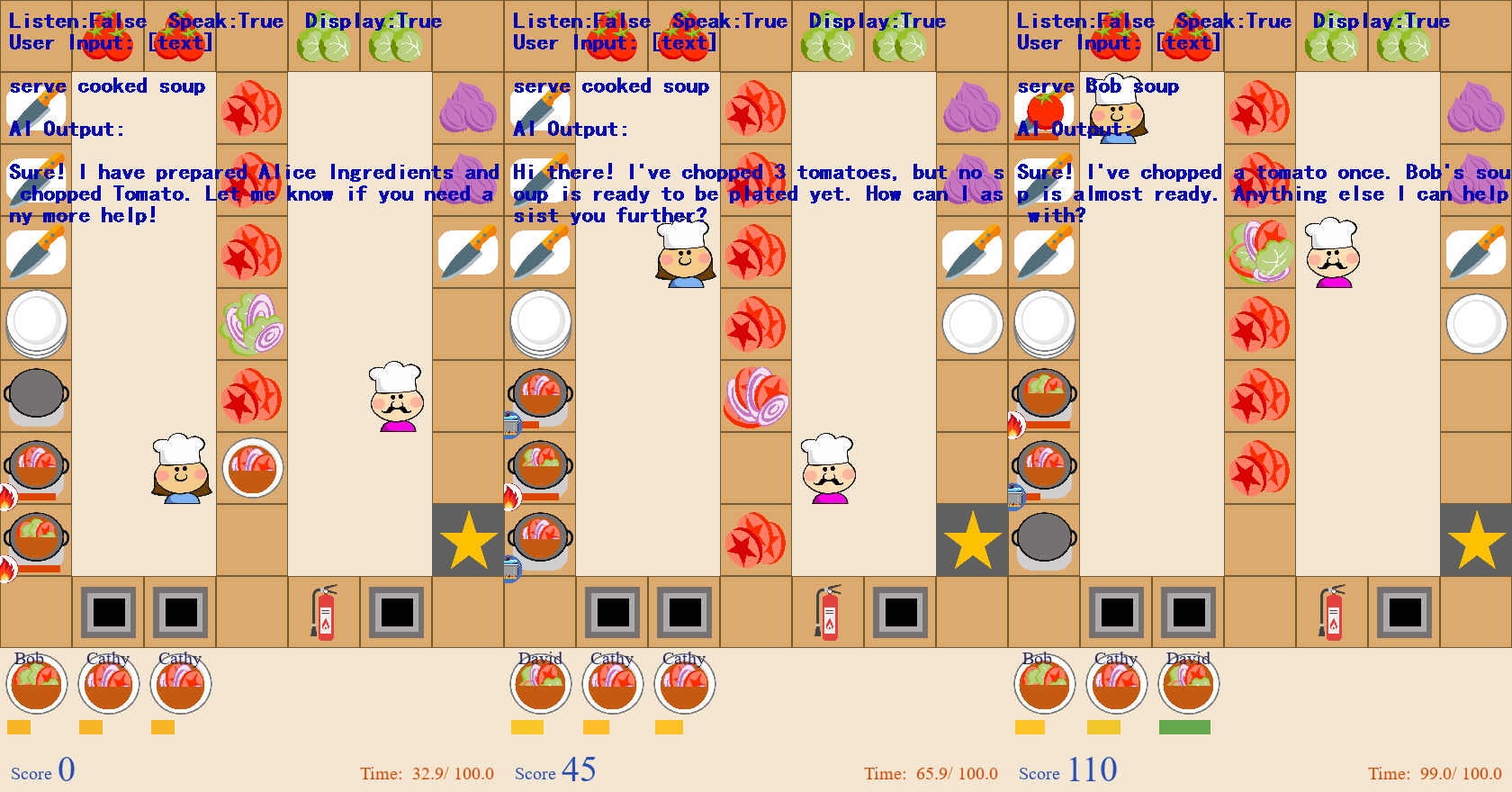}
        \caption{HLA}
    \end{subfigure}

\caption{Visualization result of player-221 in {\mappart} during the competition phase. Screenshots are captured at $1/3$ of gameplay, $2/3$ of gameplay, and upon completion of the game. }
\label{fig:exp-hci-ss-3}
\end{figure*}

\begin{figure*}[ht!]
 \centering
    \begin{subfigure}[t]{0.8\textwidth}
        \includegraphics[width=\textwidth]{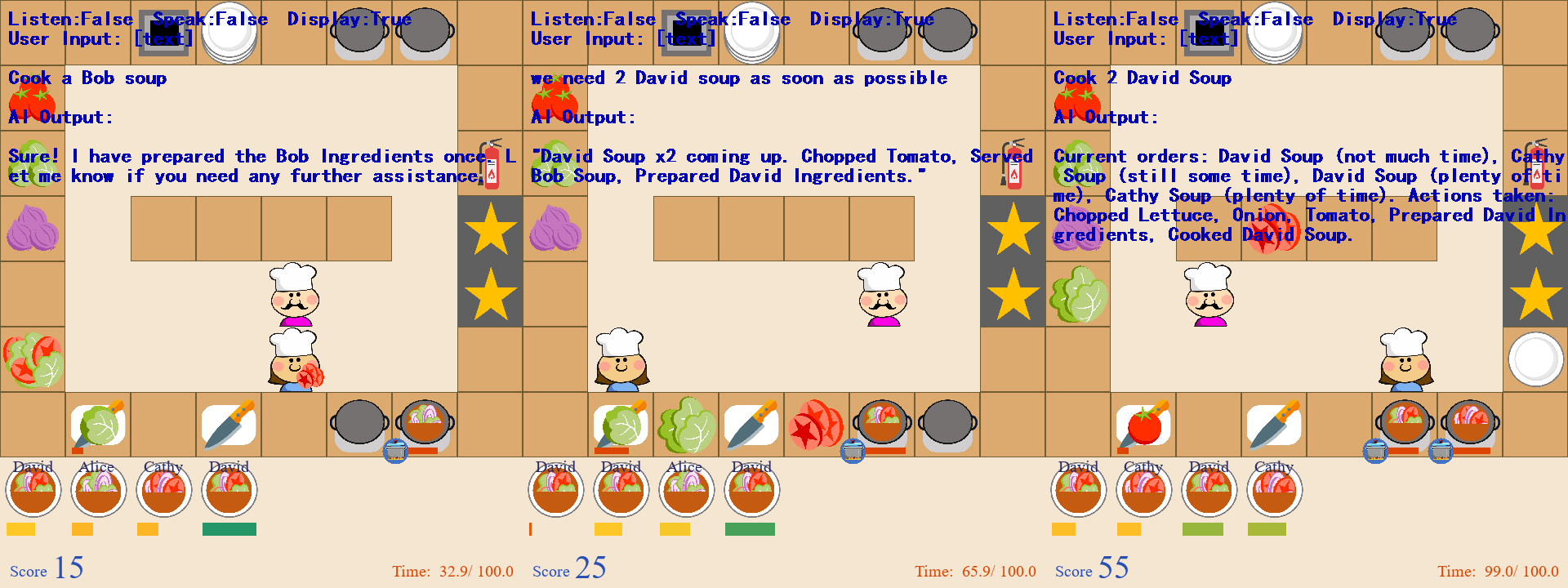}
        \caption{SMOA}
    \end{subfigure}
    \hfill
    \begin{subfigure}[t]{0.8\textwidth}
        \includegraphics[width=\textwidth]{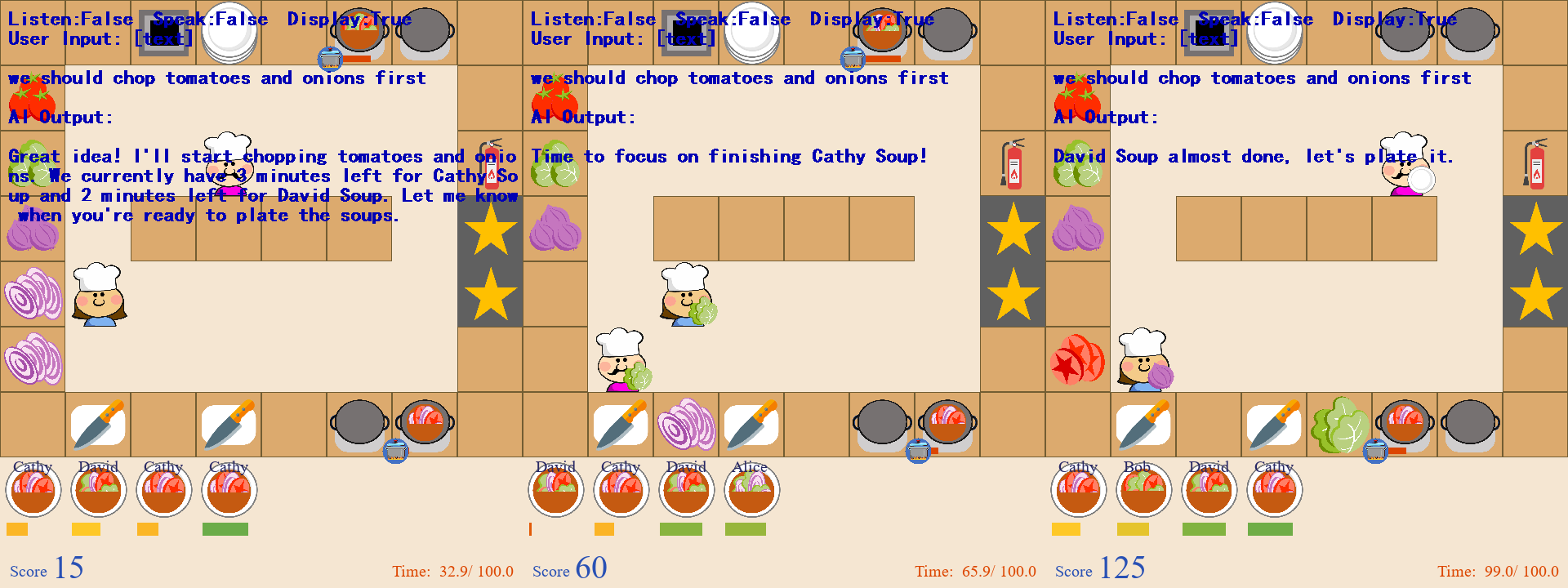}
        \caption{FMOA}
    \end{subfigure}
    
    \begin{subfigure}[t]{0.8\textwidth}
        \includegraphics[width=\textwidth]{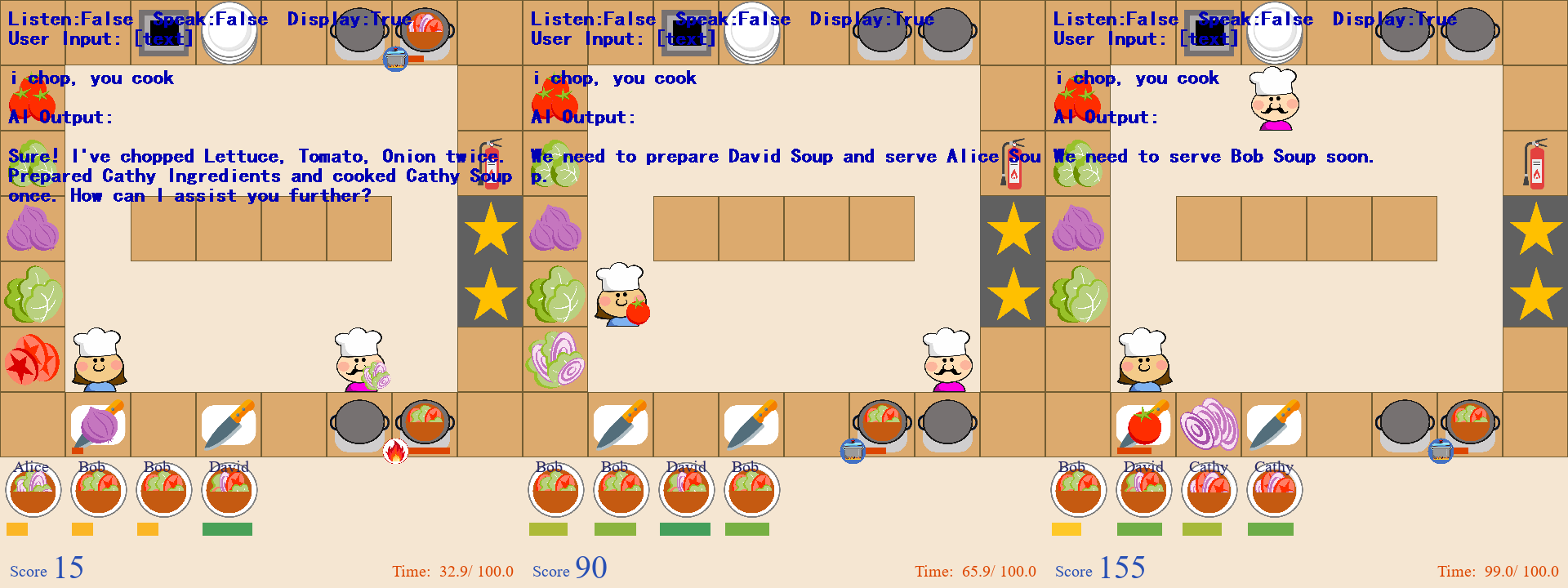}
        \caption{HLA}
    \end{subfigure}

\caption{Visualization result of player-421 in {\maphard} during the competition phase. Screenshots are captured at $1/3$ of gameplay, $2/3$ of gameplay, and upon completion of the game. }
\label{fig:exp-hci-ss-4}
\end{figure*}

We present visualizations depicting four typical human players on different maps during the competition phase. These visualizations are shown Fig.~\ref{fig:exp-hci-ss-1}, Fig.~\ref{fig:exp-hci-ss-2}, Fig.~\ref{fig:exp-hci-ss-3}, and Fig.~\ref{fig:exp-hci-ss-4} respectively. 
The player-128 remains silent through the whole game play. The player-221 and player-421 instruct AI players by texting messages while the player-322 instructs the AI player via speaking. The game screenshots are captured at $1/3$ of gameplay, $2/3$ of gameplay, and upon completion of the game.
The game score can be found in the bottom-left corner of the screenshot. The blue text within the screenshot shows the human player chat message and AI agent chat response.

Cooperating with silent players, AI agent should mainly help the human player get the highest score possible. Moreover, the pace of the game is faster when use speaking to communicate, and texting can the give human player more time to think.
As illustrated in the replay, collaborating with HLA results in the highest game scores. 
The lower latency of HLA enables more ingredients to be chopped and placed on the counters compared to other AI players in both {\mapbot} and {\mappart}. Furthermore, HLA utilizes more pots simultaneously in {\mapring} and {\mappart}.

\end{document}